\pdfoutput=1
\documentclass{iopjournal}

\usepackage{amsmath,amssymb,amsfonts}
\usepackage{bm}
\usepackage[utf8]{inputenc}
\usepackage[T1]{fontenc}
\usepackage{caption}
\usepackage{subcaption}
\usepackage{multirow}
\usepackage{booktabs}
\usepackage{nicefrac}
\usepackage{float}
\usepackage{placeins}
\usepackage[numbers,sort&compress]{natbib}  %
\usepackage{tikz}

\graphicspath{{figures/}}

\newcommand{\uten}{\ensuremath{u_{10}}}
\newcommand{\vten}{\ensuremath{v_{10}}}
\newcommand{\ttwom}{\ensuremath{T_{2}}}
\newcommand{\dtwom}{\ensuremath{T_{d,2}}}

\renewcommand{\articletype}[1]{{\vspace*{-8mm}\noindent \Large \sf Preprint}

\vspace*{8mm} \noindent}

\usepackage{fancyhdr}
\begin{document}

\articletype{Paper}

\title{Benchmarking Generative Models for Near-Surface Data Assimilation on Real Station Observations}

\author{Ruizhe Huang$^{1}$\orcid{0009-0004-9197-1597}, Qidong Yang$^{1}$\orcid{0009-0000-0188-3667}, Jonathan Giezendanner$^{1}$\orcid{0000-0001-8809-8513}, and Sherrie Wang$^{1,*}$}

\affil{$^1$Massachusetts Institute of Technology, Cambridge, MA, USA}

\affil{$^*$Author to whom any correspondence should be addressed.}

\email{sherwang@mit.edu}


\begin{abstract}
Weather reanalysis products rely on computationally intensive numerical weather predictions followed by data assimilation that corrects the forecast toward observations. Recent advances in deep generative models offer a cheaper alternative that shifts much of this cost from inference to offline training. However, existing generative approaches have been evaluated on synthetic observations or under different datasets and evaluation schemes, making it unclear which design choices actually improve real-world data assimilation. We present the first controlled benchmark of generative data assimilation for single-time near-surface analysis from real weather station observations. Using 11{,}849 NOAA MADIS stations across the contiguous United States and four near-surface variables, we hold the dataset, observation operator, and deep learning architecture fixed, and measure spatial generalization at held-out stations. The benchmark compares the major design choices proposed for generative data assimilation, including diffusion versus flow matching, pixel versus latent-space formulations, and multiple inference-time conditioning strategies, against a classical 3D-Var baseline. The benchmark reveals three clear conclusions. First, the best generative methods outperform 3D-Var (35.7\% vs.\ 33.3\% RMSE improvement over ERA5), although 3D-Var receives the ERA5 field at the analysis time as its background and the generative methods receive none. Second, full-gradient guidance consistently outperforms stop-gradient and initial-noise optimization. Third, other choices provide little measurable benefit: diffusion and flow matching perform nearly identically under matched conditions, and latent-space variable mixing does not help. We further evaluate both dense and sparse station settings and find advantages from generative methods and full-gradient guidance more pronounced under sparsity. Together, these results identify which components of generative data assimilation improve spatial generalization in near-surface analysis.
\end{abstract}

\section{Introduction}
\label{sec:introduction}

Weather reanalysis combines historical observations with a fixed numerical weather prediction (NWP) model through data assimilation (DA), producing the gridded atmospheric records that underpin climate research, renewable-energy planning, and extreme-event attribution \citep{slivinski2019towards,kalnay1996ncep}. Reanalysis, however, is expensive to produce. The state-of-the-art global reanalysis ERA5 runs a high-resolution 4D-Var \citep{rabier2003variational} analysis every 12 hours, using a ten-member ensemble to estimate background-error covariances that depend on the current atmospheric state. Even on the European Centre for Medium-Range Weather Forecasts (ECMWF) high-performance computing facility, this process yields only 6--9 days of reanalysis produced per day of wall-clock time  \citep{hersbach2020era5}. ERA5 also provides primary training data for recent machine-learning weather prediction (MLWP) models. These models now match traditional NWP forecast skill at a fraction of the inference cost \citep{pathak2022fourcastnet,bi2023pangu,lam2023learning,price2024gencast,bodnar2025aurora}. At inference, however, most of them still start from ECMWF's operational 4D-Var analysis \citep{moldovan2026aifs}. As a result, data assimilation dominates the computational cost of MLWP pipelines. Replacing 4D-Var with a learned DA method has the potential to cut this cost and make the operational forecasting fully data-driven.

Beyond its cost, classical DA also leaves systematic biases in the reanalyses it produces. It has produced coherent, multi-decadal records \citep{kalnay1996ncep,dee2011era,hersbach2020era5}, but it assumes Gaussian, unbiased errors and linearized dynamics \citep{van2019particle,dee2005bias,grooms2022comparison}. Because DA treats the model forecast as unbiased, the model's systematic errors pass into the reanalysis wherever observations constrain the analysis only weakly. In ERA5, regional evaluations have documented persistent deviations of ERA5 from \emph{in situ} observations in near-surface wind \citep{wilczak2024evaluation,chen2024evaluation,yang2025local} and precipitation \citep{chen2024evaluation}.

\begin{figure}[htbp]
\centering
\includegraphics[width=0.85\textwidth]{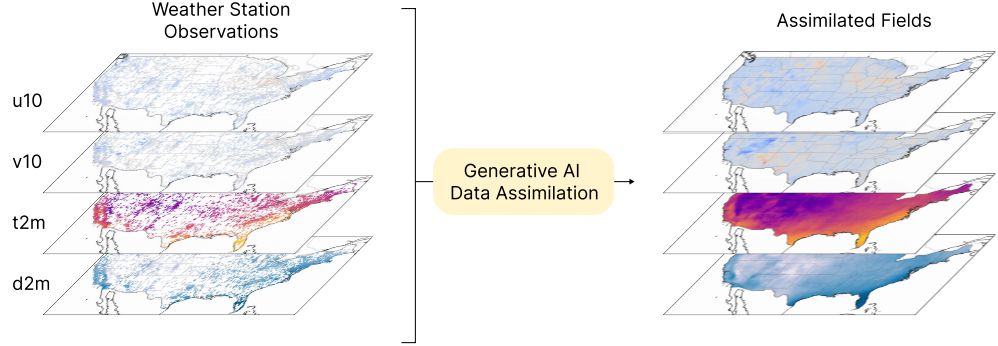}
\caption{Conceptual illustration of generative data assimilation: starting from sparse station observations on the four target variables, a generative prior trained on ERA5 produces a full gridded analysis consistent with the observations.}
\label{fig:concept}
\end{figure}

Just as machine learning has become a competitive alternative to NWP for forecasting, recent work asks whether it can also replace the two components that dominate the cost of 4D-Var: the ensemble that estimates background-error covariances and the iterative variational minimization. Deep generative models address both: the ensemble is replaced by a prior learned offline from reanalysis, and the iterative minimization by sampling from this prior conditioned on observations \citep{rozet2023score,manshausen2025generative} (\autoref{fig:concept}). Several studies further argue that generative DA can reduce the systematic biases of classical reanalysis \citep{qu2024deep,sun2025sda,fan2026physically,qu2025pnp}.

Most generative DA studies, however, evaluate on synthetic observations: 1--5\% of grid points sampled uniformly, so the observation operator $\mathcal{H}$ reduces to a mask \citep{huang2024diffda,wang2024accurate,andry2025appa,sun2025sda,wang2025four,cheng2026flowda}. Real station observations differ in three ways. First, they are off-grid, so $\mathcal{H}$ must interpolate and account for sub-grid terrain within a $\sim$30\,km cell. Second, they carry systematic biases from siting and sensor drift. Third, their density is spatially heterogeneous: methods risk overfitting local noise in dense networks and lack constraint in sparse ones. Studies that do assimilate real observations typically use ERA5 as the reference, either discarding observations that disagree with it or evaluating against it \citep{sun2025sda,qu2024deep,yang2025generative}. This choice cannot test whether generative DA corrects ERA5's biases.

More broadly, generative DA methods have not been compared under controlled conditions, which leaves three questions open (\autoref{sec:related_work}). First, how do generative methods compare with classical 3D-Var methods on real, off-grid station observations? Existing benchmarks either snap observations to a $1.4^\circ$ grid \citep{wang2024dabench} or use masked reanalysis \citep{wang2025dambench}, and Manshausen et al.\ \citep{manshausen2025generative} assimilate real station observations without comparing alternative methods. Second, do the choices between diffusion and flow matching, and between pixel and latent space, matter under a fixed backbone, dataset, and protocol? Third, which design choices within each family actually drive performance?

We address these questions for single-time near-surface analysis over the contiguous United States. We evaluate spatial generalization: how well each method propagates information from assimilated stations to held-out stations. This evaluation complements cycled assimilation experiments, which assess forecast impact. This paper makes three contributions, one for each of the questions above:

\begin{enumerate}
     \item \textbf{A controlled benchmark on real station observations.} We present the first comparison of generative methods and classical 3D-Var methods on real, off-grid station observations: five generative methods and two classical 3D-Var baselines (pixel and latent), evaluated on 11{,}849 MADIS stations across the contiguous United States (CONUS) in 2023 with fixed train/validation/test station splits and identical observation operators. On held-out stations, the best generative method reduces ERA5's root-mean-square error (RMSE) by $35.7\%$, versus $33.3\%$ for 3D-Var, without an explicit background field; the margin grows from $2.4$ percentage points (pp) on the dense benchmark to $7.9$\,pp on the sparse one.

    \item \textbf{A controlled comparison of generative frameworks and conditioning strategies.} With backbone, data, and protocol fixed, the generative framework matters little (flow matching $35.7\%$ vs.\ diffusion $35.4\%$), but the conditioning strategy matters more: full-gradient guidance (Flow Guidance) consistently outperforms stop-gradient guidance (FlowDPS) and initial-noise optimization (D-Flow), especially under sparse observations.

    \item \textbf{Systematic ablation of individual design choices.} Varying the guidance schedule, stop-gradient approximation, terminal extrapolation, solver, and operating space (pixel vs.\ latent) one at a time, we find that several commonly adopted choices, including latent-space variable mixing, provide little measurable benefit.
\end{enumerate}

\section{Related Work}
\label{sec:related_work}

Machine-learning DA methods fall into two families: deterministic models that regress the analysis from background and observations, and generative models that sample from posterior distribution. Among deterministic methods, 4DVarFormer \citep{wang2024accurate} and its adversarially trained variant 4DVarGAN \citep{wang2025four} replace the iterative 4D-Var minimization with a single forward pass through a network trained to predict its solution. FuXi-DA \citep{xu2025fuxi} extends this approach to satellite observations by replacing the observation operator with learned encoders. Like classical 3D- and 4D-Var, these methods return only a point estimate of the analysis and need a separate mechanism, such as an ensemble, to quantify uncertainty.

Generative models provide this uncertainty directly: they model the posterior distribution of the analysis and return samples from it. Most build on diffusion models \citep{ho2020denoising,song2020score,song2020denoising} or flow matching \citep{lipman2022flow}, which learn to map random noise to samples from a target distribution (\autoref{sec:methods}). By default, these models generate unconditional samples. Data assimilation must instead condition generation on observations. We focus on spatial conditioning, in which observations of atmospheric states enter the generative process. Methods for this conditioning divide into two families, depending on whether they must be trained for a specific observation layout.

\paragraph{Observation-layout-dependent conditioning.}
\label{sec:layout_dependent}
Layout-dependent methods train at least one component on a specific observation layout. End-to-end conditional models take the observations directly as input \citep{cheng2026flowda}. Hybrid setups instead freeze an unconditional prior and train a separate amortized controller to inject the observations \citep{srivastava2025control}. Both must be retrained whenever the layout changes, which is impractical for station and satellite networks whose coverage varies in space and time.

\paragraph{Observation-layout-independent conditioning.}
\label{sec:conditioning_strategies}
Layout-independent methods instead train the generative model on atmospheric states alone and inject observations only at inference time. Because the prior never sees an observation layout, a single pretrained model can assimilate any layout. All methods benchmarked in this paper take this approach, which conditions sampling through guidance, optimization, or inpainting.

\emph{Guidance-based methods} correct each sampling step with the gradient of an observation-space loss, steering the trajectory toward states consistent with the observations. They differ in how they compute this gradient and how sampling continues after each correction.

\emph{Full-gradient guidance} backpropagates through the denoiser or velocity network to obtain the gradient with respect to the current noisy state, then continues sampling from the corrected state \citep{chung2023diffusion,pandey2025variational,rozet2023score,rozet2024learning,shysheya2024conditional,manshausen2025generative,andry2025appa,yang2025generative}. These methods further differ in their guidance schedule: how strongly they weight the observation gradient at each noise level (\autoref{sec:methods}).

\emph{Stop-gradient guidance} avoids this backpropagation and takes the gradient with respect to the field itself. Its two variants differ in how sampling resumes: one continues from the corrected state at the same ODE step \citep{meng2022diffusion}, while the other resamples from the clean-data endpoint through the generative model, as in FlowDPS \citep{kim2025flowdps} and related methods \citep{qu2025pnp,sun2025sda}. Studies disagree on which of these three guidance variants performs best, so we ablate all three under a common protocol (\autoref{app:sg}).

\emph{Optimization-based methods} instead fix the generative model and optimize its initial noise so that the generated field minimizes an observation-space loss. In D-Flow \citep{ben2024d}, each optimization step therefore differentiates through the full ordinary differential equation (ODE) solve.

\emph{Inference-time inpainting} replaces observed grid points with their measured values at each denoising step \citep{lugmayr2022repaint,huang2024diffda,yang2025generative}. Because it requires on-grid observations of state variables, it cannot assimilate off-grid station data, and we do not benchmark it.

\paragraph{Latent-space formulations.}
\label{sec:bg_latent_space}
Both guidance and optimization can run either in pixel space or in a compressed latent space learned by an autoencoder. Latent-space assimilation predates generative models: Peyron et al.\ \citep{peyron2021latent} applied an EnKF in an autoencoder's latent space on idealized systems. Recent generative methods apply guidance in such a latent space \citep{fan2026physically,andry2025appa,sun2025sda}. Fan et al.\ \citep{fan2026physically} argue that the autoencoder absorbs nonlinear inter-variable relationships, so the background-error covariance in latent space is approximately diagonal. This argument implies that the autoencoder should mix variables across channels. Existing latent-space methods, however, have tested neither this choice nor the effect of compression depth.

\section{Methods}
\label{sec:methods}

\label{sec:methods_overview}

This section describes the methods compared in this work. We first define the classical baseline, 3D-Var (\autoref{sec:3dvar}), and the two generative priors, diffusion models and flow matching (\autoref{sec:generative_priors}). We then present the inference-time conditioning methods, which steer a frozen prior toward observations without retraining (\autoref{sec:flow_da}). Finally, we move both generative and classical methods into a learned latent space (\autoref{sec:latent_space}) and summarize all methods (\autoref{sec:methods_summary}).

We follow the flow-matching time convention throughout: $t \in [0,1]$ runs from noise ($t=0$) to data ($t=1$). The clean data $\mathbf{x}_1 \sim p_{\text{data}}$ and the initial noise $\mathbf{x}_0$ are the endpoints of the generative trajectory ($\mathbf{x}_0 = \boldsymbol{\epsilon} \sim \mathcal{N}(\mathbf{0}, \mathbf{I})$ for flow matching; \autoref{sec:generative_priors}), and $\mathbf{x}$ denotes the state at the current time; we add the subscript $\mathbf{x}_t$ only when several times appear together. Latent-space methods follow the same conventions with $\mathbf{z}$ in place of $\mathbf{x}$ (\autoref{sec:latent_space}).

\subsection{Classical Baseline: 3D-Var}
\label{sec:3dvar}

3D-Var estimates the analysis by minimizing
\begin{equation}
    \mathcal{L}_{\text{3D-Var}}(\mathbf{x})
    = \frac{w_b}{2}\|\mathbf{x} - \mathbf{x}_b\|_{\mathbf{B}^{-1}}^2
    + \frac{1}{2}\|\mathbf{y}^* - \mathcal{H}(\mathbf{x})\|_{\mathbf{R}^{-1}}^2,
    \label{eq:3dvar}
\end{equation}
where $\mathbf{x}_b$ is the background field, $\mathbf{y}^*$ the observations, $\mathcal{H}$ the observation operator, and $\mathbf{B}$ and $\mathbf{R}$ the background- and observation-error covariances. We use the ERA5 field at the analysis time as the background, so 3D-Var corrects ERA5 directly; the generative methods below use no background field. Because only the ratio of the two terms affects the minimizer, $w_b > 0$ absorbs the overall error scale, and we set $\mathbf{R} = \mathbf{I}$ on standardized variables, assuming equal observation-error variance across variables and stations.

Following the diffusion-based correlation operators used in operational DA \citep{weaver2001correlation}, we set $\mathbf{B}^{-1} = \mathbf{I} + \ell^{2}\mathbf{L}$ for each variable, where $\mathbf{L}$ is the positive semi-definite discrete Laplacian and $\ell$ is a correlation length scale. Larger $\ell$ penalizes rough increments more strongly and therefore yields smoother analyses \citep{lindgren2011explicit}. This $\mathbf{B}$ is univariate and isotropic, far simpler than operational covariances. Operational covariances are estimated from forecast differences \citep{parrish1992national} or ensembles, both of which require a forecast model outside our single-time setting. We tune $\ell$ and $w_b$ on the validation stations and minimize \autoref{eq:3dvar} with L-BFGS, initialized at the background ERA5 fields.

\subsection{Generative Priors: Diffusion Models and Flow Matching}
\label{sec:generative_priors}

We use two generative model families as learned priors: diffusion models and flow matching. Both share the same U-Net backbone and are trained on the same ERA5 fields (\autoref{app:arch}). \autoref{app:edm} develops them in the unified EDM framework \citep{karras2022elucidating}, in which their probability-flow ODEs take the same form and their network outputs are interconvertible through the velocity--denoiser relation (\autoref{eq:velocity_denoiser}).

The diffusion forward process perturbs clean data $\mathbf{x}_1 \sim p_{\text{data}}$ by adding Gaussian noise at level $\sigma$ and applying a scaling function $s(\sigma)$:
\begin{equation}
    \mathbf{x} = s(\sigma)\bigl(\mathbf{x}_1 + \sigma\,\boldsymbol{\epsilon}\bigr), \quad \boldsymbol{\epsilon} \sim \mathcal{N}(\mathbf{0}, \mathbf{I}),
    \label{eq:edm_forward}
\end{equation}
where $\sigma$ ranges from $\sigma_{\min} \approx 0$ (clean data) to $\sigma_{\max}$ (nearly pure noise), and $s(\sigma)$ controls the signal magnitude. The diffusion network is trained as a \emph{denoiser} $D_\theta(\mathbf{x}/s;\,\sigma)$: given a noisy input divided by the scaling factor, it predicts the clean data $\mathbf{x}_1$. Flow matching instead interpolates linearly, $\mathbf{x} = (1-t)\,\mathbf{x}_0 + t\,\mathbf{x}_1$ with $\mathbf{x}_0 = \boldsymbol{\epsilon}$, and trains a velocity network $\mathbf{v}_\theta$. Its terminal extrapolation $\hat{\mathbf{x}}_1 = \mathbf{x} + (1-t)\,\mathbf{v}_\theta(\mathbf{x},t)$ plays the role of $D_\theta$.

Their practical differences lie in training, parameterization, and sampling (\autoref{tab:diffusion_vs_flow}).

\begin{table}[htbp]
\centering
\caption{Comparison of diffusion model and flow matching. $P_{\text{mean}}$ and $P_{\text{std}}$ parameterize the EDM noise-sampling distribution (\autoref{app:edm}).}
\label{tab:diffusion_vs_flow}
\small
\begin{tabular}{lll}
\toprule
 & \textbf{Diffusion Model} & \textbf{Flow Matching} \\
\midrule
Network output & Denoiser $D_\theta \approx \mathbf{x}_1$ & Velocity $\mathbf{v}_\theta \approx \mathbf{x}_1 - \mathbf{x}_0$ \\
Training schedule & $\ln\sigma \sim \mathcal{N}(P_\text{mean}, P_\text{std}^2)$ & $t \sim \mathcal{U}(0,1)$ \\
Score function & $(s\,D_\theta - \mathbf{x})/(s^2\sigma^2)$ & $(t\,\mathbf{v}_\theta - \mathbf{x})/(1-t)$ \\
Sampler & EI predictor + Langevin corrector & Adaptive ODE / Euler / Euler + corrector \\
\bottomrule
\end{tabular}
\end{table}

\subsection{Inference-Time Conditioning}
\label{sec:flow_da}

All generative methods in this work condition a frozen generative model on observations $\mathbf{y}^*$ at inference time, without retraining. The per-step guidance methods share one template. Each approximates the intractable likelihood $p(\mathbf{y}^* \mid \mathbf{x}_t)$ by a Gaussian centered on the denoised estimate $\hat{\mathbf{x}}_1(\mathbf{x}_t)$ with covariance $\boldsymbol{\Sigma}_{\text{eff}}(t)$ (\autoref{app:likelihood_taxonomy}), which gives the likelihood gradient
\begin{equation}
    \nabla_{\mathbf{x}_t} \log p(\mathbf{y}^* \mid \mathbf{x}_t)
    \approx
    J(\mathbf{x}_t)^\top \boldsymbol{\Sigma}_{\text{eff}}(t)^{-1}
    \bigl(\mathbf{y}^* - \mathcal{H}(\hat{\mathbf{x}}_1(\mathbf{x}_t))\bigr),
    \label{eq:guidance_template}
\end{equation}
where $J$ is the Jacobian of $\mathcal{H}(\hat{\mathbf{x}}_1(\mathbf{x}_t))$.

Guidance methods differ in three respects. First, $\boldsymbol{\Sigma}_{\text{eff}}(t)$ sets how strongly observations act at each noise level; we adopt the SDA form by default \citep{rozet2023score,manshausen2025generative,shysheya2024conditional} and summarize the alternatives in \autoref{app:likelihood_taxonomy}. Second, $J$ is either computed exactly by backpropagating through the denoiser or flow network (full-gradient guidance) or approximated by dropping the network Jacobian (stop-gradient guidance). Third, the correction is applied to the current noisy state, to the terminal estimate, or to a latent state. D-Flow falls outside this template: it optimizes the initial noise through the full ODE instead of correcting each step. \autoref{fig:method_overview} maps these choices onto the five generative methods:
\begin{itemize}
    \item \textbf{Diffusion-SDA} (\autoref{sec:diffusion_sda}): full-gradient guidance through the denoiser (panel b).
    \item \textbf{Flow Guidance} (\autoref{sec:guidance}): full-gradient guidance through the velocity network (panel c).
    \item \textbf{FlowDPS} (\autoref{app:sg}): stop-gradient steps on the terminal estimate, followed by resampling (panel d).
    \item \textbf{Latent Flow Guidance} (\autoref{sec:latent_space}): Flow Guidance in an autoencoder's latent space (panel e).
    \item \textbf{D-Flow} (\autoref{sec:dflow}): end-to-end optimization of the initial noise (panel f).
\end{itemize}

\subsubsection{Diffusion-SDA: Score-Based Data Assimilation}
\label{sec:diffusion_sda}
\label{sec:guided_score}

Score-based data assimilation (SDA) was introduced by Rozet and Louppe \citep{rozet2023score} and later applied to weather station observations by Manshausen et al.\ \citep{manshausen2025generative}. The key idea is to modify the score function of a pretrained diffusion model so that generated samples are steered toward observations.

By Bayes' rule \citep{chung2023diffusion,rozet2023score}, the posterior score splits into a prior score $(s\,D_\theta - \mathbf{x})/(s^2\sigma^2)$ and a likelihood score. Following \citep{rozet2023score,manshausen2025generative,shysheya2024conditional}, the SDA likelihood sets the effective covariance to $(\sigma_o^2 + \gamma\,\sigma^2)\mathbf{I}$, where $\sigma_o^2$ is the observation noise variance and $\gamma\sigma^2$ accounts for denoiser uncertainty (\autoref{app:likelihood_taxonomy}). The likelihood score is then $-\mathbf{g}/(\sigma_o^2 + \gamma\sigma^2)$, where $\mathbf{g}$ is the gradient of $\tfrac{1}{2}\|\mathbf{y}^* - \mathcal{H}(D_\theta(\mathbf{x}/s;\,\sigma))\|^2$ with respect to $\mathbf{x}$ (\autoref{eq:obs_gradient}), computed by backpropagation through $D_\theta$ and so implicitly carrying the denoiser Jacobian $\partial D_\theta/\partial \mathbf{x}$, the diffusion analogue of the terminal-extrapolation Jacobian $\partial \hat{\mathbf{x}}_1/\partial \mathbf{x}$ that appears explicitly in Flow Guidance (\autoref{sec:guidance}). Inverting the score--denoiser relation $(s\,D_\theta - \mathbf{x})/(s^2\sigma^2)$ then gives the guided denoiser
\begin{equation}
    D_\theta^{\text{guided}}
    = D_\theta
    - \frac{s\,\sigma^2}{\sigma_o^2 + \gamma\,\sigma^2}\,\mathbf{g}.
    \label{eq:guided_denoiser}
\end{equation}

Substituting the guided denoiser (\autoref{eq:guided_denoiser}) for $D_\theta$ in the Predictor-Corrector (PC) sampler yields the guided predictor and corrector updates, given in \autoref{app:samplers}.

\subsubsection{Flow Guidance}
\label{sec:guidance}

At state $\mathbf{x}_t$, the terminal extrapolation $\mathbf{x} + (1-t)\,\mathbf{v}_\theta(\mathbf{x},t)$ estimates the clean output $\hat{\mathbf{x}}_1 = \mathbf{x} + (1-t)\,\mathbf{v}_\theta(\mathbf{x},t)$, and the observation loss measures the discrepancy between this estimate and the observations, $\mathcal{L}_\text{obs}(\hat{\mathbf{x}}_1) = \tfrac{1}{2}\|\mathcal{H}(\hat{\mathbf{x}}_1) - \mathbf{y}^*\|^2$. The guided velocity field subtracts the gradient of this loss:
\begin{equation}
    \tilde{\mathbf{v}}(\mathbf{x}, t)
    = \mathbf{v}_\theta(\mathbf{x}, t)
    - \lambda(t)\,\nabla_{\mathbf{x}}\,\mathcal{L}_\text{obs}(\hat{\mathbf{x}}_1),
    \label{eq:guided_velocity}
\end{equation}
where $\lambda(t)$ is the guidance strength. The guided ODE $d\mathbf{x}/dt = \tilde{\mathbf{v}}$ is integrated from $\mathbf{x}_0 \sim \mathcal{N}(\mathbf{0},\mathbf{I})$ to $t=1$.

\paragraph{Guidance schedule.}
The schedule $\lambda(t)$ in \autoref{eq:guided_velocity} is the Flow Guidance counterpart of $\boldsymbol{\Sigma}_{\text{eff}}(t)^{-1}$ in \autoref{eq:guidance_template}. Diffusion-SDA uses the same guidance mechanism: rewriting its guided denoiser (\autoref{eq:guided_denoiser}) as a velocity recovers the form of \autoref{eq:guided_velocity} (\autoref{eq:guided_ode}), with the schedule $\lambda_{\text{SDA}}(t)$ of \autoref{eq:sda_schedule} under the flow-matching parameterization. Any accuracy gap between the two methods therefore comes from the prior, the schedule, or the sampler, not from how observations enter the update. We use a constant schedule $\lambda(t) = \lambda$ by default and ablate $\lambda_{\text{SDA}}$ and three heuristic schedules in \autoref{app:schedule}.

\paragraph{Stop-gradient guidance.}
Computing $\nabla_{\mathbf{x}}\mathcal{L}_\text{obs}$ in \autoref{eq:guided_velocity} requires backpropagation through $\mathbf{v}_\theta$ via the terminal extrapolation $\hat{\mathbf{x}}_1 = \mathbf{x} + (1-t)\mathbf{v}_\theta$. Stop-gradient guidance avoids this cost by treating $\mathbf{v}_\theta$ as constant, which drops the network Jacobian from $\nabla_{\mathbf{x}}\mathcal{L}_\text{obs} = \bigl(\mathbf{I} + (1-t)\,\partial\mathbf{v}_\theta/\partial\mathbf{x}\bigr)^{\top}\nabla_{\hat{\mathbf{x}}_1}\mathcal{L}_\text{obs}$ and leaves only the gradient on the terminal state (\autoref{app:sg}). The network Jacobian $\partial \mathbf{v}_\theta/\partial \mathbf{x}$ plays the role of $\mathbf{B}$ in 3D-Var: it spreads each station residual to surrounding grid points through the correlations learned by the prior, whereas stop-gradient variants spread it only indirectly, through subsequent sampling steps. We use the full gradient by default and compare the stop-gradient variants in \autoref{app:sg}.

\paragraph{Sampler.}
The guided ODE can be solved with any of the three samplers from \autoref{sec:fm_sampling}; for the Euler + corrector variant, the Langevin corrector uses the full posterior score, same as Diffusion-SDA (\autoref{app:samplers}).

\subsubsection{D-Flow}
\label{sec:dflow}

D-Flow \citep{ben2024d} optimizes the initial noise $\mathbf{x}_0$ end-to-end through a frozen flow to match observations, rather than modifying the velocity field during sampling.

Starting from $\mathbf{x}_0 \sim \mathcal{N}(\mathbf{0}, \mathbf{I})$, D-Flow propagates $\mathbf{x}_0$ forward through the ODE to obtain the terminal state $\mathbf{x}_1$, then minimizes:
\begin{equation}
    \mathcal{L}_{\text{D-Flow}}(\mathbf{x}_0)
    = \underbrace{\tfrac{1}{2}\bigl\|\mathcal{H}(\mathbf{x}_1) - \mathbf{y}^*\bigr\|_2^2}_{\mathcal{L}_{\text{obs}}}
    + \lambda_{\text{init}}
      \underbrace{\bigl(\tfrac{1}{2}\|\mathbf{x}_0\|_2^2
      - (d-1)\ln\|\mathbf{x}_0\|_2\bigr)}_{\mathcal{L}_{\text{init}}},
    \label{eq:dflow_loss}
\end{equation}
where $\mathcal{L}_{\text{obs}}$ penalizes the observation mismatch and $\mathcal{L}_{\text{init}}$ regularizes $\mathbf{x}_0$ toward the typical set of a $d$-dimensional Gaussian via the negative log-$\chi_d$ density \citep{ben2024d}. The initial noise is updated by backpropagating through the full ODE integration. Throughout optimization we monitor the physical RMSE on a held-out validation set and return the iterate with the lowest validation error.

\subsection{Latent-Space Formulation}
\label{sec:latent_space}

We extend flow matching and 3D-Var framework to operate in a learned latent space to investigate whether dimensionality reduction and the decoder's spatial structure affect assimilation quality.

\subsubsection{Autoencoder}
\label{sec:autoencoder}
A convolutional variational autoencoder (VAE) encodes physical states $\mathbf{x}$ into a latent representation $\mathbf{z} = \mathcal{E}(\mathbf{x})$ and decodes via $\hat{\mathbf{x}} = \mathcal{D}(\mathbf{z})$. The encoder--decoder follows a convolutional pyramid with $S$ scales ($S{-}1$ downsampling stages), reducing spatial resolution by $2^{S-1}$ with a latent channel dimension of 16.

We design a single architecture and vary two factors: whether weather variables interact inside the network ($G$) and the compression depth ($S$). A group parameter $G$ controls all convolutions and attention layers. The \emph{variable-plus-spatial-mixing} variant ($G{=}1$) uses standard convolutions, so the four variables interact at every layer. The \emph{spatial-mixing-only} variant ($G{=}4$) uses grouped convolutions with one group per variable, keeping each variable in an isolated stream so the encoder can only exploit spatial structure. At matched channel width, grouped convolutions have $G\times$ fewer parameters, so $G{=}1$ has roughly $4\times$ more parameters than $G{=}4$. We sweep $S \in \{2, 3, 4\}$ for both settings, yielding six configurations with compression ratios from $0.99\times$ to $14.7\times$. Architecture and training details are given in \autoref{app:arch}.

\subsubsection{Latent 3D-Var}
\label{sec:latent_3dvar}

Fan et al.\ \citep{fan2026physically} argue that an autoencoder absorbs the spatial and inter-variable correlations of the state, so that 3D-Var in its latent space can use a diagonal background-error covariance. We test this idea with the autoencoder of \autoref{sec:autoencoder}, minimizing
\begin{equation}
    \mathcal{L}_{\text{Latent 3D-Var}}(\mathbf{z})
    = \frac{w_b}{2}\|\mathbf{z} - \mathcal{E}(\mathbf{x}_b)\|_2^2
    + \frac{1}{2}\|\mathbf{y}^* - \mathcal{H}(\mathcal{D}(\mathbf{z}))\|_2^2,
    \label{eq:latent_3dvar}
\end{equation}
which sets $\mathbf{B} = \mathbf{I}$ in latent space and $\mathbf{R} = \mathbf{I}$ in observation space. Because the misfit is evaluated after decoding, the latent increment reaches the stations only through $\mathcal{D}$, which must therefore supply the spatial structure that $\mathbf{B}$ provides in pixel space.

\subsubsection{Latent Flow Matching}

After freezing the autoencoder, a U-Net flow model with the same architecture as the pixel-space model is trained on latent trajectories $\mathbf{z} = (1-t)\,\mathbf{z}_0 + t\,\mathcal{E}(\mathbf{x}_1)$ with $\mathbf{z}_0 \sim \mathcal{N}(\mathbf{0}, \mathbf{I})$. Sampling integrates the latent ODE $d\mathbf{z}_t/dt = \mathbf{v}_\theta(\mathbf{z}_t, t)$ from $t=0$ to $t=1$, then decodes via $\mathbf{x}_1 = \mathcal{D}(\mathbf{z}_1)$.

\subsubsection{Latent Flow Guidance}

The pixel-space Flow Guidance method (\autoref{sec:guidance}) transfers to latent space by replacing $\mathbf{x}$ with $\mathbf{z}$: the terminal extrapolation becomes $\hat{\mathbf{z}}_1 = \mathbf{z} + (1-t)\,\mathbf{v}_\theta(\mathbf{z}, t)$ and the guided velocity field takes the form of \autoref{eq:guided_velocity} with $\mathbf{z}$ in place of $\mathbf{x}$. The observation loss is always computed in pixel space after decoding, $\mathcal{L}_\text{obs} = \tfrac{1}{2}\|\mathcal{H}(\mathcal{D}(\hat{\mathbf{z}}_1)) - \mathbf{y}^*\|^2$, so gradients backpropagate through $\mathcal{D}$ before reaching the latent space. Hyperparameters are given in \autoref{app:arch}.

\subsection{Summary of Methods}
\label{sec:methods_summary}

\begin{figure}[htbp]
\centering
\includegraphics[width=\textwidth]{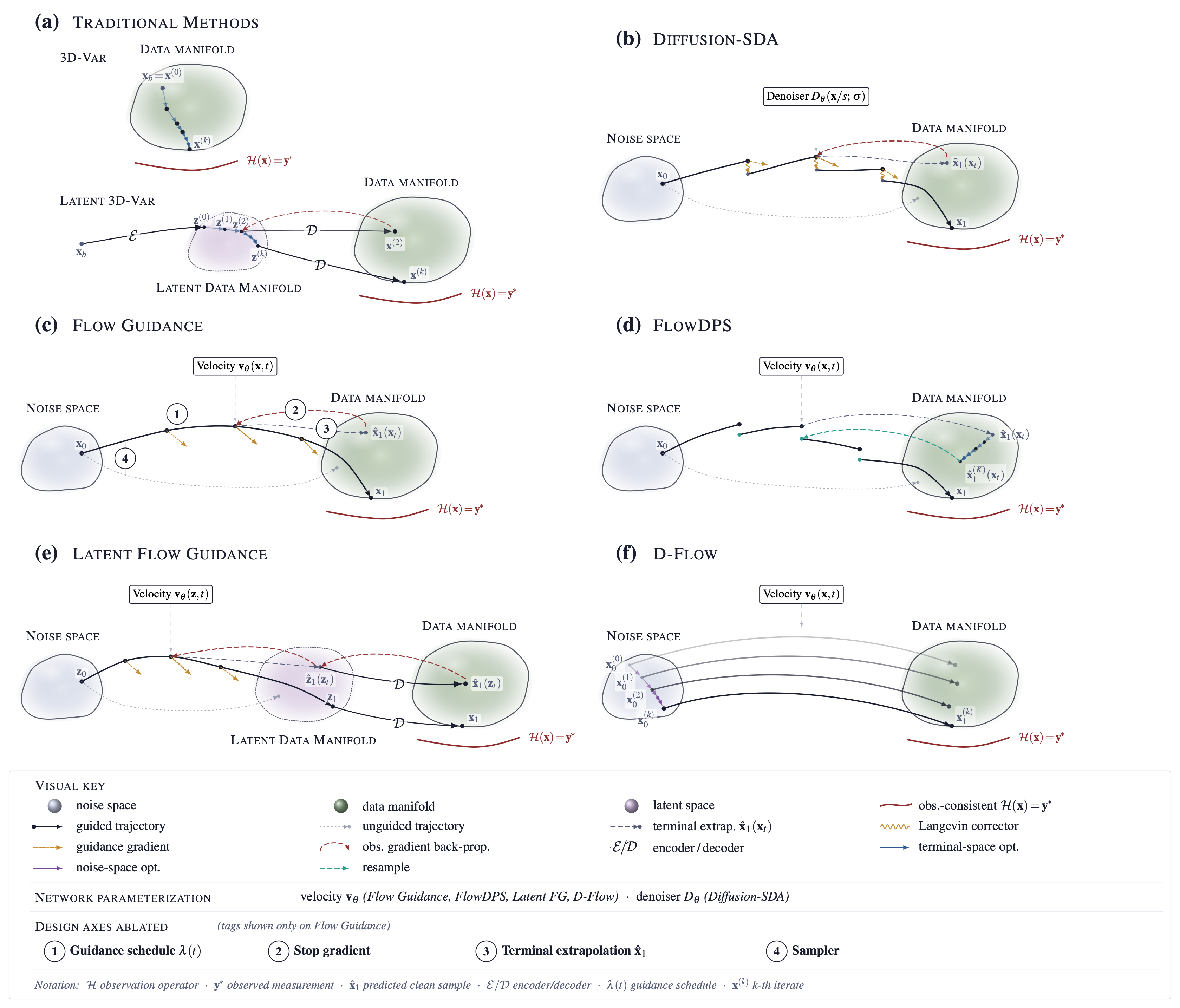}
\caption{Overview of the data assimilation strategies compared in this work. \textbf{(a) Classical methods}: 3D-Var and its latent variant minimize a background-plus-observation cost with no generative prior. \textbf{(b) Diffusion-SDA} (\autoref{sec:diffusion_sda}): per-step guidance through the denoiser $D_\theta$, optionally followed by a Langevin corrector, toward the observation-consistent region $\mathcal{H}(\mathbf{x}) = \mathbf{y}^*$. \textbf{(c) Flow Guidance} (\autoref{sec:guidance}): the flow-matching analogue, backpropagating the observation loss through the terminal extrapolation $\hat{\mathbf{x}}_1(\mathbf{x}_t)$. \textbf{(d) FlowDPS} \citep{kim2025flowdps}: iterates $K$ optimization steps on the terminal estimate $\hat{\mathbf{x}}_1^{(K)}(\mathbf{x}_t)$ and resamples to the next ODE step. \textbf{(e) Latent Flow Guidance}: Flow Guidance inside the latent space of an autoencoder $\mathcal{E}/\mathcal{D}$; the observation loss is evaluated in pixel space via the decoder. \textbf{(f) D-Flow} (\autoref{sec:dflow}): no per-step guidance; an outer loop updates the initial noise $\mathbf{x}_0^{(0)} \to \mathbf{x}_0^{(k)}$ by backpropagating through the full ODE until $\mathbf{x}_1$ satisfies the observations. Numbered circles on panel~(c) mark the four design axes ablated in \autoref{app:design_space}: (1) guidance schedule $\lambda(t)$, (2) stop gradient, (3) terminal extrapolation $\hat{\mathbf{x}}_1$, (4) sampler; a fifth axis (pixel vs.\ latent) is captured by the (c)/(e) contrast.}
\label{fig:method_overview}
\end{figure}

\autoref{fig:method_overview} illustrates the data assimilation strategies compared in this work, and \autoref{tab:method_summary} summarizes them.

\begin{table}[htbp]
\centering
\caption{Summary of data assimilation methods. Columns: model class (classical, diffusion, or flow matching), operating space (pixel or learned latent), DA mechanism (how observations are injected), and default ensemble size $M$.}
\label{tab:method_summary}
\small
\begin{tabular}{@{}l l l l c@{}}
\toprule
\textbf{Method} & \textbf{Model class} & \textbf{Space} & \textbf{DA mechanism} & $M$ \\
\midrule
3D-Var               & Classical     & Pixel  & Variational optimization   & 1  \\
Latent 3D-Var        & Classical     & Latent & Variational optimization   & 1  \\
Diffusion-SDA        & Diffusion     & Pixel  & Full-gradient guidance     & 16 \\
Flow Guidance        & Flow matching & Pixel  & Full-gradient guidance     & 16 \\
FlowDPS              & Flow matching & Pixel  & Stop-gradient guidance     & 16 \\
D-Flow               & Flow matching & Pixel  & Initial-noise optimization & 1  \\
Latent Flow Guidance & Flow matching & Latent & Full-gradient guidance     & 16 \\
\bottomrule
\end{tabular}
\end{table}

\section{Experimental Setup}
\label{sec:data}

\subsection{Data}
\label{sec:data_data}

\subsubsection{ERA5 Reanalysis}

We use hourly ERA5 reanalysis grids (UTC) at $0.25^\circ$ resolution over the contiguous United States ($23$--$51^\circ$N, $127$--$65^\circ$W) for 2016--2023. The domain contains 113 (lat) $\times$ 249 (lon) = 28{,}137 grid points per variable per time step. We evaluate four variables: 10\,m u-wind (\uten, m/s), 10\,m v-wind (\vten, m/s), 2\,m temperature (\ttwom, K), and 2\,m dewpoint (\dtwom, K). All fields are z-scored per variable using training-set (2017--2022) statistics.

\subsubsection{MADIS Weather Station Observations}
\label{sec:madis}

We use surface weather observations from 11{,}849 stations in the National Oceanic and Atmospheric Administration (NOAA) Meteorological Assimilation Data Ingest System (MADIS) \citep{noaa2025madis} archive for 2023, accessed through the curated ERA5--MADIS dataset of Giezendanner et al.\ \citep{giezendanner2026era5madis}. MADIS observations are treated as targets and denoted $\mathbf{y}^*$. Stations are split randomly once (70/15/15) into train/validation/test sets using random seed 42, and this split is held fixed across all experiments.   To test robustness when observations are scarce, we also build a \emph{sparse} benchmark from a random subsample of the full pool (seed 42). The \textbf{dense} benchmark uses all $11{,}849$ stations (Train $8{,}294$ / Val $1{,}777$ / Test $1{,}778$), and the \textbf{sparse} benchmark uses $1{,}000$ (Train 700 / Val 150 / Test 150).

\begin{figure}[htbp]
\centering
\includegraphics[width=0.8\textwidth]{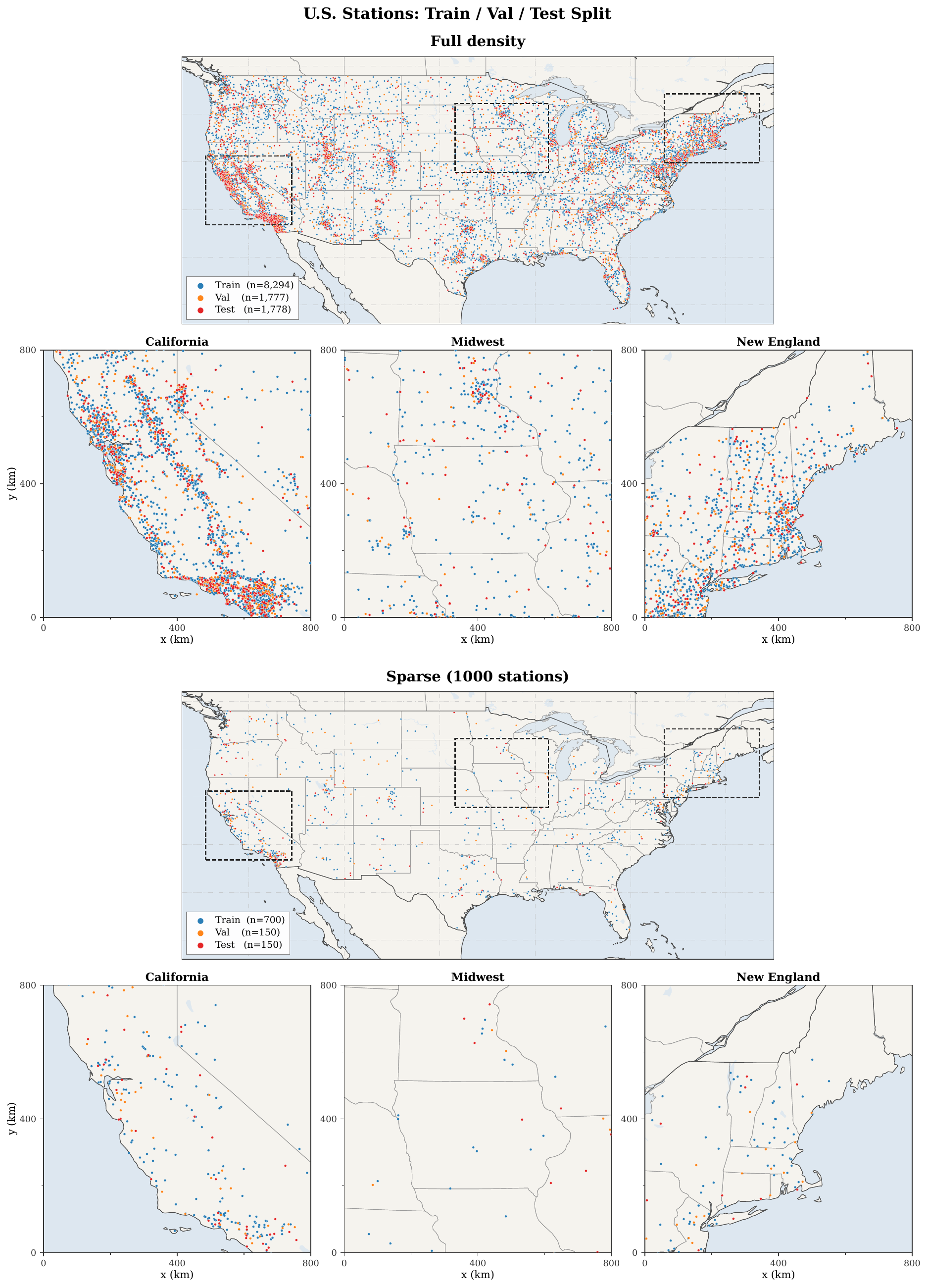}
\caption{Dataset split and regional station density under the two benchmarks. Top block (Dense, 11{,}849 stations): U.S.\ map with stations colored by partition (Train: 8{,}294 / Val: 1{,}777 / Test: 1{,}778) and dashed boxes marking three zoom regions, followed by California, Midwest, and New England panels in Lambert Conformal projection with identical physical extent ($800 \times 800$\,km). Bottom block (Sparse, 1{,}000 stations): same layout for a uniformly subsampled set (Train: 700 / Val: 150 / Test: 150). Matching extents enable fair density comparison across regions and between the dense and sparse benchmarks.}
\label{fig:dataset}
\end{figure}

The weather stations are spatially heterogeneous, with higher density in populated regions like New England and California, and lower density in the Midwest and mountainous areas (\autoref{fig:dataset}).

\subsubsection{Temporal Splits and Evaluation Snapshots}
\label{sec:temporal_splits}

We train the flow-matching model on 2017--2022 ERA5, validate it on 2016, and use 2023 for assimilation and testing. Because ERA5 fields do not depend on the station network, the model is trained once and reused for both benchmarks. Only the observations $\mathbf{y}^*$ differ between them.

We draw assimilation snapshots by sampling one hour uniformly at random within each calendar day of 2023 (seed 12), giving 365 candidates. Five fall in MADIS data outages and are dropped, leaving 360 snapshots in both benchmarks (thresholds and per-snapshot station availability in \autoref{app:full_year}). These snapshots cover all 24 hours of the day (10--26 per hour, mean 15) and all seasons (89 winter, 90 spring, 92 summer, 89 fall).

Station-hour missingness over 2023 averages $18.11\%$ in the dense set and $17.64\%$ in the sparse set, slightly higher for \dtwom\ than for the other variables.\footnote{Dense: \uten/\vten/\ttwom\ $17.73\%$, \dtwom\ $19.26\%$. Sparse: $17.26\%$ and $18.80\%$.}

Network architectures, training hyperparameters, autoencoder configurations, assimilation hyperparameters, and hardware are detailed in \autoref{app:arch}.

\subsection{Evaluation Protocol}
\label{sec:evaluation}

We evaluate all methods using RMSE at held-out test weather stations under both the dense and sparse benchmarks described in \autoref{sec:temporal_splits}. Both benchmarks share the ERA5 baseline, the bicubic observation operator $\mathcal{H}$ (\autoref{app:interpolation}), and the 360 snapshots. We report:
\begin{itemize}
    \item \textbf{Absolute RMSE}: mean $\pm$ std across 360 time steps, for each variable and averaged.
    \item \textbf{RMSE improvement (\%)}: percentage RMSE reduction relative to the ERA5 baseline, computed as $\Delta\% = 100 \times (1 - \text{RMSE}_{\text{method}} / \text{RMSE}_{\text{ERA5}})$. Throughout this paper, ``improvement'' denotes this quantity.
    \item \textbf{Computational cost}: GPU memory (GB) and total time (min).
\end{itemize}

We find empirically that assimilation improvement is strongly correlated with the distance from a test station to its nearest training neighbors, as nearby stations benefit from denser local constraints. A simple per-station average therefore overrepresents the dense clusters and masks performance variation across different levels of observational sparsity.

To expose this variation, we compute for each test station the mean distance to its five nearest training stations with observations at that snapshot. We then bin test stations by this distance and report metrics per bin, which shows how performance changes with distance. The same binning procedure is applied in both the dense and sparse benchmarks, though the sparse benchmark's reduced station count naturally shifts stations toward larger distance bins.

\section{Results}
\label{sec:results}

We present results for the methods of \autoref{sec:methods} under the protocol of \autoref{sec:evaluation}. We first compare methods on the dense $11{,}849$-station benchmark, including pixel- versus latent-space assimilation (\autoref{sec:dense_benchmark}). We then stress-test them on the $1{,}000$-station sparse benchmark (\autoref{sec:sparse_benchmark}). Finally, we analyze how improvement decays with 5-NN distance (\autoref{sec:spatial}) and examine regional case studies (\autoref{sec:case_studies}). Two further analyses appear in the appendix: the role of the Langevin corrector (\autoref{app:corrector}, \autoref{app:perturbation}) and ablations of the Flow Guidance design space (\autoref{app:design_space}).

\subsection{Dense Benchmark: The Best Generative Methods Outperform 3D-Var}
\label{sec:dense_benchmark}

We evaluate the headline pixel-space methods (3D-Var, Diffusion-SDA, Flow Guidance, and D-Flow) against the ERA5 baseline; FlowDPS is evaluated with the stop-gradient family in \autoref{app:sg} (best $28.2\%$ at $K{=}20$). All methods use the default hyperparameters listed in \autoref{app:arch}; deviations are noted where applicable. \autoref{tab:benchmark_main} consolidates per-variable test improvements, the train/test gap, memory and total compute time cost; full RMSE and validation-station numbers are in \autoref{app:full_benchmark}. Re-running the three headline methods on all $8{,}664$ valid hours of 2023 changes their average test improvement by at most $0.1$\,pp and leaves their ranking unchanged (\autoref{app:full_year}), so the $360$ snapshots are representative of the full year.

\begin{table}[htbp]
\centering
\caption{Dense benchmark on the $11{,}849$-station network. RMSE improvement (\%) over ERA5 at training (Tr) and held-out test (Te) stations; ``Average'' is the unweighted mean across the four variables. ``Mem'' is peak GPU memory (GB); ``Total'' is wall-clock time (min) for $360$ snapshots on a single RTX8000. Higher improvement and lower cost are better. ``PO'' = predictor-only (no Langevin corrector).}
\label{tab:benchmark_main}
\tiny
\begin{tabular}{l cccc cccc cc rr}
\toprule
& \multicolumn{2}{c}{\uten} & \multicolumn{2}{c}{\vten} & \multicolumn{2}{c}{\ttwom} & \multicolumn{2}{c}{\dtwom} & \multicolumn{2}{c}{Average} & \multicolumn{2}{c}{Cost} \\
\cmidrule(lr){2-3}\cmidrule(lr){4-5}\cmidrule(lr){6-7}\cmidrule(lr){8-9}\cmidrule(lr){10-11}\cmidrule(lr){12-13}
Method & Tr & Te & Tr & Te & Tr & Te & Tr & Te & Tr & Te & Mem & Total \\
\midrule
3D-Var                   & 51.7 & 35.2 & 51.8 & 35.1 & 46.0 & 29.2 & 49.5 & 33.6 & 49.7 & 33.3 & 0.17 & 4 \\
3D-Var (w/o ERA5 loss)  & 66.0 & 15.0 & 66.2 & 13.7 & 61.8 &  9.4 & 64.8 & 12.3 & 64.7 & 12.6 & 0.20  & 36 \\
Diffusion-SDA            & 52.7 & 37.8 & 52.2 & 38.6 & 39.9 & 29.6 & 47.6 & 35.6 & 48.1 & 35.4 & 6.12 & 246 \\
Diffusion-SDA (PO)       & 48.3 & 37.9 & 47.9 & 38.3 & 29.5 & 19.9 & 36.7 & 25.0 & 40.6 & 30.3 & 6.11 & 82 \\
D-Flow                   & 50.7 & 34.7 & 49.6 & 35.8 & 33.3 & 23.0 & 39.8 & 28.5 & 43.4 & 30.5 & 36.01 & 2563 \\
Flow Guidance            & 53.2 & 38.2 & 53.0 & 38.8 & 42.9 & 29.9 & 49.1 & 35.9 & 49.6 & 35.7 & 3.05 & 380 \\
\bottomrule
\end{tabular}
\end{table}

\paragraph{Flow Guidance and Diffusion-SDA tie, both ahead of 3D-Var.}
Flow Guidance and Diffusion-SDA reach essentially the same test improvement over ERA5 ($35.7\%$ vs.\ $35.4\%$; \autoref{tab:benchmark_main}), ahead of 3D-Var by ${\sim}2$\,pp and D-Flow by ${\sim}5$\,pp. Their generalization gaps, defined as training minus test improvement, are also similar ($13.9$ vs.\ $12.7$\,pp). 3D-Var's gap is comparable ($16.4$\,pp). Without its ERA5 background term, however, its training-station improvement rises to $64.7\%$ while test improvement falls to $12.6\%$, a ${\sim}52$\,pp gap.

\paragraph{Flow Guidance is the most practical choice.}
Total wall-clock time spans roughly $640\times$ (\autoref{tab:benchmark_main}): $4$ minutes for 3D-Var, $246$ for Diffusion-SDA, $380$ for Flow Guidance, and $2{,}563$ for D-Flow. Among the two leading methods, Flow Guidance is ${\sim}1.5\times$ slower than Diffusion-SDA but uses half the peak memory ($3.05$ vs.\ $6.12$\,GB) and has one hyperparameter to tune instead of five ($\lambda$ versus $\sigma_o$, $\gamma$, $\eta$, $\tau$, $K$). It is therefore the more practical choice when memory or tuning budget is limited.

\paragraph{Latent-space assimilation does not improve on pixel space.}
\label{sec:latent}
We also run Flow Guidance in the latent space of an autoencoder, varying how much it compresses the field ($S$) and whether it lets the four variables interact ($G{=}1$) or keeps them separate ($G{=}4$). Neither latent space, deeper compression, nor variable mixing improves accuracy. First, no latent configuration beats pixel space: the best one ($S{=}2$) is $1.4$\,pp less accurate and ${\sim}10\%$ slower, and its only advantage is peak memory ($1.85$ vs.\ $3.05$\,GB). Second, compressing further lowers accuracy and raises cost. The best improvement falls from $34.3\%$ at $S{=}2$ to $32.8\%$ at $S{=}3$ and $31.2\%$ at $S{=}4$, mostly on temperature and dewpoint, while reconstruction error grows ${\sim}20\times$ and wall-clock time increases (\autoref{fig:latent}, \autoref{tab:ae_recon}). Third, letting variables interact does not help: at $S{=}2$, keeping them separate ($G{=}4$) matches mixing them ($G{=}1$) on both assimilation and reconstruction accuracy with $4\times$ fewer parameters. The same holds for classical 3D-Var: 3D-Var reaches $33.3\%$ in pixel space and $33.3\%$ and $33.5\%$ in the $G{=}1$ and $G{=}4$ latent spaces (\autoref{tab:ae_mixing_3dvar}). The full accuracy--compute trade-off and the compression-depth sweep are in \autoref{app:latent_study}.

\paragraph{Wind improves more than temperature and dewpoint; the corrector helps only the latter.}
All methods improve wind more than temperature and dewpoint (\autoref{tab:benchmark_main}). The Langevin corrector's gain is confined to the latter: it adds $5.1$\,pp to Diffusion-SDA on average, all of it on temperature and dewpoint (\ttwom\ $+9.7$\,pp, \dtwom\ $+10.6$\,pp; wind within $\pm0.3$\,pp).

\subsection{Sparse Benchmark: The Generative Advantage Widens}
\label{sec:sparse_benchmark}

The preceding sections evaluate all methods on the full $11{,}849$-station network. To assess robustness when observations are scarce, we repeat the evaluation on the $1{,}000$-station sparse benchmark (\autoref{sec:temporal_splits}). \autoref{tab:sparse_benchmark} includes the best configurations identified in the design-space ablations (\autoref{app:design_space}). Four findings stand out.

\begin{table}[htbp]
\centering
\caption{Sparse benchmark, $1{,}000$ randomly subsampled stations. Columns as in \autoref{tab:benchmark_main}. ``SG'' applies the stop-gradient approximation; ``Euler-128'' replaces the adaptive dopri5 solver with a $128$-step fixed-step Euler solver; ``PO'' = predictor-only; ``PC'' adds a Langevin corrector.}
\label{tab:sparse_benchmark}
\tiny
\setlength{\tabcolsep}{2.5pt}\resizebox{0.85\textwidth}{!}{%
\begin{tabular}{l cccc cccc cc rr}
\toprule
& \multicolumn{2}{c}{\uten} & \multicolumn{2}{c}{\vten} & \multicolumn{2}{c}{\ttwom} & \multicolumn{2}{c}{\dtwom} & \multicolumn{2}{c}{Average} & \multicolumn{2}{c}{Cost} \\
\cmidrule(lr){2-3}\cmidrule(lr){4-5}\cmidrule(lr){6-7}\cmidrule(lr){8-9}\cmidrule(lr){10-11}\cmidrule(lr){12-13}
Method & Tr & Te & Tr & Te & Tr & Te & Tr & Te & Tr & Te & Mem & Total \\
\midrule
3D-Var                              & 49.0 & 14.1 & 49.9 & 15.4 & 46.6 & 11.7 & 48.8 & 14.1 & 48.6 & 13.8 & 0.11 & 3 \\
Latent 3D-Var ($S{=}2$, $G{=}1$)     & 71.2 & 14.6 & 71.5 & 16.4 & 65.7 & 13.3 & 69.2 & 15.5 & 69.4 & 15.0 & 0.25 & 11 \\
Latent 3D-Var ($S{=}2$, $G{=}4$)     & 68.1 & 15.2 & 68.8 & 16.4 & 56.6 & 13.6 & 63.2 & 15.5 & 64.2 & 15.2 & 0.34 & 13 \\
Diffusion-SDA                       & 69.3 & 25.7 & 69.0 & 28.1 & 46.6 & 14.2 & 60.0 & 17.9 & 61.2 & 21.5 & 6.12 & 242 \\
Diffusion-SDA (PO)                  & 62.0 & 28.6 & 61.1 & 28.6 & 30.3 & $-2.4$ & 42.6 & $-2.8$ & 49.0 & 13.0 & 6.10 & 83 \\
D-Flow                              & 61.1 & 21.5 & 59.6 & 23.1 & 30.4 & 3.1 & 40.9 & 2.7 & 48.0 & 12.6 & 36.01 & 2085 \\
Flow Guidance                       & 73.4 & 26.4 & 74.2 & 28.4 & 62.0 & 13.4 & 69.5 & 18.6 & 69.8 & 21.7 & 3.04 & 142 \\
Flow Guidance (SG)                  & 81.0 & 21.0 & 82.3 & 21.2 & 76.5 & $-8.8$ & 80.6 & $-3.3$ & 80.1 & 7.6 & 0.36 & 40 \\
Flow Guidance (Euler-128, PO)       & 61.2 & 29.1 & 61.2 & 30.2 & 37.6 & 8.4 & 47.8 & 13.3 & 52.0 & 20.3 & 2.91 & 52 \\
Flow Guidance (Euler-128, PC)       & 61.3 & 28.9 & 61.2 & 31.1 & 39.1 & 11.4 & 50.1 & 18.1 & 52.9 & 22.4 & 2.91 & 208 \\
Flow Guidance (latent, $G{=}1$)     & 60.2 & 29.2 & 59.6 & 30.5 & 33.8 &  9.5 & 42.8 & 15.5 & 49.1 & 21.2 & 1.84 & 93 \\
Flow Guidance (latent, $G{=}4$)     & 64.6 & 28.4 & 64.8 & 29.9 & 41.7 & 11.4 & 53.4 & 17.5 & 56.1 & 21.8 & 1.84 & 115 \\
\bottomrule
\end{tabular}%
}
\end{table}

\paragraph{Sparsity separates the methods.}
Cutting the number of stations from $11{,}849$ to $1{,}000$ lowers the average test improvement from $30$--$36\%$ to $7.6$--$22.4\%$. It also spreads the methods apart: a ${\sim}5$\,pp range on the dense benchmark becomes ${\sim}15$\,pp here. The largest drops fall on stop-gradient Flow Guidance ($7.6\%$), D-Flow ($12.6\%$), and Diffusion-SDA without its corrector ($13.0\%$). 

\paragraph{Flow Guidance and Diffusion-SDA remain comparable.}
The two leading methods stay at parity under sparsity ($21.7\%$ vs.\ $21.5\%$), and Flow Guidance with the Euler-128 PC sampler is the best single configuration ($22.4\%$). The corrector again matters more for Diffusion-SDA ($8.5$\,pp) than for Flow Guidance ($2.1$\,pp).

\paragraph{Latent space matches pixel space at lower cost under sparsity.}
Pixel space no longer holds the accuracy advantage it had on the dense benchmark. Latent Flow Guidance ($S{=}2$, $G{=}4$) matches pixel Flow Guidance ($21.8\%$ vs.\ $21.7\%$) while using ${\sim}40\%$ less memory ($1.84$ vs.\ $3.04$\,GB) and running ${\sim}19\%$ faster ($115$ vs.\ $142$\,min). Latent 3D-Var even slightly outperforms pixel 3D-Var ($15.0$--$15.2\%$ vs.\ $13.8\%$).

\paragraph{Stop-gradient guidance and D-Flow collapse.}
On the dense benchmark, both methods trailed the leaders by only a few points ($32.2\%$ and $30.5\%$ against $35.7\%$). Here they fall to $7.6\%$ and $12.6\%$, and they show different training patterns. Stop-gradient guidance fits the assimilated stations better than any other method ($80.1\%$), yet temperature and dewpoint end up worse than ERA5 at test stations ($-8.8\%$ and $-3.3\%$). D-Flow fits even the assimilated stations poorly ($48.0\%$, against $69.8\%$ for Flow Guidance), and its temperature and dewpoint barely improve on test stations ($3.1\%$ and $2.7\%$).

\subsection{Spatial Propagation of Corrections}
\label{sec:spatial}

\begin{figure}[htbp]
\centering
\begin{subfigure}[t]{0.49\textwidth}
    \centering
    \includegraphics[width=\linewidth]{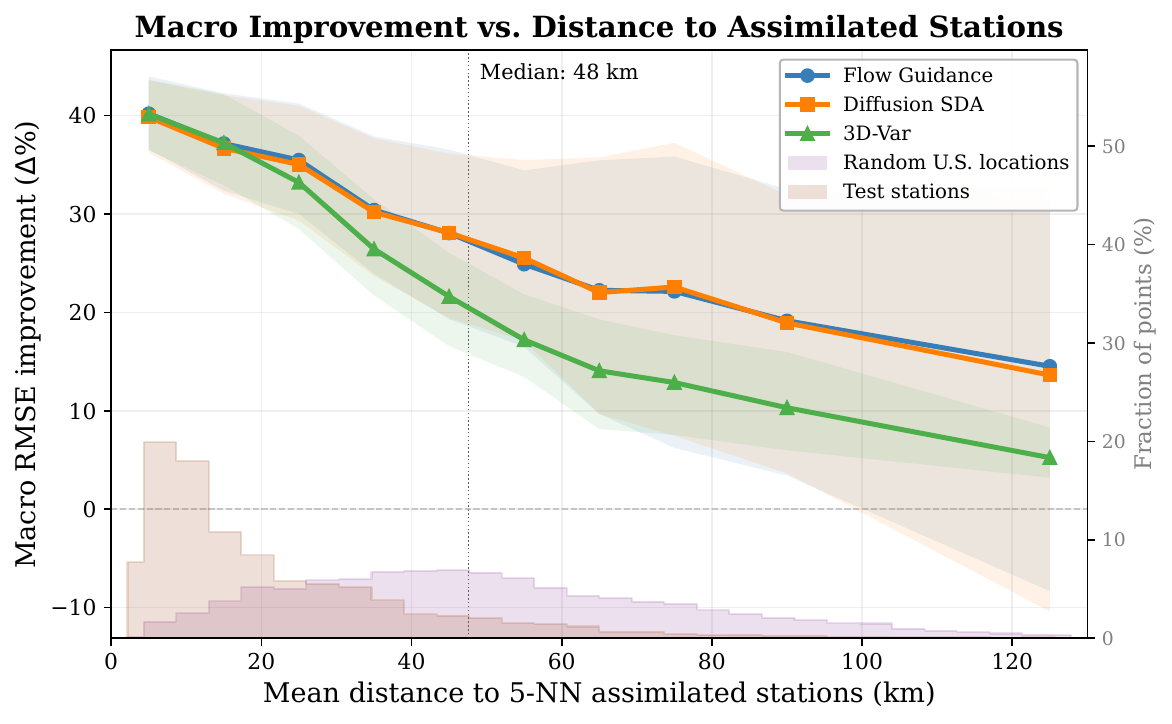}
    \caption{Dense benchmark ($11{,}849$ stations).}
    \label{fig:spatial_dense}
\end{subfigure}\hfill
\begin{subfigure}[t]{0.49\textwidth}
    \centering
    \includegraphics[width=\linewidth]{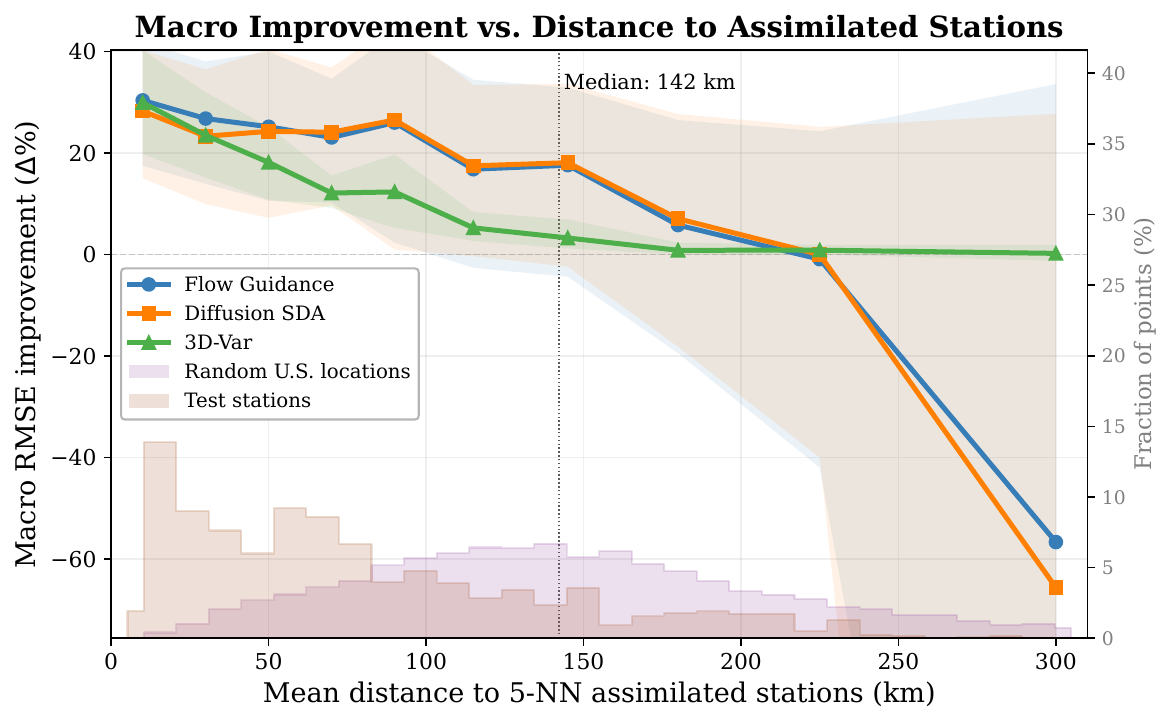}
    \caption{Sparse benchmark ($1{,}000$ stations).}
    \label{fig:spatial_sparse}
\end{subfigure}
\caption{Macro RMSE improvement ($\Delta\%$) over ERA5 for Flow Guidance, Diffusion-SDA, and 3D-Var, as a function of mean 5-NN distance to assimilated stations, for (a) the dense and (b) the sparse benchmark. Lines show the median across $360$ time steps; shaded bands span the 10th--90th percentile range. Vertical dashed lines mark the median 5-NN distance of a random CONUS location under each benchmark ($48$\,km dense; $142$\,km sparse). In the sparse benchmark, beyond ${\sim}150$\,km temperature and dewpoint improvements from the generative methods turn negative while wind improvements remain positive.}
\label{fig:spatial}
\end{figure}

The benchmark RMSEs aggregate across station locations and snapshots, hiding the spatial scale at which assimilation operates. We bin every test station by its mean distance to the five nearest assimilated stations (5-NN distance) and compute per-bin RMSE improvement. The 5-NN metric is used because the guidance signal draws on a neighborhood of stations. \autoref{fig:spatial_dense} shows the macro decay curve (the improvement averaged over the four variables) for the dense benchmark; \autoref{fig:spatial_sparse} repeats the analysis under the $1{,}000$-station sparse benchmark.

\paragraph{Improvement decays with distance from stations.}
Under dense benchmark, improvement is largest near the assimilated stations, about $40\%$ within $10$\,km for all three methods, and decays monotonically with distance (\autoref{fig:spatial_dense}). It remains positive out to ${\sim}120$\,km, four times the ERA5 grid spacing (${\sim}30$\,km), so assimilation corrects structures much larger than a single grid cell.

\paragraph{Generative methods retain more improvement away from stations.}
Under dense benchmark, within $10$\,km, all three methods perform comparably (${\sim}40\%$). At medium range, however, the generative methods appear to retain slightly more of their correction: by $50$--$60$\,km they still achieve ${\sim}25\%$, whereas 3D-Var reaches ${\sim}17\%$.

\paragraph{Wind corrections reach farther than temperature and dewpoint.}
Under dense benchmark, at $80$--$100$\,km, the generative methods still improve wind by more than $24\%$, but temperature by less than $15\%$ and dewpoint by less than $20\%$. (\autoref{fig:per_var_dense_si}, \autoref{app:per_var}).

\paragraph{Far from stations, the generative methods lack a fallback.}
Under sparse benchmark, the generative methods widen their medium-range margin over 3D-Var ($16$--$18\%$ vs.\ $3$--$5\%$ at $100$--$160$\,km; \autoref{fig:spatial_sparse}). Far from observations, however, their temperature and dewpoint improvements turn negative (\autoref{fig:per_var_sparse_si}), while 3D-Var stays near zero. The difference is the fallback. Where no station constrains the field, 3D-Var keeps the current ERA5 state, whereas the generative methods have no background field and fall back on the learned prior, which gives a plausible field but not the current local state.

\subsection{Regional Case Studies}
\label{sec:case_studies}

The analyses so far are statistical: the benchmark scores and the distance-decay curves average over all of CONUS and all $360$ snapshots. These averages show how much each method improves on ERA5, but not what the analysis fields look like or how the improvement varies from region to region. We therefore examine individual snapshots in two contrasting regions. In the Rocky Mountains, the ERA5 bias is largely a near-uniform regional offset, and all methods perform comparably. In the Great Lakes, open water leaves parts of the domain without stations, and the generative methods show a larger margin over 3D-Var at the test stations. Two further cases, the Los Angeles basin and the Florida peninsula, appear in \autoref{app:case_studies}.

\begin{figure}[htbp]
\centering
\includegraphics[width=\textwidth]{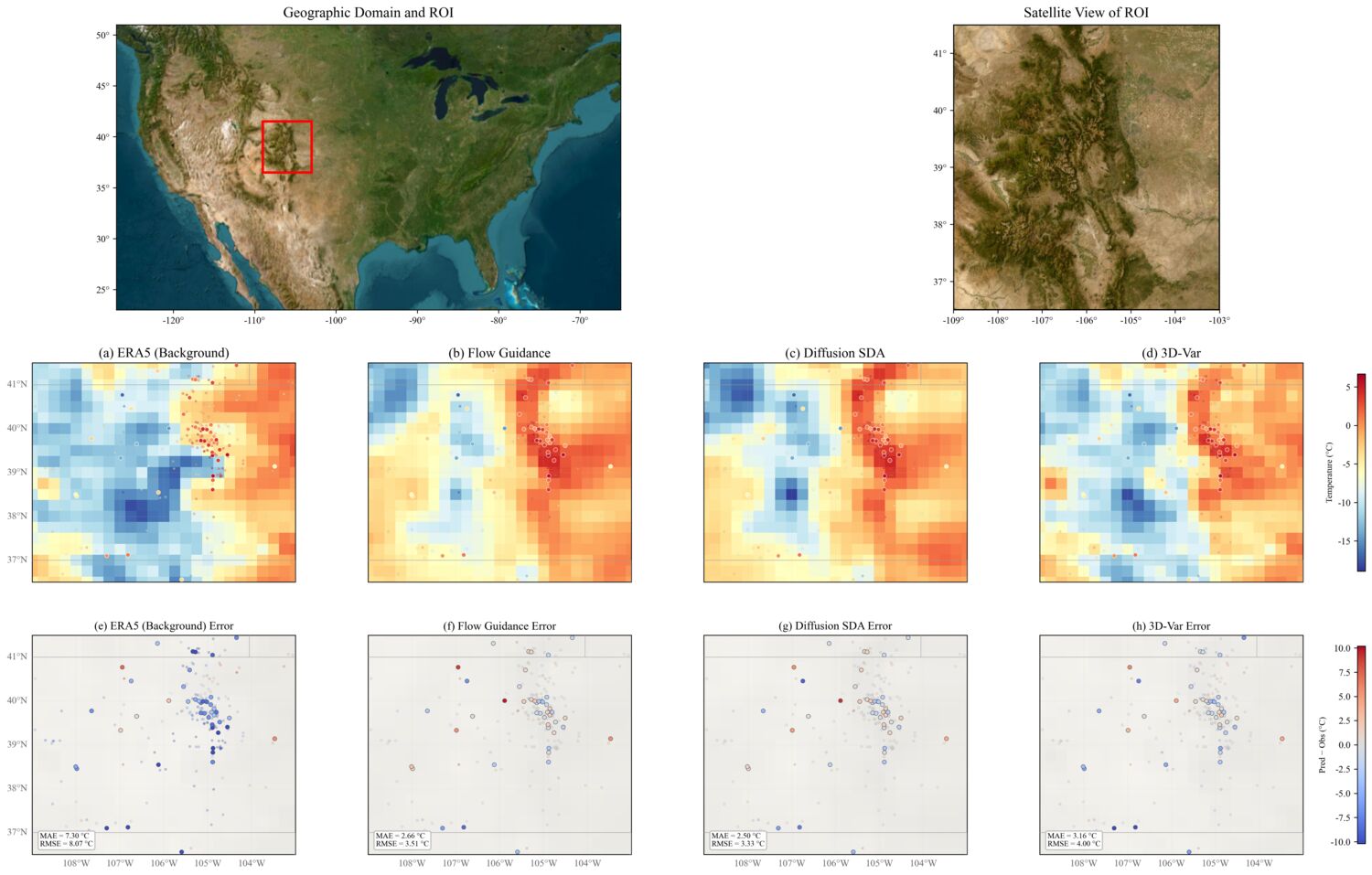}
\caption{Rocky Mountains case study for \ttwom. \textbf{Top row:} geographic context (left) and satellite view (right) of the region of interest. \textbf{Middle row:} (a) ERA5 background, (b) Flow Guidance, (c) Diffusion-SDA, (d) 3D-Var. \textbf{Bottom row:} per-station errors (analysis $-$ observation) for each method, with mean absolute error (MAE) and RMSE annotated. ERA5 shows a near-uniform ${\sim}6.5$\,K cold bias across the region; all three methods substantially reduce the error, with the generative methods (RMSE $3.3$--$3.5$\,K) modestly outperforming 3D-Var (RMSE $4.0$\,K).}
\label{fig:case_rockies}
\end{figure}

\paragraph{Rocky Mountains: a large cold bias corrected by all methods.}
Some of ERA5's largest \ttwom\ errors in our evaluation occur in the Rocky Mountains. \autoref{fig:case_rockies} shows a snapshot in which ERA5 is too cold at $91\%$ of test stations, with a near-uniform bias of $-6.5$\,K and an RMSE of $8.1$\,K. All three methods correct most of this error. Flow Guidance reduces the RMSE to $3.5$\,K, Diffusion-SDA to $3.3$\,K, and 3D-Var to $4.0$\,K.

\begin{figure}[htbp]
\centering
\includegraphics[width=\textwidth]{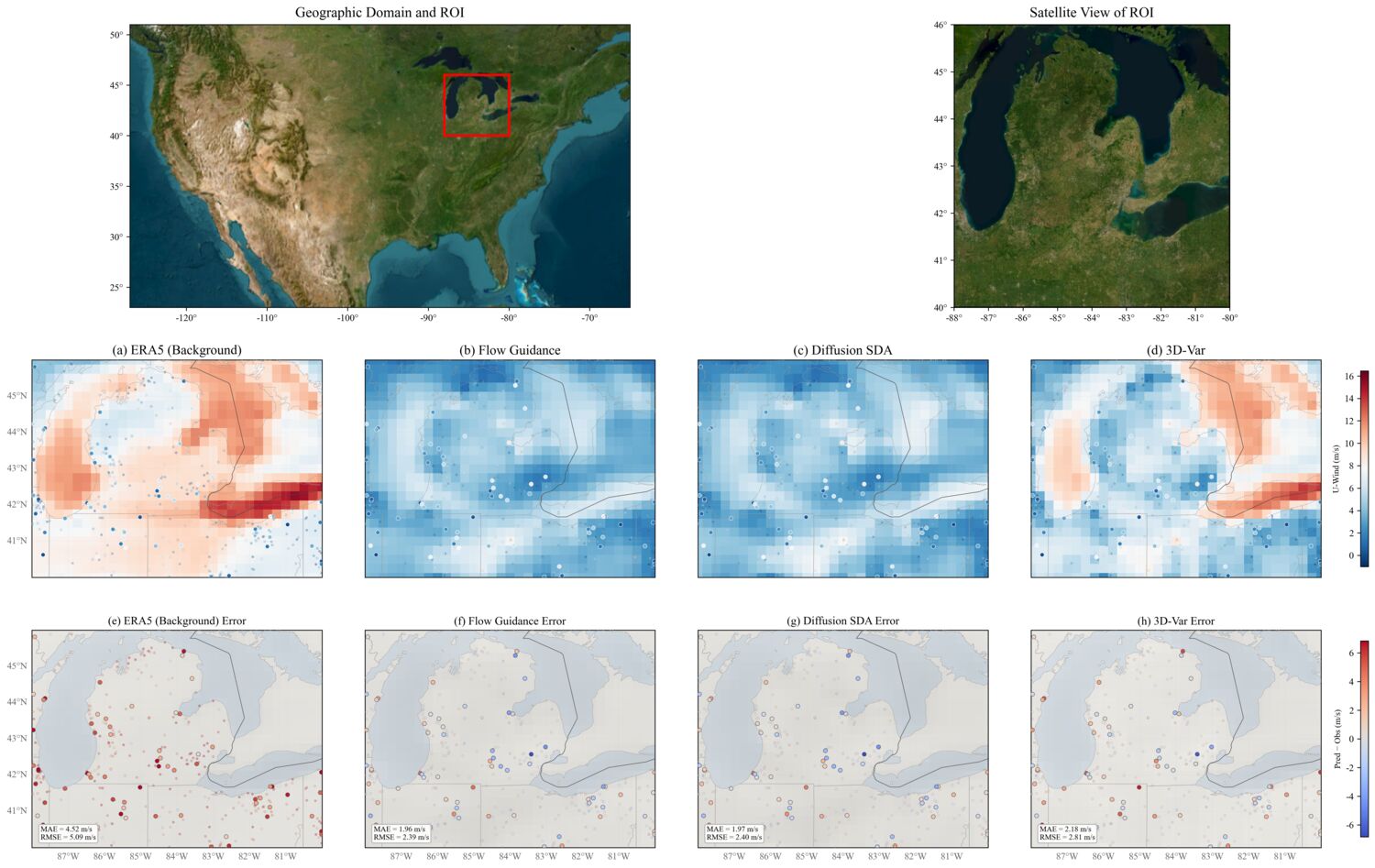}
\caption{Great Lakes case study for \uten. Panels as in \autoref{fig:case_rockies}. ERA5 overestimates wind speed by ${\sim}4.5$\,m/s across the region. Flow Guidance and Diffusion-SDA reduce the error at the shoreline test stations (RMSE 2.39 m/s, 53.0\% reduction), while 3D-Var retains larger residual errors (RMSE $2.81$\,m/s, $44.8\%$ reduction).}
\label{fig:case_lakes}
\end{figure}

\paragraph{Great Lakes: lake-effect winds and ungauged water surfaces.}
Much of the Great Lakes region is open water with no stations. \autoref{fig:case_lakes} shows a snapshot in which ERA5 overestimates the zonal wind \uten\ at the lakeshore stations: its test-station RMSE is $5.09$\,m/s, with a mean bias of ${+}4.5$\,m/s, and it is biased high at all 64 test stations. All three methods reduce this error. Flow Guidance and Diffusion-SDA both reach ${\sim}2.4$\,m/s, while 3D-Var reaches $2.81$\,m/s. Over the lakes, the generative methods appear to produce a spatially coherent correction, whereas 3D-Var's correction weakens away from the shore. We caution that the lake-interior fields cannot be directly verified, since no observations are available over the lakes themselves; the verifiable comparison is at the shoreline test stations.

\section{Discussion}
\label{sec:discussion}

\subsection{Learned Priors Provide More Effective Regularization}
\label{sec:disc_regularization}

Our results indicate that ERA5-trained generative priors provide more effective regularization than the ERA5-centered background in 3D-Var in our benchmarks. Regularization with either prior or background is essential because clustered, noisy station observations cannot fully constrain the gridded state. Without its background term, 3D-Var fits the training stations but fails to generalize (\autoref{sec:dense_benchmark}). Flow Guidance and Diffusion-SDA instead use learned priors and outperform 3D-Var on every variable in the dense benchmark, with larger gains under sparse benchmark (\autoref{tab:benchmark_main}, \autoref{tab:sparse_benchmark}). They achieve these gains despite starting from noise and receiving no ERA5 field at inference, whereas 3D-Var receives ERA5 as its background.

The spatial results suggest that learned priors extend the influence of observations farther from stations. All methods perform similarly near stations, but the generative methods retain more of their improvement as distance increases (\autoref{sec:spatial}). Learned priors may also allow more complete corrections of regional biases. The 3D-Var background term penalizes departures from ERA5, which can suppress corrections even when observations support them. A learned prior can accommodate coherent regional shifts that remain consistent with its training distribution. This flexibility may help explain the $6$--$8$\,pp advantage of the generative methods over 3D-Var in the Rocky Mountains case (\autoref{sec:case_studies}).

However, the generative methods lack an explicit background to fall back on where observations provide little information. Under sparse benchmark, their temperature and dewpoint estimates far from stations become less accurate than ERA5 (\autoref{sec:spatial}). Conditioning generation on ERA5 could provide this fallback while preserving the flexibility of the learned prior. Future work could introduce ERA5 as an additional input, a warm start, or a soft constraint.

\subsection{Full-Gradient Guidance is the best Conditioning Strategy}
\label{sec:disc_guidance}

Full-gradient guidance is the only conditioning strategy tested that outperforms 3D-Var on both dense and sparse benchmarks (\autoref{tab:benchmark_main}, \autoref{tab:sparse_benchmark}). We compare three strategies: full-gradient guidance (Flow Guidance and Diffusion-SDA), stop-gradient guidance (including FlowDPS), and initial-noise optimization (D-Flow; \autoref{tab:method_summary}). They differ in which sampling state they modify and how observation gradients reach that state (\autoref{sec:flow_da}).

Full-gradient guidance spreads observation corrections through spatial dependencies learned by the denoiser. It evaluates the observation loss on the denoised estimate and differentiates that loss with respect to the current sample (\autoref{eq:terminal_tweedie}). Through the denoiser Jacobian, a station residual can update grid locations beyond those directly connected to that station by the observation operator (\autoref{sec:guidance}). Observation information can therefore reach unobserved locations within each guidance step, providing a mechanism for the stronger generalization observed in our benchmarks.

Stop-gradient guidance removes the learned spatial coupling from each observation update. It replaces the denoiser Jacobian with the identity, so each station residual directly modifies only the grid cells connected to that station by the observation operator (\autoref{eq:stop_gradient}). The correction can reach other locations through subsequent sampling steps, but this indirect propagation appears insufficient. The benchmark results show a pattern of overfitting: stop-gradient guidance fits assimilated stations more closely but predicts test stations less accurately, especially under sparse coverage (\autoref{tab:ablation_sg}, \autoref{tab:sparse_benchmark}). FlowDPS uses the same approximation and repeats the observation updates on the terminal-state estimate. Increasing the number of updates further strengthens the training fit, while test performance first improves and then deteriorates, indicating that additional updates eventually exacerbate overfitting (\autoref{tab:ablation_sg}).

D-Flow optimizes the initial noise, so each observation gradient must pass through the entire sampling trajectory (\autoref{sec:dflow}). Its poor fit even at assimilated stations suggests difficulty with this optimization (\autoref{tab:benchmark_main}, \autoref{tab:sparse_benchmark}). It also requires roughly an order of magnitude more memory and time than Flow Guidance, making it less competitive in these experiments (\autoref{tab:benchmark_main}).

\subsection{Sparse Observations Amplify Differences in Generalization}
\label{sec:disc_sparsity}

Sparse coverage increases the risk of overfitting and exposes differences in how methods constrain unobserved regions. In a dense network, nearby stations jointly constrain the local field: a correction that fits one station's noise may worsen the fit at its neighbors, encouraging the assimilated field to follow their shared signal. Under sparse coverage, isolated stations become easier to fit, while predictions between stations depend more strongly on each method's ability to spread useful corrections. A tighter station fit can therefore coexist with poorer generalization, and differences between methods become more pronounced. Our results support this interpretation: under sparse observations, training-station improvements increase while train--test gaps widen, and stop-gradient guidance falls farther behind full-gradient guidance (\autoref{tab:benchmark_main}, \autoref{tab:sparse_benchmark}, \autoref{tab:ablation_sg}).

\subsection{Pixel-Space Flow Guidance Offers the Best Practical Trade-off}
\label{sec:disc_flow_vs_diffusion}

Flow Guidance matches Diffusion-SDA's accuracy with a simpler sampler (\autoref{sec:dense_benchmark}, \autoref{sec:sparse_benchmark}). With predictor-only sampling, flow matching performs better than diffusion (\autoref{tab:ablation_solver}, \autoref{tab:sparse_benchmark}), possibly because its linear training paths make the learned dynamics easier to integrate. Diffusion's deficit lies mainly in temperature and dewpoint (\autoref{tab:ablation_solver}). One possible explanation is weaker observation forcing for these variables. Their normalized ERA5 residuals are smaller than those of wind (${\sim}0.2$ versus ${\sim}0.5$; \autoref{tab:normalized_era5}), which may correspond to weaker observation gradients during sampling. The posterior corrector reinforces observation constraints through repeated updates at each noise level, combining them with the learned prior (\autoref{eq:guided_corrector}). The diffusion-derived score appears better suited to these corrective updates than the flow-derived corrector: its corrector supports larger stable Langevin steps and suppresses random perturbations more effectively than the flow-derived corrector, even on the same flow trajectory (\autoref{app:perturbation}, \autoref{fig:perturbation}). These additional updates from diffusion corrector substantially improve temperature and dewpoint at both assimilated and test stations (\autoref{tab:benchmark_main}, \autoref{tab:sparse_benchmark}). Diffusion therefore recovers much of its accuracy through corrective updates, whereas the default Flow Guidance sampler reaches comparable performance without them.

Latent compression can provide useful spatial constraints, but its value depends on the structure already learned by the assimilation model. In latent 3D-Var, the decoder introduces a learned spatial mapping for observation corrections, which may explain the gains under sparse coverage compared to 3D-Var, whose background covariance applies the same fixed spatial smoothing everywhere (\autoref{tab:sparse_benchmark}). Flow Guidance already captures spatial dependencies through its learned prior, so additional compression may offer less benefit while introducing reconstruction errors and restricting the available corrections (\autoref{sec:latent}, \autoref{app:autoencoder}). In both methods, spatial-only autoencoders remain competitive with those that also mix variables, suggesting limited additional benefit from explicit variable mixing (\autoref{tab:ae_mixing_3dvar}, \autoref{fig:latent}, \autoref{tab:sparse_benchmark}). Latent space therefore offers a practical trade-off: pixel space retains higher accuracy under dense coverage, while latent space matches pixel space accuracy at lower cost under sparse coverage (\autoref{sec:latent}, \autoref{sec:sparse_benchmark}).

Pixel-space Flow Guidance is therefore our practical default, with one tuning parameter and roughly half Diffusion-SDA's peak memory in our implementations (\autoref{sec:dense_benchmark}, \autoref{tab:benchmark_main}, \autoref{app:arch}). The ablations support full-gradient guidance with a constant schedule, single-step terminal extrapolation, and an adaptive solver without a corrector (\autoref{app:sg}, \autoref{app:schedule}, \autoref{app:terminal}, \autoref{app:solver}). A shallow spatial-only latent representation remains useful when memory is limited or observations are sparse (\autoref{sec:latent}, \autoref{sec:sparse_benchmark}, \autoref{fig:latent}).

\section{Conclusion}
\label{sec:conclusions}

This work establishes the first controlled benchmark of generative data assimilation for single-time near-surface analysis from real, noisy station observations. It covers $11{,}849$ MADIS stations across CONUS, four near-surface variables, and $360$ snapshots, and it holds the backbone architecture, dataset, and evaluation protocol fixed across flow matching, diffusion, and classical baselines. This evaluation of spatial generalization at held-out stations complements cycled assimilation experiments, which assess forecast impact. The benchmark supports three conclusions.

First, the best generative methods outperform 3D-Var ($35.7\%$ vs.\ $33.3\%$ RMSE improvement over ERA5) even though they use no ERA5 background field at inference. Their advantage widens from $2.4$\,pp on the dense benchmark to $7.9$\,pp on the sparse benchmark. Second, full-gradient velocity-space guidance consistently outperforms stop-gradient guidance and initial-noise optimization, especially under sparse observations. Third, the remaining design choices matter little. Diffusion and flow matching reach parity on both benchmarks, and neither latent-space assimilation nor inter-variable mixing in the autoencoder improves accuracy. Among these equally accurate options, Flow Guidance is the practical choice: it needs half the memory of Diffusion-SDA and only one hyperparameter to tune.

Several limitations remain. First, far from stations the generative prior has no fallback, whereas 3D-Var reverts to the ERA5 background. Second, the prior enforces no temporal consistency between successive snapshots \citep{hodyss2026using}. Third, it is confined to ERA5's distribution and grid, so it inherits ERA5's systematic biases and cannot represent sub-grid structure. Fourth, we assimilate only four near-surface variables from surface stations over CONUS, although our main comparisons hold for each of them and should extend to other variables. Fifth, our evaluation relies on deterministic RMSE and compares only against ERA5 and 3D-Var, not against operational regional analyses such as RTMA/URMA \citep{de2011real} or HRRR \citep{dowell2022high}.

Each limitation points to a direction for future work. A fallback far from stations could come from conditioning the generative prior on ERA5, as an additional input, a warm start, or a soft constraint. Temporal consistency could come from priors over $6$--$12$\,h windows, in the spirit of video diffusion \citep{ho2022video, chen2024diffusion} and AI weather forecasters \citep{lam2023learning, bodnar2025aurora}, or from coupling the framework with a forecast model for cycled assimilation. Sub-grid structure could come from coupling with downscaling conditioned on terrain, land cover, and high-resolution Earth-observation imagery \citep{giezendanner2026partial}. Other observation types could enter through modality-specific observation operators for radiosondes, aircraft, radar, GPS radio occultation, and satellites \citep{yang2025local}. Finally, a fuller evaluation would add ensemble-calibration metrics such as the CRPS \citep{gneiting2007strictly} and the spread--skill relationship \citep{price2024gencast}, which the generative framework supports at no extra cost, together with comparisons against RTMA/URMA and HRRR.

\ack{We thank ECMWF and the Copernicus Climate Change Service for producing and openly distributing the ERA5 reanalysis, and NOAA for maintaining the Meteorological Assimilation Data Ingest System (MADIS) archive of surface observations. Computational resources were provided by the Laboratory for Information \& Decision Systems (LIDS) at MIT.}

\funding{This work was supported in part by research funding to Sherrie Wang and in part by a Massachusetts Institute of Technology teaching assistantship awarded to Ruizhe Huang.}

\roles{%
\textbf{Ruizhe Huang}: Conceptualization, Data curation, Formal analysis, Funding acquisition, Investigation, Methodology.
\textbf{Qidong Yang}: Conceptualization, Data curation, Methodology.
\textbf{Jonathan Giezendanner}: Conceptualization, Data curation, Methodology.
\textbf{Sherrie Wang}: Conceptualization, Data curation, Formal analysis, Funding acquisition, Investigation, Methodology, Project administration.%
}

\data{ERA5 reanalysis is publicly available from the Copernicus Climate Data Store (\url{https://cds.climate.copernicus.eu}). MADIS surface observations are publicly available from NOAA (\url{https://madis.ncep.noaa.gov}). The curated ERA5--MADIS pairing used in this work is archived on Zenodo (\url{https://doi.org/10.5281/zenodo.18598860}) and described in Giezendanner et al.\ \citep{giezendanner2026era5madis}. The full benchmark are available through the project page \citep{huang2025website}.} Code, trained model weights, will be released upon acceptance.

\suppdata{Supplementary information appended below: sampler discretizations, design-space ablations, a perturbation analysis of the corrector, the latent-space study, autoencoder reconstruction metrics, additional regional case studies, and full benchmark tables.}

\bibliographystyle{iopart-num}
\bibliography{references}

@article{moldovan2026aifs,
  title={AIFS Single 1.1. 0: an update to ECMWF's machine-learned weather forecast model AIFS},
  author={Moldovan, Gabriel and Pinnington, Ewan and Prieto Nemesio, Ana and Lang, Simon and Ben Bouall{\`e}gue, Zied and Dramsch, Jesper and Alexe, Mihai and Santa Cruz, Mario and Hahner, Sara and Cook, Harrison and others},
  journal={Geoscientific Model Development},
  volume={19},
  number={10},
  pages={4703--4724},
  year={2026},
  publisher={Copernicus Publications G{\"o}ttingen, Germany}
}

@article{gneiting2007strictly,
  title={Strictly proper scoring rules, prediction, and estimation},
  author={Gneiting, Tilmann and Raftery, Adrian E},
  journal={Journal of the American statistical Association},
  volume={102},
  number={477},
  pages={359--378},
  year={2007},
  publisher={Taylor \& Francis}
}

@article{albergo2025stochastic,
  title={Stochastic interpolants: A unifying framework for flows and diffusions},
  author={Albergo, Michael and Boffi, Nicholas M and Vanden-Eijnden, Eric},
  journal={Journal of Machine Learning Research},
  volume={26},
  number={209},
  pages={1--80},
  year={2025}
}

@inproceedings{chung2023diffusion,
  title={Diffusion posterior sampling for general noisy inverse problems},
  author={Chung, Hyungjin and Kim, Jeongsol and Mccann, Michael T and Klasky, Marc L and Ye, Jong Chul},
  booktitle={International Conference on Learning Representations},
  year={2023}
}

@article{de2011real,
  title={The real-time mesoscale analysis at {NOAA}'s {National Centers for Environmental Prediction}: current status and development},
  author={De Pondeca, Manuel SFV and Manikin, Geoffrey S and DiMego, Geoff and Benjamin, Stanley G and Parrish, David F and Purser, R James and Wu, Wan-Shu and Horel, John D and Myrick, David T and Lin, Ying and others},
  journal={Weather and Forecasting},
  volume={26},
  number={5},
  pages={593--612},
  year={2011}
}

@article{dowell2022high,
  title={The High-Resolution Rapid Refresh (HRRR): An hourly updating convection-allowing forecast model. Part I: Motivation and system description},
  author={Dowell, David C and Alexander, Curtis R and James, Eric P and Weygandt, Stephen S and Benjamin, Stanley G and Manikin, Geoffrey S and Blake, Benjamin T and Brown, John M and Olson, Joseph B and Hu, Ming and others},
  journal={Weather and Forecasting},
  volume={37},
  number={8},
  pages={1371--1395},
  year={2022}
}

@article{price2024gencast,
  title={Probabilistic weather forecasting with machine learning},
  author={Price, Ilan and Sanchez-Gonzalez, Alvaro and Alet, Ferran and Andersson, Tom R and El-Kadi, Andrew and Masters, Dominic and Ewalds, Timo and Stott, Jacklynn and Mohamed, Shakir and Battaglia, Peter and others},
  journal={Nature},
  year={2024}
}

@article{rozet2024learning,
  title={Learning diffusion priors from observations by expectation maximization},
  author={Rozet, Fran{\c{c}}ois and Andry, G{\'e}r{\^o}me and Lanusse, Fran{\c{c}}ois and Louppe, Gilles},
  journal={Advances in Neural Information Processing Systems},
  volume={37},
  pages={87647--87682},
  year={2024}
}

@article{lam2023learning,
  title={Learning skillful medium-range global weather forecasting},
  author={Lam, Remi and Sanchez-Gonzalez, Alvaro and Willson, Matthew and Wirnsberger, Peter and Fortunato, Meire and Alet, Ferran and Ravuri, Suman and Ewalds, Timo and Eaton-Rosen, Zach and Hu, Weihua and others},
  journal={Science},
  volume={382},
  number={6677},
  pages={1416--1421},
  year={2023},
  publisher={American Association for the Advancement of Science}
}

@article{bodnar2025aurora,
  title={A foundation model for the Earth system},
  author={Bodnar, Cristian and Bruinsma, Wessel P and Lucic, Ana and Stanley, Megan and Brandstetter, Johannes and Garvan, Patrick and Riechert, Maik and Weyn, Jonathan and Dong, Haiyu and Vaughan, Anna and others},
  journal={Nature},
  year={2025}
}

@article{bi2023pangu,
  title={Accurate medium-range global weather forecasting with 3D neural networks},
  author={Bi, Kaifeng and Xie, Lingxi and Zhang, Hengheng and Chen, Xin and Gu, Xiaotao and Tian, Qi},
  journal={Nature},
  year={2023}
}

@article{pathak2022fourcastnet,
  title={Fourcastnet: A global data-driven high-resolution weather model using adaptive fourier neural operators},
  author={Pathak, Jaideep and Subramanian, Shashank and Harrington, Peter and Raja, Sanjeev and Chattopadhyay, Ashesh and Mardani, Morteza and Kurth, Thorsten and Hall, David and Li, Zongyi and Azizzadenesheli, Kamyar and others},
  journal={arXiv preprint arXiv:2202.11214},
  year={2022}
}

@inproceedings{lugmayr2022repaint,
  title={{RePaint}: Inpainting using denoising diffusion probabilistic models},
  author={Lugmayr, Andreas and Danelljan, Martin and Romero, Andres and Yu, Fisher and Timofte, Radu and Van Gool, Luc},
  booktitle={Proceedings of the IEEE/CVF Conference on Computer Vision and Pattern Recognition},
  pages={11461--11471},
  year={2022}
}

@inproceedings{lipman2022flow,
  title={Flow matching for generative modeling},
  author={Lipman, Yaron and Chen, Ricky T. Q. and Ben-Hamu, Heli and Nickel, Maximilian and Le, Matthew},
  booktitle={International Conference on Learning Representations (ICLR)},
  year={2023}
}

@inproceedings{liu2022flow,
  title={Flow straight and fast: Learning to generate and transfer data with rectified flow},
  author={Liu, Xingchao and Gong, Chengyue and Liu, Qiang},
  booktitle={International Conference on Learning Representations (ICLR)},
  year={2023}
}

@article{ben2024d,
  title={D-Flow: Differentiating through flows for controlled generation},
  author={Ben-Hamu, Heli and Puny, Omri and Gat, Itai and Karrer, Brian and Singer, Uriel and Lipman, Yaron},
  journal={arXiv preprint arXiv:2402.14017},
  year={2024}
}

@inproceedings{ho2020denoising,
  title={Denoising diffusion probabilistic models},
  author={Ho, Jonathan and Jain, Ajay and Abbeel, Pieter},
  booktitle={Advances in Neural Information Processing Systems},
  volume={33},
  pages={6840--6851},
  year={2020}
}

@inproceedings{ho2022video,
  title={Video diffusion models},
  author={Ho, Jonathan and Salimans, Tim and Gritsenko, Alexey and Chan, William and Norouzi, Mohammad and Fleet, David J.},
  booktitle={Advances in Neural Information Processing Systems},
  volume={35},
  pages={8633--8646},
  year={2022}
}

@inproceedings{song2020score,
  title={Score-based generative modeling through stochastic differential equations},
  author={Song, Yang and Sohl-Dickstein, Jascha and Kingma, Diederik P. and Kumar, Abhishek and Ermon, Stefano and Poole, Ben},
  booktitle={International Conference on Learning Representations (ICLR)},
  year={2021}
}

@article{hersbach2020era5,
  title={The {ERA5} global reanalysis},
  author={Hersbach, Hans and Bell, Bill and Berrisford, Paul and Hirahara, Shoji and Hor{\'a}nyi, Andr{\'a}s and Mu{\~n}oz-Sabater, Joaqu{\'\i}n and others},
  journal={Quarterly Journal of the Royal Meteorological Society},
  volume={146},
  number={730},
  pages={1999--2049},
  year={2020}
}

@inproceedings{rabier2003variational,
  title={Variational data assimilation: theory and overview},
  author={Rabier, Florence and Liu, Zhiquan},
  booktitle={Proc. ECMWF Seminar on Recent Developments in Data Assimilation for Atmosphere and Ocean, Reading, UK},
  pages={29--43},
  year={2003}
}

@article{dee2005bias,
  title={Bias and data assimilation},
  author={Dee, Dick P.},
  journal={Quarterly Journal of the Royal Meteorological Society},
  volume={131},
  number={613},
  pages={3323--3343},
  year={2005}
}

@article{dee2011era,
  title={The {ERA-Interim} reanalysis: Configuration and performance of the data assimilation system},
  author={Dee, Dick P. and Uppala, Sakari and Simmons, Adrian J. and Berrisford, Paul and Poli, Paul and Kobayashi, Shinya and others},
  journal={Quarterly Journal of the Royal Meteorological Society},
  volume={137},
  number={656},
  pages={553--597},
  year={2011}
}

@article{kalnay1996ncep,
  title={The {NCEP/NCAR} 40-Year Reanalysis Project},
  author={Kalnay, E and Kanamitsu, M and Kistler, R and Collins, W and Deaven, D and Gandin, L and Iredell, M and Saha, S and White, G and Woollen, J and others},
  journal={Bulletin of the American Meteorological Society},
  volume={77},
  number={3},
  pages={437--471},
  year={1996},
  publisher={American Meteorological Society}
}

@article{slivinski2019towards,
  title={Towards a more reliable historical reanalysis: Improvements for version 3 of the {T}wentieth {C}entury {R}eanalysis system},
  author={Slivinski, Laura C. and Compo, Gilbert P. and Whitaker, Jeffrey S. and Sardeshmukh, Prashant D. and Giese, Benjamin S. and McColl, Chesley and others},
  journal={Quarterly Journal of the Royal Meteorological Society},
  volume={145},
  number={724},
  pages={2876--2908},
  year={2019}
}

@article{grooms2022comparison,
  title={A comparison of nonlinear extensions to the ensemble {K}alman filter: {G}aussian anamorphosis and two-step ensemble filters},
  author={Grooms, Ian},
  journal={Computational Geosciences},
  volume={26},
  number={3},
  pages={633--650},
  year={2022}
}

@article{manshausen2025generative,
  title={Generative data assimilation of sparse weather station observations at kilometer scales},
  author={Manshausen, Peter and Cohen, Yair and Harrington, Peter and Pathak, Jaideep and Pritchard, Mike and Garg, Piyush and Mardani, Morteza and Kashinath, Karthik and Byrne, Simon and Brenowitz, Noah},
  journal={Journal of Advances in Modeling Earth Systems},
  volume={17},
  number={10},
  pages={e2024MS004505},
  year={2025},
  publisher={Wiley Online Library}
}

@article{andry2025appa,
  title={Appa: Bending Weather Dynamics with Latent Diffusion Models for Global Data Assimilation},
  author={Andry, Ga{\"e}tan and Rozet, Fran{\c{c}}ois and Lewin, Simon and Rochman, Omer and Mangeleer, Victor and Pirlet, Martin and others},
  journal={arXiv preprint arXiv:2504.18720},
  year={2025}
}

@inproceedings{qu2024deep,
  title={Deep generative data assimilation in multimodal setting},
  author={Qu, Yongquan and Nathaniel, Juan and Li, Shuolin and Gentine, Pierre},
  booktitle={Proceedings of the IEEE/CVF Conference on Computer Vision and Pattern Recognition},
  pages={449--459},
  year={2024}
}

@article{qu2025pnp,
  title={PnP-DA: Towards principled plug-and-play integration of variational data assimilation and generative models},
  author={Qu, Yongquan and Blanke, Matthieu and Shamekh, Sara and Gentine, Pierre},
  journal={arXiv preprint arXiv:2508.00325},
  year={2025}
}

@article{fan2026physically,
  title={Physically consistent global atmospheric data assimilation with machine learning in latent space},
  author={Fan, Hang and Bai, Lei and Fei, Bin and Xiao, Yi and Chen, Kang and Liu, Yong and others},
  journal={Science Advances},
  volume={12},
  number={1},
  pages={eaea4248},
  year={2026}
}

@article{rozet2023score,
  title={Score-based data assimilation},
  author={Rozet, Fran{\c{c}}ois and Louppe, Gilles},
  journal={Advances in Neural Information Processing Systems},
  volume={36},
  pages={40521--40541},
  year={2023}
}

@article{huang2024diffda,
  title={DiffDA: A diffusion model for weather-scale data assimilation},
  author={Huang, Langwen and Gianinazzi, Lukas and Yu, Yuejiang and Dueben, Peter D. and Hoefler, Torsten},
  journal={arXiv preprint arXiv:2401.05932},
  year={2024}
}

@article{hodyss2026using,
  title={Using diffusion models to do data assimilation},
  author={Hodyss, Daniel and Morzfeld, Matthias},
  journal={Monthly Weather Review},
  volume={154},
  number={2},
  pages={165--182},
  year={2026}
}

@inproceedings{chen2024diffusion,
  title={Diffusion forcing: Next-token prediction meets full-sequence diffusion},
  author={Chen, Boyuan and Mart{\'\i} Mons{\'o}, Daniel and Du, Yilun and Simchowitz, Max and Tedrake, Russ and Sitzmann, Vincent},
  booktitle={Advances in Neural Information Processing Systems},
  volume={37},
  pages={24081--24125},
  year={2024}
}

@article{pandey2025variational,
  title={Variational control for guidance in diffusion models},
  author={Pandey, Kushagra and Sofian, Faris M. and Draxler, Felix and Karaletsos, Theofanis and Mandt, Stephan},
  journal={arXiv preprint arXiv:2502.03686},
  year={2025}
}

@inproceedings{shysheya2024conditional,
  title={On conditional diffusion models for {PDE} simulations},
  author={Shysheya, Aliaksandra and Diaconu, Cristian and Bergamin, Federico and Perdikaris, Paris and Hern{\'a}ndez-Lobato, Jos{\'e} Miguel and Turner, Richard and Mathieu, Emile},
  booktitle={Advances in Neural Information Processing Systems},
  volume={37},
  pages={23246--23300},
  year={2024}
}

@article{srivastava2025control,
  title={Control-augmented autoregressive diffusion for data assimilation},
  author={Srivastava, Pratyush and Sofian, Faris M. and Immorlano, Francesco and Pandey, Kushagra and Mandt, Stephan},
  journal={arXiv preprint arXiv:2510.06637},
  year={2025}
}

@article{wilczak2024evaluation,
  title={Evaluation and bias correction of the {ERA5} reanalysis over the {U}nited {S}tates for wind and solar energy applications},
  author={Wilczak, James M. and Akish, Evelyn and Capotondi, Antonietta and Compo, Gilbert P.},
  journal={Energies},
  volume={17},
  number={7},
  pages={1667},
  year={2024}
}

@article{chen2024evaluation,
  title={Evaluation of {ERA5} precipitation and 10-m wind speed associated with extratropical cyclones using station data over {N}orth {A}merica},
  author={Chen, Ting-Chen and Collet, Fabrice and Di Luca, Alejandro},
  journal={International Journal of Climatology},
  volume={44},
  number={3},
  pages={729--747},
  year={2024}
}

@misc{noaa2025madis,
  title={Meteorological Assimilation Data Ingest System ({MADIS})},
  author={{NOAA/NCEP}},
  year={2025},
  howpublished={\url{https://madis.ncep.noaa.gov/}},
  note={Accessed: 2025-08-18}
}

@article{yang2025local,
  title={Local off-grid weather forecasting with multi-modal earth observation data},
  author={Yang, Qidong and Giezendanner, Jonathan and Civitarese, Daniel Salles and Jakubik, Johannes and Schmitt, Eric and Chandra, Anirban and Vila, Jeremy and Hohl, Detlef and Hill, Chris and Watson, Campbell and others},
  journal={Journal of Advances in Modeling Earth Systems},
  volume={17},
  number={12},
  pages={e2025MS005207},
  year={2025},
  publisher={Wiley Online Library}
}

@article{giezendanner2026partial,
  title={Partial recovery of meter-scale surface weather},
  author={Giezendanner, Jonathan and Yang, Qidong and Schmitt, Eric and Chandra, Anirban and Civitarese, Daniel Salles and Jakubik, Johannes and Vila, Jeremy and Hohl, Detlef and Watson, Campbell and Wang, Sherrie},
  journal={arXiv preprint arXiv:2602.23146},
  year={2026}
}

@misc{giezendanner2026era5madis,
  title={{ERA5} and {MADIS} (ground stations) curated weather data for {CONUS} (2020--2023)},
  author={Giezendanner, Jonathan and Yang, Qidong and Huang, Ruizhe and Wang, Sherrie},
  year={2026},
  publisher={Zenodo},
  doi={10.5281/zenodo.18598860},
  url={https://doi.org/10.5281/zenodo.18598860}
}

@article{van2019particle,
  title={Particle filters for high-dimensional geoscience applications: A review},
  author={Van Leeuwen, Peter Jan and K{\"u}nsch, Hans R and Nerger, Lars and Potthast, Roland and Reich, Sebastian},
  journal={Quarterly Journal of the Royal Meteorological Society},
  volume={145},
  pages={2335--2365},
  year={2019}
}

@article{sun2025sda,
  title={LO-SDA: Latent Optimization for Score-based Atmospheric Data Assimilation},
  author={Sun, Jing-An and Fan, Hang and Gong, Junchao and Fei, Ben and Chen, Kun and Ling, Fenghua and Zhang, Wenlong and Xu, Wanghan and Yan, Li and Gentine, Pierre and others},
  journal={arXiv preprint arXiv:2510.22562},
  year={2025}
}

@article{karras2022elucidating,
  title={Elucidating the design space of diffusion-based generative models},
  author={Karras, Tero and Aittala, Miika and Aila, Timo and Laine, Samuli},
  journal={Advances in neural information processing systems},
  volume={35},
  pages={26565--26577},
  year={2022}
}

@article{vincent2011connection,
  title={A connection between score matching and denoising autoencoders},
  author={Vincent, Pascal},
  journal={Neural computation},
  volume={23},
  number={7},
  pages={1661--1674},
  year={2011},
  publisher={MIT Press}
}

@article{parrish1992national,
  title={The National Meteorological Center's spectral statistical-interpolation analysis system},
  author={Parrish, David F and Derber, John C},
  journal={Monthly Weather Review},
  volume={120},
  number={8},
  pages={1747--1763},
  year={1992}
}

@article{weaver2001correlation,
  title={Correlation modelling on the sphere using a generalized diffusion equation},
  author={Weaver, Anthony and Courtier, Philippe},
  journal={Quarterly Journal of the Royal Meteorological Society},
  volume={127},
  number={575},
  pages={1815--1846},
  year={2001},
  publisher={Wiley Online Library}
}

@article{lindgren2011explicit,
  title={An explicit link between Gaussian fields and Gaussian Markov random fields: the stochastic partial differential equation approach},
  author={Lindgren, Finn and Rue, H{\aa}vard and Lindstr{\"o}m, Johan},
  journal={Journal of the Royal Statistical Society Series B: Statistical Methodology},
  volume={73},
  number={4},
  pages={423--498},
  year={2011},
  publisher={Oxford University Press}
}

@article{zhang2022fast,
  title={Fast sampling of diffusion models with exponential integrator},
  author={Zhang, Qinsheng and Chen, Yongxin},
  journal={arXiv preprint arXiv:2204.13902},
  year={2022}
}

@article{dormand1980family,
  title={A family of embedded Runge-Kutta formulae},
  author={Dormand, John R and Prince, Peter J},
  journal={Journal of computational and applied mathematics},
  volume={6},
  number={1},
  pages={19--26},
  year={1980},
  publisher={Elsevier}
}

@book{liu2024rectifiedflow,
  author    = {Qiang Liu},
  title     = {Let us Flow Together},
  year      = {2024},
  url       = {https://www.cs.utexas.edu/~lqiang/PDF/flow_book.pdf},
  note      = {Chapter 5: Flow to Diffusion}
}

@article{cheng2026flowda,
  title={FlowDA: Accurate, Low-Latency Weather Data Assimilation via Flow Matching},
  author={Cheng, Ran and Zhu, Lailai},
  journal={arXiv preprint arXiv:2602.06800},
  year={2026}
}

@inproceedings{kim2025flowdps,
  title={Flowdps: Flow-driven posterior sampling for inverse problems},
  author={Kim, Jeongsol and Kim, Bryan Sangwoo and Ye, Jong Chul},
  booktitle={Proceedings of the IEEE/CVF International Conference on Computer Vision},
  pages={12328--12337},
  year={2025}
}

@article{wang2025four,
  title={A four-dimensional variational informed generative adversarial network for data assimilation},
  author={Wang, Wuxin and Duan, Boheng and Ni, Weicheng and Lu, Jingze and Yuan, Taikang and Li, Dawei and Zhao, Juan and Ren, Kaijun},
  journal={Journal of Advances in Modeling Earth Systems},
  volume={17},
  number={6},
  pages={e2024MS004437},
  year={2025},
  publisher={Wiley Online Library}
}

@article{wang2024accurate,
  title={Accurate initial field estimation for weather forecasting with a variational constrained neural network},
  author={Wang, Wuxin and Zhang, Jinrong and Su, Qingguo and Chai, Xingyu and Lu, Jingze and Ni, Weicheng and Duan, Boheng and Ren, Kaijun},
  journal={npj Climate and Atmospheric Science},
  volume={7},
  number={1},
  pages={223},
  year={2024},
  publisher={Nature Publishing Group UK London}
}

@article{peyron2021latent,
  title={Latent space data assimilation by using deep learning},
  author={Peyron, Mathis and Fillion, Anthony and G{\"u}rol, Selime and Marchais, Victor and Gratton, Serge and Boudier, Pierre and Goret, Gael},
  journal={Quarterly Journal of the Royal Meteorological Society},
  volume={147},
  number={740},
  pages={3759--3777},
  year={2021},
  publisher={Wiley Online Library}
}

@inproceedings{song2020denoising,
  title={Denoising diffusion implicit models},
  author={Song, Jiaming and Meng, Chenlin and Ermon, Stefano},
  booktitle={International Conference on Learning Representations (ICLR)},
  year={2021}
}

@article{wang2024dabench,
  title={Dabench: A benchmark dataset for data-driven weather data assimilation},
  author={Wang, Wuxin and Ni, Weicheng and Han, Tao and Bai, Lei and Duan, Boheng and Ren, Kaijun},
  journal={arXiv preprint arXiv:2408.11438},
  year={2024}
}

@article{wang2025dambench,
  title={{DAMBench}: A Multi-Modal Benchmark for Deep Learning-based Atmospheric Data Assimilation},
  author={Wang, Hao and Weng, Zixuan and Han, Jindong and Fan, Wei and Liu, Hao},
  journal={arXiv preprint arXiv:2511.01468},
  year={2025}
}

@article{xu2025fuxi,
  title={Fuxi-da: A generalized deep learning data assimilation framework for assimilating satellite observations},
  author={Xu, Xiaoze and Sun, Xiuyu and Han, Wei and Zhong, Xiaohui and Chen, Lei and Gao, Zhiqiu and Li, Hao},
  journal={npj Climate and Atmospheric Science},
  volume={8},
  number={1},
  pages={156},
  year={2025},
  publisher={Nature Publishing Group UK London}
}

@article{yang2025generative,
  title={Generative assimilation and prediction for weather and climate},
  author={Yang, Shangshang and Nai, Congyi and Liu, Xinyan and Li, Weidong and Chao, Jie and Wang, Jingnan and Wang, Leyi and Li, Xichen and Chen, Xi and Lu, Bo and others},
  journal={arXiv preprint arXiv:2503.03038},
  year={2025}
}

@article{meng2022diffusion,
  title={Diffusion model based posterior sampling for noisy linear inverse problems},
  author={Meng, Xiangming and Kabashima, Yoshiyuki},
  journal={arXiv preprint arXiv:2211.12343},
  year={2022}
}

@misc{huang2025website,
  title        = {Benchmarking Generative Models for Weather Data Assimilation on Real Station Observations: Project Page},
  author       = {Huang, Ruizhe and Yang, Qidong and Giezendanner, Jonathan and Wang, Sherrie},
  year         = {2025},
  howpublished = {\url{https://earth-intelligence-lab.github.io/generative-weather-da/}},
  note         = {Accessed 2026-04-23}
}

\appendix

\section{Architecture and training details}
\label{app:arch}

The shared U-Net backbone has $\sim$3.8\,M parameters with channel widths $[16, 32, 64, 128]$, four residual blocks per scale, group-norm activations, and time conditioning by 256-dimensional Fourier features. Training uses Adam with learning rate $2\times 10^{-4}$, effective batch size 64 across two NVIDIA RTX\,8000 GPUs (PyTorch DDP), exponential moving averaging with halflife 500\,kimg, and 200 equivalent epochs on z-scored ERA5 fields (training years 2017--2022, validation 2016, held-out test 2023). The latent-space variant is trained in two stages: the variational autoencoder is fit first on the same ERA5 split with an L1 reconstruction loss plus a KL regularizer ($\beta = 10^{-4}$, $S = 2$ scales for the production latent runs, base width 16, latent channel dimension 16, group parameter $G = 4$ for spatial-mixing-only or $G = 1$ for full mixing); the latent flow model is then trained with the same backbone and protocol. All assimilation runs use a single RTX\,8000 GPU. Per-method assimilation hyperparameters are: Diffusion-SDA $\sigma_o = 0.1$, $\gamma = 0.5$, stability parameter $\eta = 0.0316$, $K = 2$ corrector steps, $N = 128$ predictor steps, Langevin step size $\tau = 0.1$; Flow Guidance constant $\lambda = 2\times 10^{5}$ (dense) or $5\times 10^{4}$ (sparse), Dormand--Prince tolerances $\texttt{atol} = \texttt{rtol} = 10^{-5}$, $M = 16$ ensemble members, full gradient through $\mathbf{v}_\theta$; FlowDPS $K = 20$ inner iterations with step size $\zeta = 5{,}000$; D-Flow L-BFGS over the initial noise through the frozen ODE with regularizer weight $\lambda_{\text{init}} = 0.1$.

\FloatBarrier
\subsection{Interpolation scheme comparison}
\label{app:interpolation}

\autoref{tab:interpolation} compares nearest-neighbor, bilinear, and bicubic interpolation for the ERA5 baseline and for Flow Guidance assimilation on validation stations. All three yield similar ERA5 baseline RMSE, with differences within $0.06$ on average. We adopt bicubic as the observation operator $\mathcal{H}$ throughout this work (\autoref{sec:evaluation}): although the ERA5 baseline is marginally better under bilinear (average validation RMSE $2.21$ vs.\ $2.22$), bicubic yields slightly better assimilation performance for Flow Guidance on validation stations (RMSE $1.43$ vs.\ $1.44$). The same operator is used for the ERA5 baseline evaluation and for all assimilation methods (Flow Guidance, D-Flow, Diffusion-SDA, 3D-Var, and Latent 3D-Var), ensuring a consistent mapping from gridded fields to station locations across all experiments.

\begin{table}[htbp]
\centering
\small
\caption{Effect of interpolation scheme on ERA5 baseline and Flow Guidance assimilation RMSE (mean $\pm$ std, validation stations). Bicubic is selected based on validation performance. Bold marks the best value per column within each panel.}
\label{tab:interpolation}
\begin{tabular}{l ccccc}
\toprule
\textbf{Interpolation} & \textbf{\uten} & \textbf{\vten} & \textbf{\ttwom} & \textbf{\dtwom} & \textbf{Average} \\
\midrule
\multicolumn{6}{l}{\textit{ERA5 Baseline}} \\
Nearest  & 1.90$\pm$0.31 & 1.95$\pm$0.28 & 2.51$\pm$0.31 & 2.72$\pm$0.51 & 2.27$\pm$0.35 \\
Bilinear & \textbf{1.88}$\pm$0.31 & \textbf{1.93}$\pm$0.28 & \textbf{2.43}$\pm$0.30 & \textbf{2.60}$\pm$0.45 & \textbf{2.21}$\pm$0.34 \\
Bicubic  & 1.89$\pm$0.31 & 1.94$\pm$0.28 & 2.44$\pm$0.31 & 2.63$\pm$0.47 & 2.22$\pm$0.34 \\
\midrule
\multicolumn{6}{l}{\textit{Flow Guidance}} \\
Bilinear & \textbf{1.15}$\pm$0.17 & \textbf{1.18}$\pm$0.18 & 1.75$\pm$0.18 & 1.70$\pm$0.21 & 1.44$\pm$0.18 \\
Bicubic  & \textbf{1.15}$\pm$0.17 & \textbf{1.18}$\pm$0.18 & \textbf{1.73}$\pm$0.17 & \textbf{1.69}$\pm$0.21 & \textbf{1.43}$\pm$0.18 \\
\bottomrule
\end{tabular}
\end{table}

\subsection{Normalized-space ERA5 baseline}
\label{app:normalized}

The wind versus temperature and dewpoint asymmetry that recurs throughout the main text traces to a $\sim$$2.4\times$ ratio of normalized ERA5 baseline error: after z-score normalization, wind variables retain a larger residual, so any guidance correction has more signal to work with on wind and proportionally less on the other two.

\begin{table}[htbp]
\centering
\caption{ERA5 baseline error in physical and normalized units.}
\label{tab:normalized_era5}
\small
\begin{tabular}{l cccc}
\toprule
& \uten & \vten & \ttwom & \dtwom \\
\midrule
Training std (physical units, m\,s$^{-1}$ or K) & 3.45 & 3.84 & 11.70 & 11.67 \\
ERA5 validation RMSE (physical)                 & 1.89 & 1.94 & 2.44  & 2.63 \\
ERA5 validation RMSE (normalized)               & 0.547 & 0.504 & 0.209 & 0.225 \\
\bottomrule
\end{tabular}
\end{table}

\FloatBarrier
\section{Generative priors and samplers}
\label{app:edm}
\label{sec:diffusion}

\paragraph{Terminology and conventions.}
All generative DA methods produce $M$ ensemble members from independent noise draws and return the ensemble mean as the final analysis field ($M = 16$ for Diffusion-SDA, Flow Guidance, Latent Flow Guidance, and FlowDPS; $M = 1$ for D-Flow due to its high computational cost). We use $\mathcal{D}$ (calligraphic) for the autoencoder decoder and $D_\theta$ (italic, with parameter subscript) for the diffusion denoiser; the two are distinct objects, disambiguated by context. In the classical setting (\autoref{sec:3dvar}), \emph{prior} refers to the background distribution $\mathcal{N}(\mathbf{x}_b, \mathbf{B})$ centered on a forecast field; in the generative setting, \emph{(learned) prior} refers to the trained distribution $p_\theta(\mathbf{x})$, decoupled from any specific observation or background field, whose sampling begins from the base noise distribution $\mathcal{N}(\mathbf{0}, \mathbf{I})$. \emph{Posterior} always denotes the Bayesian conditional $p(\mathbf{x} \mid \mathbf{y}^*)$ over the atmospheric state given observations.

We present diffusion models in the EDM (Elucidating the Design Space of Diffusion Models) framework \citep{karras2022elucidating}, which provides a unified parameterization that clarifies the design choices underlying score-based diffusion models, and then obtain flow matching as a special case.

\subsection{Forward process and training}

The forward process (\autoref{eq:edm_forward}) is parameterized by the scaling function $s(\sigma)$. Setting $s(\sigma) = 1$ recovers the Variance Exploding (VE) formulation of \citep{karras2022elucidating}, while $s(\sigma) = 1/\sqrt{\sigma^2 + 1}$ yields the Variance Preserving (VP) formulation \citep{song2020score,manshausen2025generative}, under which $\mathbf{x}$ has unit variance at all noise levels.

Internally, the denoiser uses $\sigma$-dependent preconditioning to normalize inputs and outputs across noise levels \citep{karras2022elucidating}. The training loss is:
\begin{equation}
    \mathcal{L}_{\text{EDM}}(\theta) = \mathbb{E}_{\sigma,\, \mathbf{x}_1,\, \boldsymbol{\epsilon}} \left[ w(\sigma)\, \| D_\theta(\mathbf{x}/s;\, \sigma) - \mathbf{x}_1 \|_2^2 \right],
    \label{eq:edm_loss}
\end{equation}
where $\sigma$ is sampled from $\ln \sigma \sim \mathcal{N}(P_{\text{mean}},\, P_{\text{std}}^2)$ and $w(\sigma) = (\sigma^2 + \sigma_{\text{data}}^2)/(\sigma\,\sigma_{\text{data}})^2$ is chosen so that each noise level contributes equally to the training gradient \citep{karras2022elucidating}. A key insight from denoising score matching \citep{vincent2011connection} is that regressing onto $\mathbf{x}_1$ implicitly recovers the marginal score needed for sampling (\autoref{eq:score_denoiser}). Since the denoiser sees $\mathbf{x}/s$ and depends only on $\sigma$, it is agnostic to the choice of $s(\sigma)$, allowing a single model to be deployed with any scaling without retraining. In this work, we train with VE scaling $s = 1$ and sample with VP scaling $s(\sigma) = 1/\sqrt{\sigma^2 + 1}$.

\subsection{Score and sampling}

The trained denoiser implicitly defines a score function:
\begin{equation}
    \nabla_{\mathbf{x}} \log p_\sigma(\mathbf{x}) = \frac{s\,D_\theta(\mathbf{x}/s;\,\sigma) - \mathbf{x}}{s^2\sigma^2},
    \label{eq:score_denoiser}
\end{equation}
obtained by applying the change of variables $\mathbf{x} = s(\mathbf{x}_1 + \sigma\boldsymbol{\epsilon})$ to the denoising score matching identity \citep{vincent2011connection}. This score enables both a reverse-time stochastic differential equation (SDE) for stochastic sampling \citep{song2020score} and a probability flow ODE that shares the same marginal distributions for deterministic sampling. In this work, we only use the probability flow ODE:
\begin{equation}
    \frac{d\mathbf{x}}{d\sigma} = \frac{s^{\prime}}{s} \mathbf{x}-s^{2} \sigma \nabla_{\mathbf{x}} \log p_{\sigma}(\mathbf{x}) = \frac{s'}{s}\,\mathbf{x} + \frac{\mathbf{x} - s\,D_\theta(\mathbf{x}/s;\,\sigma)}{\sigma},
    \label{eq:edm_ode_sigma}
\end{equation}
where $s' = ds/d\sigma$. To align with the flow-matching convention (\autoref{sec:flow_matching}), we reparameterize with $t \in [0,1]$ via a monotonically decreasing noise schedule $\sigma(t)$ with $\sigma(0) = \sigma_{\max}$ and $\sigma(1) \approx 0$. Denoting the ODE trajectory as $\mathbf{x}_t$:
\begin{equation}
    \frac{d\mathbf{x}_t}{dt} = \underbrace{\frac{\dot{s}}{s}\,\mathbf{x} + \frac{\dot{\sigma}}{\sigma}\bigl(\mathbf{x} - s\,D_\theta(\mathbf{x}/s;\,\sigma)\bigr)}_{\displaystyle \mathbf{v}_\theta(\mathbf{x}_t,\, t)},
    \label{eq:edm_ode_t}
\end{equation}
where $\dot{s} = ds/dt$ and $\dot{\sigma} = d\sigma/dt < 0$ denote derivatives with respect to $t$. This ODE is integrated from $\mathbf{x}_0 \sim \mathcal{N}\bigl(\mathbf{0},\, s(0)^2\,\sigma(0)^2\,\mathbf{I}\bigr)$ at $t=0$ to $\mathbf{x}_1$ at $t=1$, matching the structure of the flow-matching ODE (\autoref{eq:fm_ode}).

\paragraph{Predictor-Corrector (PC) sampler.}
In practice we integrate this ODE with a predictor-corrector (PC) sampler \citep{manshausen2025generative}: an exponential-integrator predictor \citep{zhang2022fast} followed by $K$ Langevin corrector steps that use the score (\autoref{eq:score_denoiser}) to project samples toward the support of $p_\sigma$, at a total cost of $N(1+K)$ network evaluations. The cosine noise schedule and the discretized predictor and corrector updates are given in \autoref{app:samplers}.

\subsection{Flow matching as a special case}
\label{sec:flow_matching}

Flow matching \citep{lipman2022flow,liu2022flow,albergo2025stochastic} learns a velocity field that transports Gaussian noise to data via an ODE. We show that it is a special case of the EDM framework with a specific choice of scaling and noise schedule. Given clean data $\mathbf{x}_1 \sim p_{\text{data}}$ and noise $\mathbf{x}_0 \sim \mathcal{N}(\mathbf{0}, \mathbf{I})$, flow matching defines a linear interpolation path:
\begin{equation}
    \mathbf{x} = (1-t)\,\mathbf{x}_0 + t\,\mathbf{x}_1,
    \quad t \in [0, 1],
    \label{eq:fm_interpolant}
\end{equation}
which is the EDM forward process (\autoref{eq:edm_forward}) with $s(t) = t$ and $\sigma(t) = (1-t)/t$. A network $\mathbf{v}_\theta(\mathbf{x}, t)$ is trained to predict the conditional velocity $\mathbf{u}_t(\mathbf{x} \mid \mathbf{x}_0, \mathbf{x}_1) = \mathbf{x}_1 - \mathbf{x}_0$:
\begin{equation}
    \mathcal{L}_{\text{FM}}(\theta) = \mathbb{E}_{t,\,\mathbf{x}_0, \,\mathbf{x}_1}\left[\|\mathbf{v}_\theta(\mathbf{x}, t) - (\mathbf{x}_1 - \mathbf{x}_0)\|_2^2\right].
    \label{eq:flow_matching_loss}
\end{equation}

\paragraph{Relation to the EDM denoiser.}
Substituting $s = t$, $\sigma = (1{-}t)/t$ into the probability flow ODE (\autoref{eq:edm_ode_t}) and comparing with $d\mathbf{x}/dt = \mathbf{v}_\theta$ yields the velocity--denoiser relation:
\begin{align}
    \mathbf{v}_\theta(\mathbf{x}, t) &= \frac{D_\theta\bigl(\mathbf{x}/t;\,(1{-}t)/t\bigr) - \mathbf{x}}{1-t},
    \label{eq:velocity_denoiser} \\
    \underbrace{\mathbf{x} + (1-t)\,\mathbf{v}_\theta(\mathbf{x},t)}_{\hat{\mathbf{x}}_1\;\text{(terminal extrapolation)}} &= D_\theta\bigl(\mathbf{x}/t;\,(1{-}t)/t\bigr).
    \label{eq:terminal_tweedie}
\end{align}
Substituting this into the flow-matching loss shows that its velocity-regression form is equivalent to a denoiser-regression form with a time-dependent reweighting:
\begin{equation}
    \mathcal{L}_{\text{FM}} = \mathbb{E}_{t,\,\mathbf{x}_0, \,\mathbf{x}_1}\!\left[\frac{1}{(1-t)^2}\,\|D_\theta\bigl(\mathbf{x}/t;\,(1{-}t)/t\bigr) - \mathbf{x}_1\|_2^2\right].
    \label{eq:fm_loss_denoiser}
\end{equation}

\paragraph{Score function.}
Substituting $s = t$, $\sigma = (1{-}t)/t$ into \autoref{eq:score_denoiser}:
\begin{equation}
    \nabla_{\mathbf{x}} \log p_t(\mathbf{x})
    = \frac{t\,D_\theta - \mathbf{x}}{(1-t)^2}
    = \frac{t\,\mathbf{v}_\theta - \mathbf{x}}{1-t}.
    \label{eq:fm_score}
\end{equation}
Although flow matching does not explicitly train a score function, the velocity field implicitly defines one through this relation.

\paragraph{Sampling.}
\label{sec:fm_sampling}
Samples are generated by solving the ODE
\begin{equation}
    \frac{d\mathbf{x}_t}{dt} = \mathbf{v}_\theta(\mathbf{x}_t, t),
    \label{eq:fm_ode}
\end{equation}
from $\mathbf{x}_0 \sim \mathcal{N}(\mathbf{0}, \mathbf{I})$ at $t=0$ to $\mathbf{x}_1$ at $t=1$. We consider three solvers: an adaptive Dormand--Prince (dopri5) solver \citep{dormand1980family}, our default; a fixed-step forward Euler solver; and Euler augmented with a Langevin corrector derived from the velocity-implied score (\autoref{eq:fm_score}). The discretizations are given in \autoref{app:samplers}. A caveat specific to the flow-derived score is that, because the denominator $(1-t) \to 0$ as $t \to 1$, velocity-estimation errors are amplified near the terminal time, causing systematic overestimation and sample over-concentration \citep{liu2024rectifiedflow}; the diffusion PC sampler instead computes the score directly from $D_\theta$ with EDM preconditioning \citep{karras2022elucidating} that stabilizes it across noise levels. Sampler design choices are ablated in \autoref{app:solver}.

\subsection{Sampler discretizations}
\label{app:samplers}

This subsection collects the discretized samplers referenced in \autoref{sec:methods}.

\subsubsection{Predictor-corrector sampler}
In practice, we employ a predictor-corrector (PC) sampler \citep{manshausen2025generative}: a discretized solver for the probability flow ODE (\autoref{eq:edm_ode_t}), augmented with stochastic Langevin corrections. We discretize $0 = t_0 < t_1 < \cdots < t_N = 1$ uniformly and define $s_i \equiv s(t_i)$, $\sigma_i \equiv \sigma(t_i)$ via the cosine schedule
\begin{equation}
    s(t) = \cos^2\!\bigl(\omega\,(1-t)\bigr), \qquad
    \sigma(t) = \frac{\sqrt{1 - s(t)^2}}{s(t)},
    \qquad \omega = \arccos\sqrt{\eta},
    \label{eq:pc_schedule}
\end{equation}
so that $s_i$ increases from $\eta$ to $1$ and $\sigma_i$ decreases accordingly. The \emph{predictor} advances from $t_i$ to $t_{i+1}$ via the exponential integrator (EI) scheme \citep{zhang2022fast}, treating the denoiser output $D_\theta$ as locally constant over $[t_i,\, t_{i+1}]$:
\begin{equation}
    \mathbf{x}_{t_{i+1}}
    = \frac{s_{i+1}\,\sigma_{i+1}}{s_i\,\sigma_i}\,\mathbf{x}_{t_i}
    \;+\; s_{i+1}\!\left(1 - \frac{\sigma_{i+1}}{\sigma_i}\right)
      D_\theta\!\bigl(\mathbf{x}_{t_i}/s_i;\;\sigma_i\bigr).
    \label{eq:predictor}
\end{equation}
The \emph{corrector} applies $K$ Langevin MCMC steps at $\sigma_{i+1}$ using the score (\autoref{eq:score_denoiser}), updating the running noisy state $\mathbf{x} \equiv \mathbf{x}_{t_{i+1}}$:
\begin{equation}
    \mathbf{x} \;\leftarrow\; \mathbf{x} + \delta\,\nabla_{\mathbf{x}}\!\log p_{\sigma_{i+1}}(\mathbf{x}) + \sqrt{2\delta}\;\boldsymbol{\xi}, \qquad \boldsymbol{\xi} \sim \mathcal{N}(\mathbf{0},\,\mathbf{I})
    \label{eq:langevin_corrector}
\end{equation}
with adaptive step size $\delta = \tau \big/ \bigl\langle\bigl\|\nabla_{\mathbf{x}}\log p_{\sigma_{i+1}}(\mathbf{x})\bigr\|^2\bigr\rangle$, where $\tau$ is a scalar hyperparameter and $\langle\cdot\rangle$ denotes the spatial mean. This normalization holds the per-step signal-to-noise ratio approximately constant across noise levels; $\tau$ controls this ratio rather than a step size directly. The predictor is entirely deterministic; all stochasticity comes from the corrector, which uses the score to project samples toward the support of $p_{\sigma_{i+1}}$. With $K$ corrector steps per predictor step and $N$ predictor steps, the total cost is $N(1+K)$ network evaluations.

\subsubsection{Flow-matching solvers}
Samples are generated by solving the flow-matching ODE (\autoref{eq:fm_ode}) from $\mathbf{x}_0 \sim \mathcal{N}(\mathbf{0}, \mathbf{I})$ at $t=0$ to $\mathbf{x}_1$ at $t=1$ with one of three solvers.

\paragraph{Adaptive ODE solver.}
The Dormand--Prince method (dopri5) \citep{dormand1980family} solves \autoref{eq:fm_ode} with adaptive step-size control. At each step, the solver computes a fourth- and fifth-order Runge--Kutta estimate and adjusts the step size to keep the local truncation error below a specified tolerance. This automatically concentrates function evaluations in regions where $\mathbf{v}_\theta$ varies rapidly, without requiring a predetermined step count.

\paragraph{Fixed-step Euler.}
With a uniform grid $0 = t_0 < t_1 < \cdots < t_N = 1$, the forward Euler method advances the solution via
\begin{equation}
    \mathbf{x}_{t_{i+1}} = \mathbf{x}_{t_i} + h\,\mathbf{v}_\theta\!\bigl(\mathbf{x}_{t_i},\, t_i\bigr),
    \label{eq:euler}
\end{equation}
where $h = t_{i+1} - t_i$. This requires exactly $N$ network evaluations and a fixed memory footprint, making it attractive when computational cost must be predictable.

\paragraph{Euler with Langevin corrector.}
Although flow matching is formulated as a deterministic ODE, the velocity field implicitly defines a score function via \autoref{eq:fm_score}. This enables a predictor-corrector scheme analogous to the diffusion PC sampler (\autoref{sec:diffusion}). After each Euler predictor step (\autoref{eq:euler}), we apply $K$ Langevin corrector steps:
\begin{equation}
    \mathbf{x} \;\leftarrow\; \mathbf{x} + \delta\,\nabla_{\mathbf{x}} \log p_{t_{i+1}}(\mathbf{x}) + \sqrt{2\delta}\;\boldsymbol{\xi}, \qquad \boldsymbol{\xi} \sim \mathcal{N}(\mathbf{0},\,\mathbf{I}),
    \label{eq:fm_corrector}
\end{equation}
with $\delta = \tau / \langle\|\nabla_{\mathbf{x}} \log p_{t_{i+1}}(\mathbf{x})\|^2\rangle$, mirroring the adaptive step size used in the diffusion corrector (\autoref{eq:langevin_corrector}). Each corrector step requires one additional evaluation of $\mathbf{v}_\theta$ to compute the score, bringing the total cost to $N(1 + K)$ evaluations.

\subsubsection{Guided samplers}
Recall the observation-space gradient of \autoref{sec:diffusion_sda},
\begin{equation}
    \mathbf{g} \;\triangleq\;
    \nabla_{\mathbf{x}}\!\left[
      \tfrac{1}{2}\bigl\|\mathbf{y}^*
      - \mathcal{H}\bigl(D_\theta(\mathbf{x}/s;\,\sigma)\bigr)\bigr\|^2
    \right],
    \label{eq:obs_gradient}
\end{equation}
computed by backpropagation through $D_\theta$. With the SDA likelihood (\autoref{eq:likelihood}), the posterior score decomposes as
\begin{equation}
    \nabla_{\mathbf{x}} \log p(\mathbf{x} \mid \mathbf{y}^*)
    = \underbrace{\frac{s\,D_\theta - \mathbf{x}}{s^2\sigma^2}}_{\text{prior score}}
    \;-\; \underbrace{\frac{\mathbf{g}}{\sigma_o^2 + \gamma\,\sigma^2}}_{\text{likelihood score}}.
    \label{eq:posterior_score}
\end{equation}
Substituting the guided denoiser (\autoref{eq:guided_denoiser}) for $D_\theta$ in the PC sampler gives the guided predictor and corrector. The \emph{predictor} advances from $t_i$ to $t_{i+1}$:
\begin{equation}
    \mathbf{x}_{t_{i+1}}
    = \underbrace{\frac{s_{i+1}\,\sigma_{i+1}}{s_i\,\sigma_i}\,\mathbf{x}_{t_i}
    + s_{i+1}\!\left(1 - \frac{\sigma_{i+1}}{\sigma_i}\right) D_\theta\!\bigl(\mathbf{x}_{t_i}/s_i;\,\sigma_i\bigr)}_{\text{unguided predictor (\autoref{eq:predictor})}}
    \;-\; \frac{s_{i+1}\!\left(\sigma_i - \sigma_{i+1}\right) s_i\,\sigma_i}{\sigma_o^2 + \gamma\,\sigma_i^2}\;\mathbf{g}_i,
    \label{eq:guided_predictor}
\end{equation}
where $\mathbf{g}_i$ denotes the observation-space gradient (\autoref{eq:obs_gradient}) evaluated at $(\mathbf{x}_{t_i}, \sigma_i)$. The \emph{corrector} applies $K$ Langevin steps at $\sigma_{i+1}$ using the full posterior score:
\begin{equation}
    \mathbf{x} \;\leftarrow\; \mathbf{x} + \delta\,\nabla_{\mathbf{x}}\!\log p_{\sigma_{i+1}}(\mathbf{x} \mid \mathbf{y}^*) + \sqrt{2\delta}\;\boldsymbol{\xi}, \quad \boldsymbol{\xi} \sim \mathcal{N}(\mathbf{0},\,\mathbf{I})
    \label{eq:guided_corrector}
\end{equation}
with $\delta = \tau \big/ \bigl\langle\bigl\|\nabla_{\mathbf{x}}\log p(\mathbf{x}\mid\mathbf{y}^*)\bigr\|^2\bigr\rangle$. The corrector uses the full posterior score (\autoref{eq:posterior_score}), simultaneously projecting samples back toward the data manifold and enforcing observation consistency. Equivalently, substituting $D_\theta^{\text{guided}}$ into the probability flow ODE (\autoref{eq:edm_ode_t}) gives the guided ODE:
\begin{equation}
    \frac{d\mathbf{x}}{dt}
    = \mathbf{v}_\theta(\mathbf{x},t)
    + \frac{\dot{\sigma}\,s^2\,\sigma}{\sigma_o^2 + \gamma\,\sigma^2}\,\mathbf{g},
    \label{eq:guided_ode}
\end{equation}
where $\dot{\sigma} = d\sigma/dt < 0$, so the second term steers the trajectory toward observations.

For Flow Guidance, the guided ODE can use any of the three solvers above; for the Euler + corrector variant, substituting the flow-matching parameterization $\sigma = (1{-}t)/t$ into the Diffusion-SDA posterior (\autoref{eq:posterior_score}) gives
\begin{equation}
    \nabla_{\mathbf{x}} \log p_t(\mathbf{x} \mid \mathbf{y}^*)
    = \frac{t\,\mathbf{v}_\theta - \mathbf{x}}{1-t}
    \;-\; \frac{\mathbf{g}}
         {\sigma_o^2 + \gamma\,(1{-}t)^2/t^2},
    \label{eq:fm_posterior_score}
\end{equation}
where $\mathbf{g} = \nabla_{\mathbf{x}}\,\mathcal{L}_\text{obs}(\hat{\mathbf{x}}_1)$ is the observation-loss gradient through the terminal extrapolation. The corrector step (\autoref{eq:fm_corrector}) then uses this posterior score, simultaneously projecting samples toward the data manifold and enforcing observation consistency at each ODE step.

\FloatBarrier
\section{Inference-time conditioning}
\label{app:design_space}

\subsection{Likelihood-approximation taxonomy}
\label{app:likelihood_taxonomy}

Every per-step guidance method in this work approximates the intractable likelihood by a Gaussian centered on the denoised estimate,
\begin{equation*}
    p(\mathbf{y}^* \mid \mathbf{x}_t)
    \;\approx\;
    \mathcal{N}\!\bigl(\mathcal{H}(\hat{\mathbf{x}}_1(\mathbf{x}_t)),\;\boldsymbol{\Sigma}_{\text{eff}}(t)\bigr),
\end{equation*}
from which the universal gradient form of \autoref{sec:flow_da} follows. Following \citep{rozet2023score,manshausen2025generative,shysheya2024conditional}, we approximate the likelihood as
\begin{equation}
    p(\mathbf{y}^* \mid \mathbf{x})
    \approx \mathcal{N}\!\Big(\mathbf{y}^*;\;
    \mathcal{H}\bigl(D_\theta(\mathbf{x}/s;\,\sigma)\bigr),\;
    (\sigma_o^2 + \gamma\,\sigma^2)\,\mathbf{I}\Big),
    \label{eq:likelihood}
\end{equation}
one specific choice of the Gaussian-approximated conditional $p(\mathbf{y}^*\mid\mathbf{x}_t) \approx \mathcal{N}(\mathbf{y}^*;\,\mathcal{H}(\hat{\mathbf{x}}_1),\,\boldsymbol{\Sigma})$ used in posterior-guidance methods. The scalar form $(\sigma_o^2 + \gamma\sigma^2)\mathbf{I}$ is the $\mathbf{H}\mathbf{H}^\top \approx \mathbf{I}$ simplification of the SDA matrix form $\boldsymbol{\Sigma}_y + \mathbf{H}\boldsymbol{\Sigma}_0\mathbf{H}^\top$. \autoref{tab:likelihood_variants} summarizes the main alternatives, distinguished by how the effective covariance $\boldsymbol{\Sigma}$ accounts for uncertainty in the terminal estimate $\hat{\mathbf{x}}_1$. We use the SDA form throughout because (i) the closed-form $\boldsymbol{\Sigma}_0(t) = \gamma\sigma^2\mathbf{I}$ avoids per-step linear solves, and (ii) it interfaces cleanly with the velocity--denoiser identity (\autoref{eq:velocity_denoiser}).

\begin{table}[htbp]
\centering
\caption{Common Gaussian likelihood approximations $p(\mathbf{y}^*\mid\mathbf{x}_t) \approx \mathcal{N}(\mathbf{y}^*;\,\mathcal{H}(\hat{\mathbf{x}}_1),\,\boldsymbol{\Sigma})$ used in posterior-guidance methods. We use the SDA form throughout.}
\label{tab:likelihood_variants}
\small
\begin{tabular}{p{0.22\textwidth} p{0.32\textwidth} p{0.34\textwidth}}
\toprule
Variant & Covariance $\boldsymbol{\Sigma}$ & Notes \\
\midrule
DPS \citep{chung2023diffusion} & $\boldsymbol{\Sigma}_y$ & Ignores uncertainty in $\hat{\mathbf{x}}_1$. \\
SDA / $\Pi$GDM \citep{rozet2023score} & $\boldsymbol{\Sigma}_y + \mathbf{H}\,\boldsymbol{\Sigma}_0(t)\,\mathbf{H}^\top$ & $\boldsymbol{\Sigma}_0 = \gamma\sigma^2\mathbf{I}$ (SDA) or $r_t^2\mathbf{I}$ ($\Pi$GDM). \\
MMPS \citep{andry2025appa} & $\boldsymbol{\Sigma}_y + \mathbf{H}\,\mathbb{V}[\mathbf{x}_1\mid\mathbf{x}_t]\,\mathbf{H}^\top$ & Exact posterior covariance; conjugate-gradient solves per step. \\
NDTM \citep{pandey2025variational} & per-step variational optimization & Most general; high per-step cost. \\
\bottomrule
\end{tabular}
\end{table}

\FloatBarrier
\subsection{Flow Guidance design space}

We explore the Flow Guidance design space along four axes: guidance schedule (\autoref{app:schedule}), stop-gradient approximation (\autoref{app:sg}), terminal extrapolation (\autoref{app:terminal}), and ODE solver with corrector (\autoref{app:solver}); the pixel-vs-latent choice is addressed in \autoref{sec:latent}. Diffusion-SDA hyperparameters are not ablated and follow~\citep{manshausen2025generative}. Unless otherwise noted, all ablations use the default Flow Guidance configuration from \autoref{app:arch}: constant schedule, dopri5 adaptive solver, single-step terminal extrapolation, no stop gradient, $\lambda = 2 \times 10^5$.

\subsubsection{Guidance schedule}
\label{app:schedule}

Recall that in flow matching, $s(t) = t$ and $\sigma(t) = (1-t)/t$ (\autoref{sec:flow_matching}). Substituting this parameterization into the Diffusion-SDA guided denoiser (\autoref{eq:guided_denoiser}) and feeding the result into the probability flow ODE (\autoref{eq:edm_ode_t}) yields an analytically motivated schedule:
\begin{equation}
    \lambda_{\text{SDA}}(t)
    = \lambda\frac{(1-t)\,t}{\sigma_o^2\,t^2 + \gamma\,(1-t)^2},
    \label{eq:sda_schedule}
\end{equation}
which anneals guidance with the noise level: $\lambda_{\text{SDA}} \to 0$ as $t \to 0$ (pure noise) and $t \to 1$ (clean data), peaking at intermediate $t$. We also consider three heuristic schedules: a constant $\lambda(t) = \lambda$, a cosine-increasing schedule $\lambda(t) = \lambda\,(1 - \cos\pi t)/2$ that ramps guidance toward the clean end, and a cosine-decreasing schedule $\lambda(t) = \lambda\,(1 + \cos\pi t)/2$ that concentrates guidance at the noisy end. Only $\lambda_{\text{SDA}}$ has a principled derivation from the SDA likelihood approximation \citep{rozet2023score}; the other three are heuristics.

\begin{table}[htbp]
\centering
\caption{Ablation: guidance schedule. \textit{Top}: effect of removing guidance from 20\%-wide windows (constant schedule, $\lambda{=}2{\times}10^5$). \textit{Bottom}: comparison of four schedule shapes (\autoref{fig:guidance_schedules}), each at its optimal $\lambda$. All use dopri5, single-step terminal extrapolation, without stop gradient. RMSE improvement (\%) over ERA5 at training (Tr) and test (Te) stations; total wall-clock cost (min) for 360 snapshots.}
\label{tab:ablation_schedule}
\tiny
\begin{tabular}{l cccc cccc cc rr}
\toprule
& \multicolumn{2}{c}{\uten} & \multicolumn{2}{c}{\vten} & \multicolumn{2}{c}{\ttwom} & \multicolumn{2}{c}{\dtwom} & \multicolumn{2}{c}{Avg} & \multicolumn{2}{c}{Cost} \\
\cmidrule(lr){2-3}\cmidrule(lr){4-5}\cmidrule(lr){6-7}\cmidrule(lr){8-9}\cmidrule(lr){10-11}\cmidrule(lr){12-13}
Configuration & Tr & Te & Tr & Te & Tr & Te & Tr & Te & Tr & Te & Mem & Total \\
\midrule
\multicolumn{13}{l}{\emph{No-guidance window} ($\lambda = 2\times 10^5$, constant elsewhere)} \\
None (default)            & 53.2 & 38.2 & 53.0 & 38.8 & 42.9 & 29.9 & 49.1 & 35.9 & 49.6 & 35.7 & 3.05 & 380 \\
$[0.0,0.2]$               & 53.3 & 38.0 & 53.1 & 38.7 & 42.3 & 28.3 & 48.7 & 34.4 & 49.3 & 34.9 & 3.05 & 203 \\
$[0.2,0.4]$               & 53.1 & 38.3 & 52.9 & 38.8 & 42.7 & 29.7 & 48.9 & 35.7 & 49.4 & 35.6 & 3.05 & 393 \\
$[0.4,0.6]$               & 52.9 & 38.3 & 52.7 & 38.9 & 42.5 & 29.6 & 48.6 & 35.7 & 49.2 & 35.6 & 3.05 & 388 \\
$[0.6,0.8]$               & 52.5 & 38.5 & 52.2 & 39.0 & 41.9 & 29.3 & 48.0 & 35.6 & 48.6 & 35.6 & 3.05 & 379 \\
$[0.8,1.0]$               & 44.7 & 37.4 & 44.6 & 38.3 & 26.6 & 22.7 & 36.5 & 31.1 & 38.1 & 32.4 & 3.05 & 387 \\
\midrule
\multicolumn{13}{l}{\emph{Schedule shape} (each at its optimal $\lambda$)} \\
Constant ($2\times 10^5$)  & 53.2 & 38.2 & 53.0 & 38.8 & 42.9 & 29.9 & 49.1 & 35.9 & 49.6 & 35.7 & 3.05 & 380 \\
Cosine$\uparrow$ ($0\to 5\times 10^5$) & 58.4 & 36.1 & 58.4 & 36.7 & 50.7 & 30.3 & 55.1 & 34.6 & 55.6 & 34.4 & 3.05 & 238 \\
Cosine$\downarrow$ ($5\times 10^5\to 0$) & 44.8 & 37.4 & 44.6 & 38.3 & 26.7 & 22.8 & 36.7 & 31.3 & 38.2 & 32.5 & 3.05 & 703 \\
SDA ($\lambda=10^6$)       & 50.0 & 37.5 & 49.9 & 38.4 & 36.3 & 27.8 & 44.9 & 34.7 & 45.3 & 34.6 & 3.05 & 296 \\
\bottomrule
\end{tabular}
\end{table}

\begin{figure}[htbp]
\centering
\includegraphics[width=0.8\textwidth]{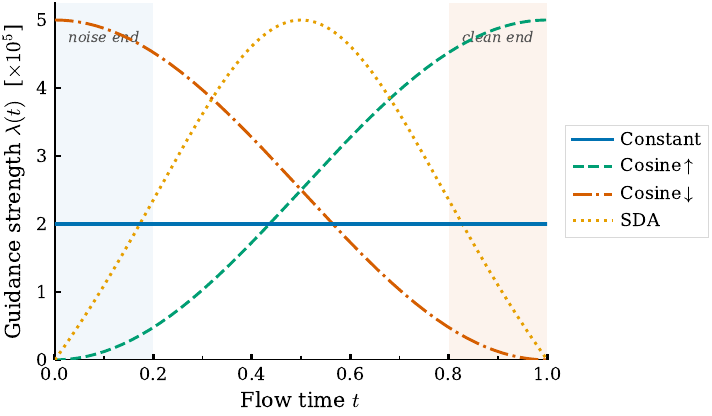}
\caption{The four guidance schedules compared in \autoref{tab:ablation_schedule}. The constant schedule maintains uniform guidance strength throughout the trajectory; cosine-increasing concentrates guidance near $t{=}1$; cosine-decreasing concentrates it near $t{=}0$; and the SDA-derived schedule (\autoref{eq:sda_schedule}) peaks at intermediate $t$ and vanishes at both endpoints.}
\label{fig:guidance_schedules}
\end{figure}

We first probe \emph{where} along the trajectory guidance matters by removing it from successive 20\%-wide windows (\autoref{tab:ablation_schedule}, top). Removing guidance near the clean endpoint ($t \in [0.8, 1.0]$) causes the largest drop ($-3.3$\,\text{pp}), removing the noisy endpoint ($t \in [0, 0.2]$) has a moderate effect ($-0.8$\,\text{pp}), while removing any middle segment ($t \in [0.2, 0.8]$) changes test accuracy by at most $0.1$\,\text{pp}. Late-stage guidance, where the terminal-state estimate is most accurate, matters most.

\autoref{tab:ablation_schedule} (bottom) compares four guidance schedules (\autoref{fig:guidance_schedules}), each at its optimal $\lambda$. The constant schedule achieves the best average test improvement ($35.7\%$), followed by the SDA-derived schedule ($34.6\%$), cosine-increasing ($34.4\%$), and cosine-decreasing ($32.5\%$). All shaped schedules underperform constant, suggesting that maintaining guidance strength throughout the trajectory is more practical than tuning the schedule profile.

\subsubsection{Stop-gradient approximation}
\label{app:sg}

\label{sec:flowdps}
FlowDPS \citep{kim2025flowdps} operates in terminal space: at each ODE step, it iterates $K$ stop-gradient optimization steps on $\hat{\mathbf{x}}_1$ in the terminal space, then resamples the corrected terminal state to the next ODE step. The $K$ inner iterations let FlowDPS accumulate a larger terminal-space correction at each ODE step than a single stop-gradient update, at the cost of $K$ extra forward passes per step and without requiring backpropagation through the full ODE. The trajectory can then either continue from the corrected state or resample from the corrected endpoint; both variants are ablated below.

\begin{table}[htbp]
\centering
\caption{Ablation: stop gradient (SG) and FlowDPS. RMSE improvement (\%) over ERA5 at training (Tr) and test (Te) stations, and total wall-clock cost (min) for 360 snapshots. Without SG, gradients backpropagate through the velocity network. FlowDPS-$K$ denotes FlowDPS with $K$ optimization steps per ODE step.}
\label{tab:ablation_sg}
\tiny
\begin{tabular}{l cccc cccc cc rr}
\toprule
& \multicolumn{2}{c}{\uten} & \multicolumn{2}{c}{\vten} & \multicolumn{2}{c}{\ttwom} & \multicolumn{2}{c}{\dtwom} & \multicolumn{2}{c}{Avg} & \multicolumn{2}{c}{Cost} \\
\cmidrule(lr){2-3}\cmidrule(lr){4-5}\cmidrule(lr){6-7}\cmidrule(lr){8-9}\cmidrule(lr){10-11}\cmidrule(lr){12-13}
Method & Tr & Te & Tr & Te & Tr & Te & Tr & Te & Tr & Te & Mem & Total \\
\midrule
w/o SG (default)   & 53.2 & 38.2 & 53.0 & 38.8 & 42.9 & 29.9 & 49.1 & 35.9 & 49.6 & 35.7 & 3.05 & 380 \\
w/  SG             & 60.9 & 34.1 & 61.0 & 34.9 & 54.6 & 27.8 & 58.1 & 32.0 & 58.6 & 32.2 & 0.37 &  77 \\
\midrule
\multicolumn{13}{l}{\textit{FlowDPS (terminal-space optimization)}} \\
FlowDPS, $K=5$     & 52.7 & 35.5 & 52.5 & 35.7 & 27.3 &  2.1 & 38.2 & 12.0 & 42.7 & 21.3 & 0.25 &  28 \\
FlowDPS, $K=10$    & 57.8 & 34.5 & 57.7 & 35.2 & 45.2 & 16.9 & 51.5 & 22.6 & 53.0 & 27.3 & 0.25 &  31 \\
FlowDPS, $K=20$    & 60.5 & 32.4 & 60.6 & 33.4 & 51.8 & 21.4 & 56.2 & 25.6 & 57.3 & 28.2 & 0.25 &  36 \\
FlowDPS, $K=50$    & 62.7 & 29.7 & 62.8 & 30.6 & 55.3 & 21.8 & 59.2 & 25.4 & 60.0 & 26.9 & 0.25 &  52 \\
FlowDPS, $K=100$   & 64.0 & 27.1 & 64.1 & 27.9 & 57.0 & 19.7 & 60.8 & 23.0 & 61.5 & 24.4 & 0.25 &  82 \\
\bottomrule
\end{tabular}
\end{table}

\autoref{tab:ablation_sg} compares the stop-gradient approximation with full-gradient Flow Guidance. With stop gradient, the guidance gradient is computed only through the terminal extrapolation $\hat{\mathbf{x}}_1 = \mathbf{x} + (1-t)\,\mathbf{v}_\theta$, treating $\mathbf{v}_\theta$ as a fixed function; without stop gradient, the gradient backpropagates through $\mathbf{v}_\theta$. Since $\hat{\mathbf{x}}_1$ is then affine in $\mathbf{x}$,
\begin{equation}
    \nabla_{\mathbf{x}}\,\mathcal{L}_\text{obs}(\hat{\mathbf{x}}_1)
    \;\overset{\text{stop-grad}}{=}\;
    \nabla_{\hat{\mathbf{x}}_1}\,\mathcal{L}_\text{obs}(\hat{\mathbf{x}}_1),
    \label{eq:stop_gradient}
\end{equation}
i.e., the guidance gradient in $\mathbf{x}$-space equals the gradient in terminal space, avoiding Jacobian-vector products through the network. After correcting the terminal state, the trajectory can either continue from the corrected state at the same ODE step \citep{meng2022diffusion}, or resample from the corrected terminal endpoint through the generative model \citep{kim2025flowdps,qu2025pnp,sun2025sda}: the \emph{continue} variant corresponds to the ``SG'' rows of \autoref{tab:ablation_sg}, and the \emph{resample} variant corresponds to the FlowDPS-$K$ rows.

The stop-gradient variant reduces test improvement from $35.7\%$ to $32.2\%$ ($-3.5$\,\text{pp}) while lowering memory by $8\times$ (3.05 to 0.37\,GB) and time by ${\sim}5\times$ (380 to 77\,min). The accuracy difference is roughly uniform across variables on the dense benchmark; under sparsity, the degradation concentrates in temperature and dewpoint (\autoref{tab:sparse_benchmark}).

\autoref{tab:ablation_sg} also includes FlowDPS. It peaks at $K{=}20$ ($28.2\%$) and degrades with additional iterations (down to $24.4\%$ at $K{=}100$). Like the stop-gradient variant, it operates in terminal space without backpropagating through $\mathbf{v}_\theta$. Iterative optimization in terminal space additionally tends to overfit observation noise on real station data, similar to 3D-Var without ERA5 regularization (\autoref{sec:dense_benchmark}): as $K$ grows, Tr improves while Te plateaus and then declines. Neither stop-gradient variant approaches full-gradient Flow Guidance.

\subsubsection{Terminal extrapolation}
\label{app:terminal}

\begin{table}[htbp]
\centering
\caption{Ablation: terminal extrapolation. RMSE improvement (\%) over ERA5 at training (Tr) and test (Te) stations, plus total wall-clock cost (min) for 360 snapshots.}
\label{tab:ablation_terminal}
\tiny
\begin{tabular}{l cccc cccc cc rr}
\toprule
& \multicolumn{2}{c}{\uten} & \multicolumn{2}{c}{\vten} & \multicolumn{2}{c}{\ttwom} & \multicolumn{2}{c}{\dtwom} & \multicolumn{2}{c}{Avg} & \multicolumn{2}{c}{Cost} \\
\cmidrule(lr){2-3}\cmidrule(lr){4-5}\cmidrule(lr){6-7}\cmidrule(lr){8-9}\cmidrule(lr){10-11}\cmidrule(lr){12-13}
Extrapolation & Tr & Te & Tr & Te & Tr & Te & Tr & Te & Tr & Te & Mem & Total \\
\midrule
Single-step (default)  & 53.2 & 38.2 & 53.0 & 38.8 & 42.9 & 29.9 & 49.1 & 35.9 & 49.6 & 35.7 & 3.05 & 380 \\
RK4 two-step           & 55.7 & 37.6 & 55.4 & 38.3 & 46.4 & 31.0 & 51.7 & 36.4 & 52.3 & 35.8 & 25.21 & 10{,}659 \\
\bottomrule
\end{tabular}
\end{table}

The terminal extrapolation $\hat{\mathbf{x}}_1 = \mathbf{x} + (1-t)\,\mathbf{v}_\theta(\mathbf{x}, t)$ used in the observation loss (\autoref{eq:guided_velocity}) is a single-step forward Euler extrapolation. Replacing this with an RK4 two-step extrapolation (\autoref{tab:ablation_terminal}) yields essentially no gain in average test accuracy ($35.8\%$ vs.\ $35.7\%$), though the per-variable pattern shows a small temperature and dewpoint improvement (\ttwom\ $+1.1$\,\text{pp}, \dtwom\ $+0.5$\,\text{pp}) offset by a wind penalty (\uten\ $-0.6$\,\text{pp}, \vten\ $-0.5$\,\text{pp}). Meanwhile, memory increases ${\sim}8\times$ ($3.05$ to $25.21$\,GB) and wall-clock ${\sim}28\times$ ($380$ to $10{,}659$\,min) due to the additional gradient-tracked evaluations of $\mathbf{v}_\theta$. Given the cost, single-step extrapolation is used as the default throughout this work.

\subsubsection{ODE solver choice}
\label{app:solver}

\begin{table}[htbp]
\centering
\caption{Ablation: ODE solver and corrector. RMSE improvement (\%) over ERA5 at training (Tr) and test (Te) stations, plus total wall-clock cost (min) for 360 snapshots. Flow Guidance dopri5 uses $\lambda{=}2{\times}10^5$; Euler-128 variants use $\lambda{=}10^5$. ``PO'' = predictor-only; ``PC'' adds a Langevin corrector between predictor steps.}
\label{tab:ablation_solver}
\tiny
\setlength{\tabcolsep}{2.5pt}\resizebox{0.85\textwidth}{!}{%
\begin{tabular}{l cccc cccc cc rr}
\toprule
& \multicolumn{2}{c}{\uten} & \multicolumn{2}{c}{\vten} & \multicolumn{2}{c}{\ttwom} & \multicolumn{2}{c}{\dtwom} & \multicolumn{2}{c}{Avg} & \multicolumn{2}{c}{Cost} \\
\cmidrule(lr){2-3}\cmidrule(lr){4-5}\cmidrule(lr){6-7}\cmidrule(lr){8-9}\cmidrule(lr){10-11}\cmidrule(lr){12-13}
Sampler & Tr & Te & Tr & Te & Tr & Te & Tr & Te & Tr & Te & Mem & Total \\
\midrule
Flow Guidance dopri5 (PO, default) & 53.2 & 38.2 & 53.0 & 38.8 & 42.9 & 29.9 & 49.1 & 35.9 & 49.6 & 35.7 & 3.05 & 380 \\
Flow Guidance Euler-128 (PO)       & 49.2 & 38.6 & 48.8 & 39.0 & 34.7 & 24.1 & 40.8 & 29.6 & 43.4 & 32.8 & 2.92 &  53 \\
Flow Guidance Euler-128 (PC)       & 48.9 & 38.7 & 48.7 & 39.3 & 35.4 & 26.9 & 42.9 & 34.5 & 44.0 & 34.9 & 2.93 & 208 \\
Diffusion-SDA PC (default)         & 52.7 & 37.8 & 52.2 & 38.6 & 39.9 & 29.6 & 47.6 & 35.6 & 48.1 & 35.4 & 6.12 & 246 \\
Diffusion-SDA PO                   & 48.3 & 37.9 & 47.9 & 38.3 & 29.5 & 19.9 & 36.7 & 25.0 & 40.6 & 30.3 & 6.11 &  82 \\
\bottomrule
\end{tabular}%
}
\end{table}

\autoref{tab:ablation_solver} compares the default adaptive Dormand--Prince (dopri5) solver against a fixed-step Euler-128 solver, with and without a Langevin predictor-corrector (PC). Diffusion-SDA with and without corrector is included for reference.

\paragraph{The corrector helps diffusion substantially but flow matching modestly.}
Removing the Langevin corrector from Diffusion-SDA drops test improvement by $5.1$\,\text{pp} ($35.4\% \to 30.3\%$), while adding a corrector to Flow Guidance Euler-128 improves accuracy by only $2.1$\,\text{pp} ($32.8\% \to 34.9\%$). The asymmetry is more pronounced under sparsity ($+8.5$\,\text{pp} for diffusion vs.\ $+2.1$\,\text{pp} for flow matching; \autoref{tab:sparse_benchmark}). This controlled comparison (128-step fixed predictor for both frameworks: Euler on the flow ODE, EI on the VP-SDE) reveals complementary parameterization trade-offs: Flow Guidance starts $2.5$\,\text{pp} ahead of Diffusion-SDA without the corrector ($32.8\%$ vs.\ $30.3\%$), reflecting its more accurate velocity predictor, but gains less from the corrector because its flow-derived score is weaker (\autoref{app:corrector}).

\paragraph{Adaptive vs.\ fixed-step solver.}
The adaptive dopri5 solver achieves higher test accuracy ($35.7\%$) than fixed-step Euler-128 ($32.8\%$) in the dense benchmark, but with a larger generalization gap ($13.9$ vs.\ $10.6$\,\text{pp}). This ranking reverses under sparsity (\autoref{tab:sparse_benchmark}): Euler-128 PC ($22.4\%$) outperforms dopri5 ($21.7\%$), suggesting that the adaptive solver's tighter fit to training stations becomes a liability when observations are few. Flow Guidance Euler-128 PC and Diffusion-SDA PC use comparable sampling procedures (128 fixed-size predictor-corrector steps), so their similar accuracy ($34.9\%$ vs.\ $35.4\%$) and gap ($9.1$ vs.\ $12.7$\,\text{pp}) reflect the two frameworks under a controlled solver configuration.

\paragraph{All three frameworks converge.}
Flow Guidance dopri5, Euler-128 PC, and Diffusion-SDA all achieve ${\sim}35\%$ test improvement; the choice among them is driven by memory, speed, and hyperparameter complexity rather than accuracy.

\FloatBarrier
\section{The role of the Langevin corrector}
\label{app:corrector}

Although Flow Guidance and Diffusion-SDA achieve comparable test accuracy in both the dense and sparse benchmarks (\autoref{sec:dense_benchmark},~\ref{sec:sparse_benchmark}), their sampling mechanisms differ in an important way: Diffusion-SDA includes a Langevin corrector that projects each sample back toward the learned data manifold at every denoising step, whereas Flow Guidance (with its default adaptive solver) does not. The corrector suppresses off-manifold drift introduced by the observation-loss gradient, and it is central to diffusion's accuracy.

On the dense benchmark the corrector's gain is confined to temperature and dewpoint (\autoref{sec:dense_benchmark}). Under sparsity the same asymmetry persists, and the corrector remains more impactful for Diffusion-SDA ($-8.5$\,\text{pp} when removed) than for Flow Guidance ($+2.1$\,\text{pp} when added; \autoref{tab:sparse_benchmark}).

The corrector transfers only partially to flow matching, and the perturbation analysis of \autoref{app:perturbation} shows why: the flow score, derived indirectly from the velocity field, is noisier than the diffusion denoiser's score (a $5\times$ smaller stable Langevin step), so a flow-space corrector suppresses off-manifold perturbations less effectively. This exposes the complementary trade-off behind their parity: flow matching has the more accurate predictor but the weaker corrector, while diffusion has the weaker predictor but the stronger corrector, so each framework compensates for its weak component through its strong one and they land at parity overall.

\paragraph{The corrector as manifold projection.}
During guided sampling, observation-loss gradients steer the generated field toward consistency with station data, but they point in directions determined by the adjoint $\mathcal{H}^\top$, which spreads residuals from sparse station locations back onto the grid. Each gradient step therefore risks pushing the sample off the manifold of physically plausible states, introducing artifacts that are invisible at station locations but degrade the field elsewhere. The Langevin corrector counteracts this by projecting the sample back toward the data manifold at every denoising step: each corrector step adds noise scaled to the current noise level and then denoises using the learned score function, a local projection onto the high-probability region of the learned distribution.

This mechanism is directly analogous to how $\mathbf{B}^{-1}$ regularizes classical 3D-Var: in both cases, the regularizer prevents the analysis from being pulled too far from the prior by noisy or sparse observations. The difference is that $\mathbf{B}^{-1}$ enforces a Gaussian prior, while the Langevin corrector enforces a learned, non-Gaussian prior that adapts to the local structure of the data manifold. This explains the corrector's outsized benefit for Diffusion-SDA, with the degradation from removing it concentrated in temperature and dewpoint, whose corrections require finer precision in normalized space (\autoref{tab:normalized_era5}). For flow matching, the analogous corrector built from the velocity-derived score is weaker (\autoref{app:perturbation}), yet still provides a modest benefit on both benchmarks (\autoref{tab:ablation_solver},~\ref{tab:sparse_benchmark}).

\subsection{Perturbation analysis}
\label{app:perturbation}

\begin{figure}[htbp]
\centering
\includegraphics[width=\textwidth]{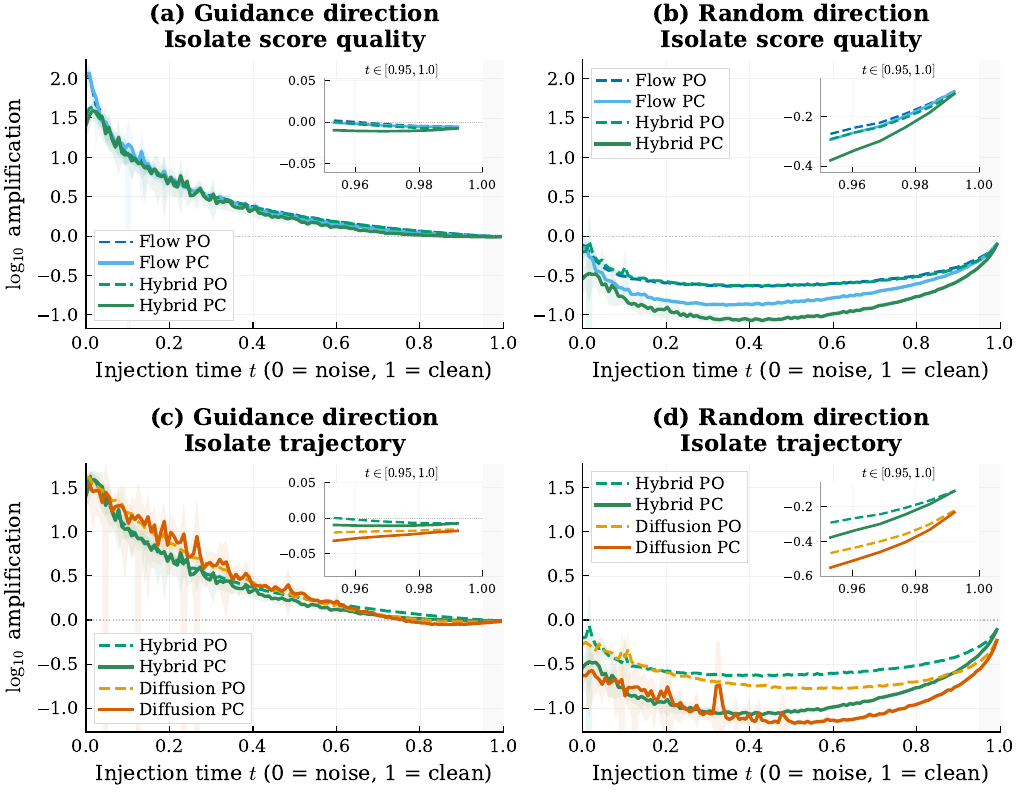}
\caption{Perturbation amplification ($\log_{10}$ scale) as a function of injection time $t$, averaged over $16$ trajectories with $\delta{=}0.01$. \textbf{Top row (a,b):} Isolate score quality by fixing the flow ODE trajectory. Flow PO/PC use the flow-derived score ($\tau{=}0.02$); Hybrid PO/PC use the diffusion denoiser's score ($\tau{=}0.1$) on the same trajectory. \textbf{Bottom row (c,d):} Isolate trajectory by fixing the diffusion score. Hybrid PO/PC integrate the flow ODE; Diffusion PO/PC integrate the VP-SDE ($\tau{=}0.1$). Left panels show guidance-direction perturbations; right panels show random-direction perturbations. Insets zoom into $t \in [0.95, 1.0]$.}
\label{fig:perturbation}
\end{figure}

To probe the corrector's stabilizing role, we inject small perturbations ($\delta = 0.01$) at each of $128$ evenly spaced points along the generative trajectory and measure the amplification ratio $\|\mathbf{x}_1^\text{pert} - \mathbf{x}_1^\text{ref}\|_2 / \delta$ at the output $t = 1$, averaged over $16$ trajectories (\autoref{fig:perturbation}). Perturbations are injected in two directions: along the observation-loss gradient (guidance direction) and along a random unit vector (random direction). We compare six sampler configurations that vary two factors: the ODE trajectory and the score function used for the corrector. \emph{Flow PO/PC} integrate the flow ODE with the flow-derived score ($\tau{=}0.02$). \emph{Hybrid PO/PC} integrate the same flow ODE but substitute the diffusion denoiser's score for the corrector ($\tau{=}0.1$). \emph{Diffusion PO/PC} integrate the VP-SDE with the diffusion score ($\tau{=}0.1$). ``PO'' denotes predictor-only (no corrector); ``PC'' adds a Langevin corrector.

\paragraph{Score quality is the primary factor.}
To isolate score quality, we fix the flow ODE trajectory and compare Flow PC with Hybrid PC (\autoref{fig:perturbation}a,b). At mid-trajectory, Hybrid PC attenuates random-direction perturbations to ${\approx}0.09$, versus ${\approx}0.14$ for Flow PC, confirming that score quality limits corrector effectiveness. The flow-matching score is derived indirectly from the velocity field, with estimation errors amplified by the $(1-t)^{-1}$ factor near the terminal time. This limits the maximum stable Langevin step size to $\tau \approx 0.02$ for the flow score, versus $\tau = 0.1$ for the diffusion score, a $5\times$ gap.

\paragraph{The noise schedule governs late-stage suppression.}
To isolate the trajectory, we compare Hybrid PC with Diffusion PC (\autoref{fig:perturbation}c,d). Both use the diffusion score; their similar mid-trajectory attenuation (${\approx}0.09$ vs.\ ${\approx}0.07$) confirms that the trajectory is not the primary factor. However, near $t = 1$ (inset of panel d), Diffusion PC retains stronger suppression than Hybrid PC, and the gap between Hybrid PO and PC narrows to zero on the flow trajectory. The adaptive step size in the Langevin corrector (\autoref{eq:langevin_corrector}) is $\delta = \tau / \langle\|\nabla \log p_\sigma\|^2\rangle$. Since $\|\nabla \log p_\sigma\| \propto 1/\sigma$ (the score grows as the marginal concentrates), the effective step size scales as $\delta_L \propto \tau\,\sigma^2$, vanishing quadratically as $\sigma \to 0$ near $t = 1$. Both schedules decay linearly in $(1{-}t)$ near the terminal time, but the VP cosine schedule (\autoref{eq:pc_schedule}) maintains $\sigma \approx \sqrt{2}\,\omega\,(1{-}t)$ with $\omega = \arccos\sqrt{\eta} \approx 1.39$, roughly $2\times$ larger than the flow-matching $\sigma = (1{-}t)/t \approx (1{-}t)$. The resulting ${\approx}4\times$ larger $\delta_L$ is visible in \autoref{fig:perturbation}(d) inset: Diffusion PC retains stronger suppression than Hybrid PC near $t = 1$, even though both use the same diffusion score and $\tau = 0.1$. In practice, a small noise floor $\sigma_{\min}$ prevents the corrector from becoming completely inert at the final steps.

\FloatBarrier
\section{Latent-space study}
\label{app:latent_study}

A priori we expect this to help: the four surface variables we assimilate (\uten, \vten, \ttwom, \dtwom) are tied together by well-known physical relationships, the two wind components share the same synoptic flow, and dewpoint is bounded above by temperature with the two co-varying through near-surface humidity, so an encoder that captures these dependencies should learn a latent representation in which the cross-variable background-error covariance is approximately diagonal, mirroring the role of $\mathbf{B}$ in classical variational methods (cf.\ \citep{fan2026physically}). The comparison is therefore conservative: if the smaller, variable-isolated $G{=}4$ matches $G{=}1$, cross-variable mixing in the autoencoder is unnecessary for effective latent-space assimilation.

\begin{figure}[htbp]
\centering
\includegraphics[width=0.7\textwidth]{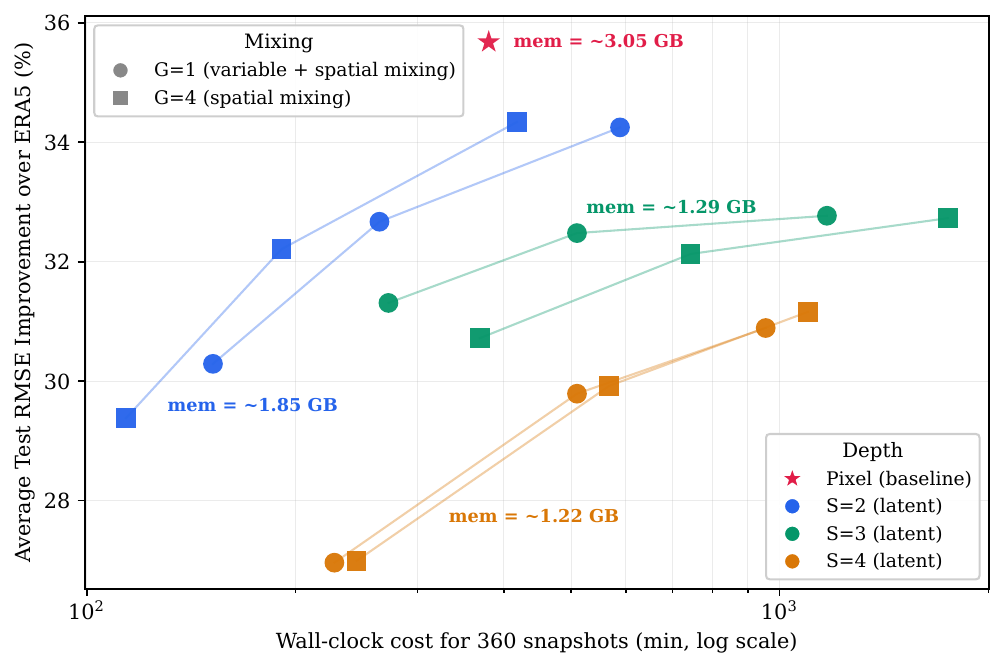}
\caption{Accuracy--compute trade-off for Flow Guidance in pixel vs.\ latent space. $x$-axis: total wall-clock cost for $360$ snapshots on $1$~GPU (log scale). $y$-axis: test-set RMSE improvement over ERA5, averaged over $\{\uten, \vten, \ttwom, \dtwom\}$. Colors denote downsampling depth $S$; marker shapes denote the attention grouping (circle: $G{=}1$, variable-plus-spatial-mixing; square: $G{=}4$, spatial-mixing-only). Lines connect configurations of the same architecture across guidance strengths~$\lambda$. Peak memory (approximately constant within each $S$-group) is annotated per group. Reported runtimes for $G{=}4$ at $S{\geq}3$ are inflated by a PyTorch grouped-convolution inefficiency that a mathematically-equivalent batch-reshape implementation removes ($1.5$--$4\times$ faster).}
\label{fig:latent}
\end{figure}

\autoref{fig:latent} compares Flow Guidance across pixel-space and latent-space autoencoders, varying the downsampling depth $S$ and the attention grouping ($G{=}1$ variable-plus-spatial-mixing, $G{=}4$ spatial-mixing-only). Three findings emerge.

\paragraph{Shallower autoencoders are preferred.}
All $S{\geq}3$ configurations are dominated by $S{=}2$: best improvement drops from $34.3\%$ at $S{=}2$ to $32.8\%$ at $S{=}3$ to $31.2\%$ at $S{=}4$, with the loss concentrated in temperature and dewpoint (${\sim}6$\,\text{pp} on \ttwom/\dtwom\ vs.\ ${\sim}1$\,\text{pp} on wind). This tracks reconstruction error, which rises sharply with compression (${\sim}20\times$ increase in mean absolute error (MAE) from $S{=}2$ to $S{=}4$; see \autoref{app:autoencoder}). Deeper downsampling also raises wall-clock cost: because every guidance step backpropagates the observation loss through the decoder, a larger $S$ increases the depth of the backward pass. The reported $G{=}4$ runtimes at $S{\geq}3$ are further inflated by a PyTorch grouped-convolution inefficiency (see caption of \autoref{fig:latent}).

\paragraph{Cross-variable mixing in the autoencoder is unnecessary.}
Running 3D-Var in three nonlinear regimes isolates the role of the autoencoder (\autoref{tab:ae_mixing_3dvar}): pixel space (no AE), latent space with cross-variable mixing ($G{=}1$), and latent space without it ($G{=}4$). The three variants are statistically indistinguishable ($33.30\%$, $33.29\%$, $33.49\%$). Flow Guidance shows the same pattern at $S{=}2$: the spatial-mixing-only $G{=}4$ configuration reaches peak accuracy ($34.34\%$) at ${\sim}27\%$ lower wall-clock than $G{=}1$ (peak $34.26\%$), using $4\times$ fewer autoencoder parameters. Cross-variable mixing in the AE therefore adds parameters without improving either DA or reconstruction accuracy (\autoref{app:autoencoder}).

\begin{table}[htbp]
\centering
\caption{3D-Var accuracy in three nonlinear regimes. RMSE improvement (\%) over ERA5 at training (Tr) and test (Te) stations, averaged across the four variables. The three variants are indistinguishable on Te.}
\label{tab:ae_mixing_3dvar}
\small
\begin{tabular}{l cc}
\toprule
3D-Var variant & Tr (\%) & Te (\%) \\
\midrule
Pixel (no AE)             & 49.7 & 33.3 \\
Latent ($S{=}2$, $G{=}1$) & 53.4 & 33.3 \\
Latent ($S{=}2$, $G{=}4$) & 55.5 & 33.5 \\
\bottomrule
\end{tabular}
\end{table}

\paragraph{The pixel baseline dominates except on memory.}
Pixel-space Flow Guidance achieves ${\sim}1.4$\,\text{pp} higher improvement than the best $S{=}2$ latent configuration at roughly $0.9\times$ the wall-clock, and is dominated only on peak memory ($3.05$ vs.\ $1.85$\,GB; ${\sim}1.6\times$). On the dense benchmark, latent compression is therefore useful only when memory is the binding constraint; otherwise pixel space is preferred on both accuracy and speed, and deeper latent compression provides no useful operating point.

\paragraph{Why variable mixing does not help.}
First, the normalized-space signal imbalance (\autoref{tab:normalized_era5}) makes variable mixing not just redundant but potentially harmful. Wind exhibits more fine-scale spatial structure than temperature and dewpoint, reflected in a ${\sim}2.4\times$ larger normalized RMSE. Therefore, a shared decoder ($G{=}1$) spends most of its capacity on the wind channels, diluting the temperature and dewpoint signal; \autoref{app:autoencoder} (\autoref{tab:ae_recon}) confirms the trade-off empirically: at $S{=}4$, $G{=}1$ reconstructs wind ${\sim}33\%$ better than $G{=}4$ but reconstructs \ttwom/\dtwom\ $19\%$ and $12\%$ worse. Variable isolation ($G{=}4$) removes this competition, yielding more balanced per-variable reconstruction error while matching $G{=}1$'s assimilation quality with $4\times$ fewer autoencoder parameters (\autoref{tab:ae_mixing_3dvar}).

Second, $G{=}4$ does not lose variable-mixing capacity overall because the flow model's U-Net already operates on all four channels jointly, capturing intervariable correlations during generation. The autoencoder's contribution is spatial: its decoder upsamples the compressed latent representation back to pixel space, spreading gradient signals from point station locations to the surrounding grid via learned convolutional kernels.

\subsection{Autoencoder reconstruction MAE}
\label{app:autoencoder}

\autoref{tab:ae_recon} reports per-variable reconstruction MAE for the variational autoencoder under both grouping choices ($G = 4$ spatial-mixing-only and $G = 1$ variable-plus-spatial mixing) and three compression depths $S \in \{2, 3, 4\}$. At $S = 2$ the two grouping choices reach essentially identical end-to-end assimilation accuracy ($34.34$ vs.\ $34.26\%$; \autoref{fig:latent}), confirming that variable mixing in the autoencoder is unnecessary. At deeper compression ($S = 4$), $G = 1$ reconstructs wind better but loses temperature and dewpoint detail; $G = 4$ keeps those two channels isolated and recovers them more faithfully.

\begin{table}[htbp]
\centering
\caption{Autoencoder reconstruction MAE on test fields in normalized units. ``Compr.'' is the spatial compression factor.}
\label{tab:ae_recon}
\tiny
\begin{tabular}{l l l cccc c}
\toprule
$G$ & $S$ & Compr. & \uten & \vten & \ttwom & \dtwom & MAE \\
\midrule
4 & 2 & 0.99$\times$  & 0.0069 & 0.0078 & 0.0175 & 0.0177 & 0.0125 \\
4 & 3 & 3.85$\times$  & 0.0819 & 0.0808 & 0.1482 & 0.1735 & 0.1211 \\
4 & 4 & 14.7$\times$  & 0.1982 & 0.2024 & 0.2688 & 0.3322 & 0.2504 \\
\midrule
1 & 2 & 0.99$\times$  & 0.0063 & 0.0067 & 0.0213 & 0.0218 & 0.0140 \\
1 & 3 & 3.85$\times$  & 0.0533 & 0.0573 & 0.1494 & 0.1897 & 0.1124 \\
1 & 4 & 14.7$\times$  & 0.1307 & 0.1377 & 0.3185 & 0.3720 & 0.2397 \\
\bottomrule
\end{tabular}
\end{table}

\FloatBarrier
\section{Additional results}

\subsection{Per-variable spatial decay}
\label{app:per_var}

\begin{figure}[htbp]
\centering
\includegraphics[width=\textwidth]{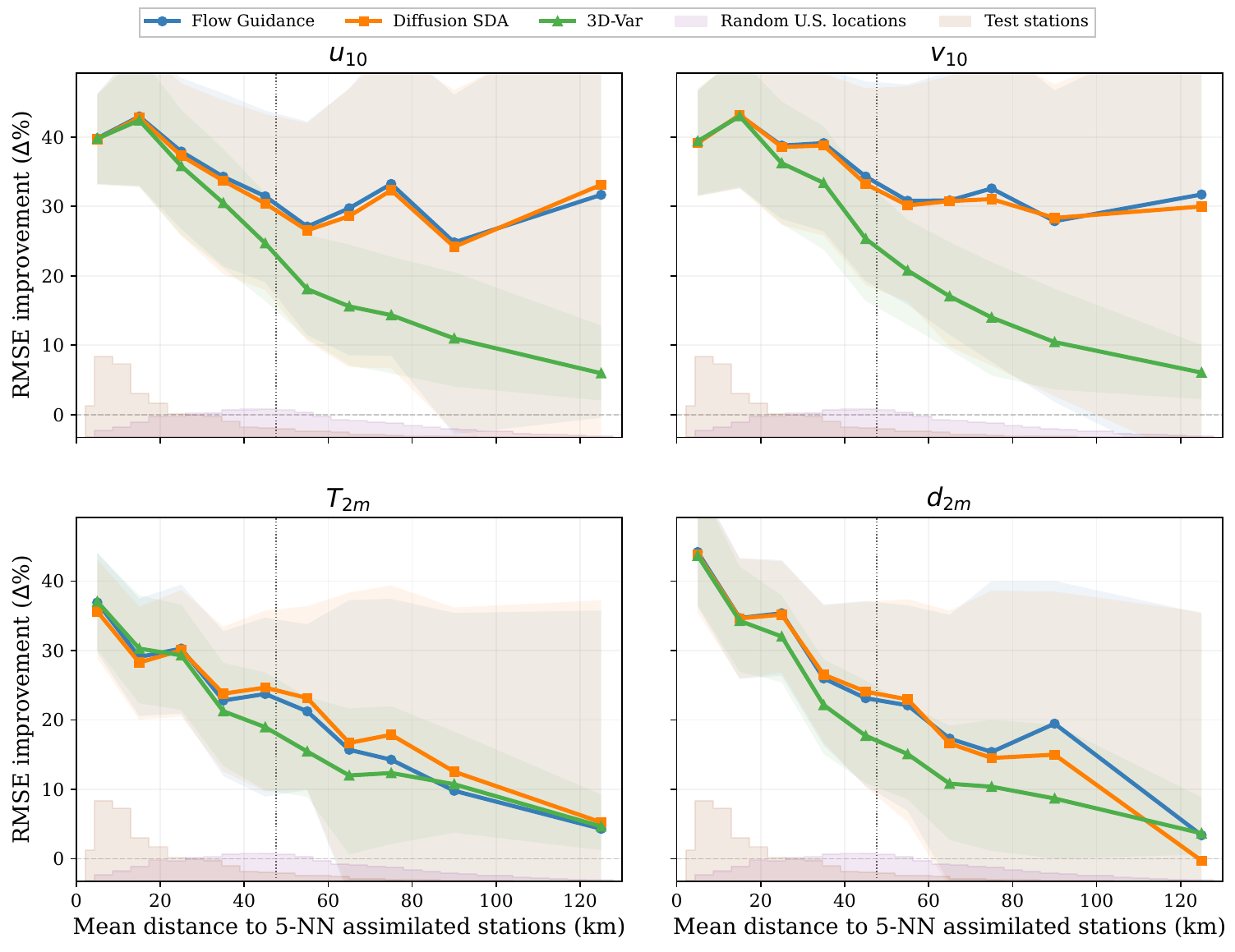}
\caption{Per-variable RMSE improvement vs.\ mean 5-NN distance, dense benchmark.}
\label{fig:per_var_dense_si}
\end{figure}

\begin{figure}[htbp]
\centering
\includegraphics[width=\textwidth]{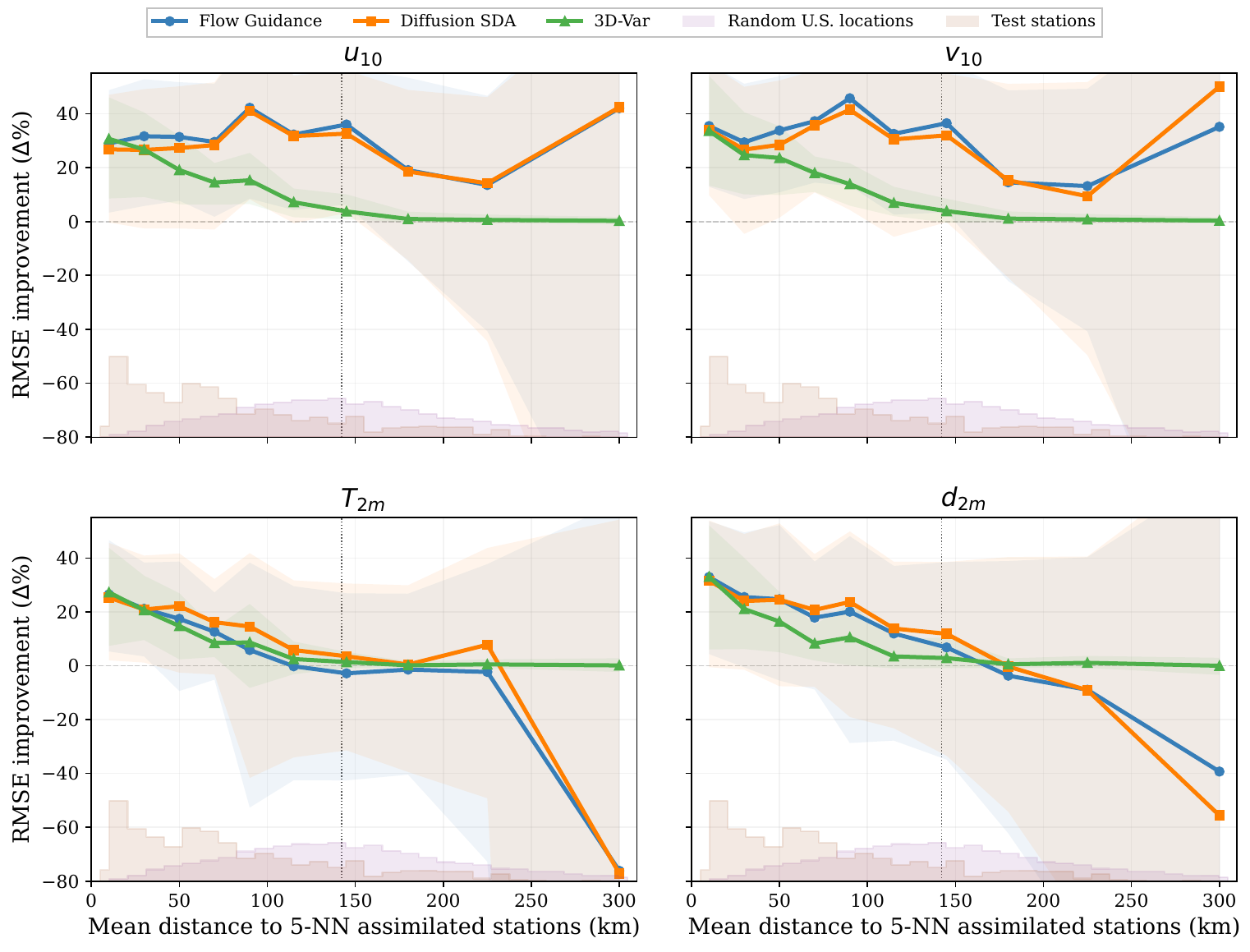}
\caption{Per-variable RMSE improvement vs.\ mean 5-NN distance, sparse benchmark.}
\label{fig:per_var_sparse_si}
\end{figure}

\autoref{fig:per_var_dense_si} and \autoref{fig:per_var_sparse_si} resolve the macro spatial-decay curves of \autoref{fig:spatial_dense} and \autoref{fig:spatial_sparse} per variable. Wind variables (\uten, \vten) decay more slowly than temperature and dewpoint (\ttwom, \dtwom), consistent with the $\sim$$2.4\times$ normalized-space signal imbalance documented in \autoref{app:normalized}. The sign reversal beyond $\sim$$150$\,km in the sparse benchmark is fully driven by negative temperature and dewpoint improvements; wind improvements remain positive past 200\,km.

\subsection{Additional regional case studies}
\label{app:case_studies}

\autoref{sec:case_studies} presented two regional case studies, Rocky Mountains and Great Lakes, that bracket the range of behaviors between methods. This section presents two additional cases, the Los Angeles basin and the Florida peninsula, that reinforce the same pattern: the gap between methods is narrow when the station network densely covers the region and widens when corrections must propagate through poorly-sampled areas.

\begin{figure}[htbp]
\centering
\includegraphics[width=\textwidth]{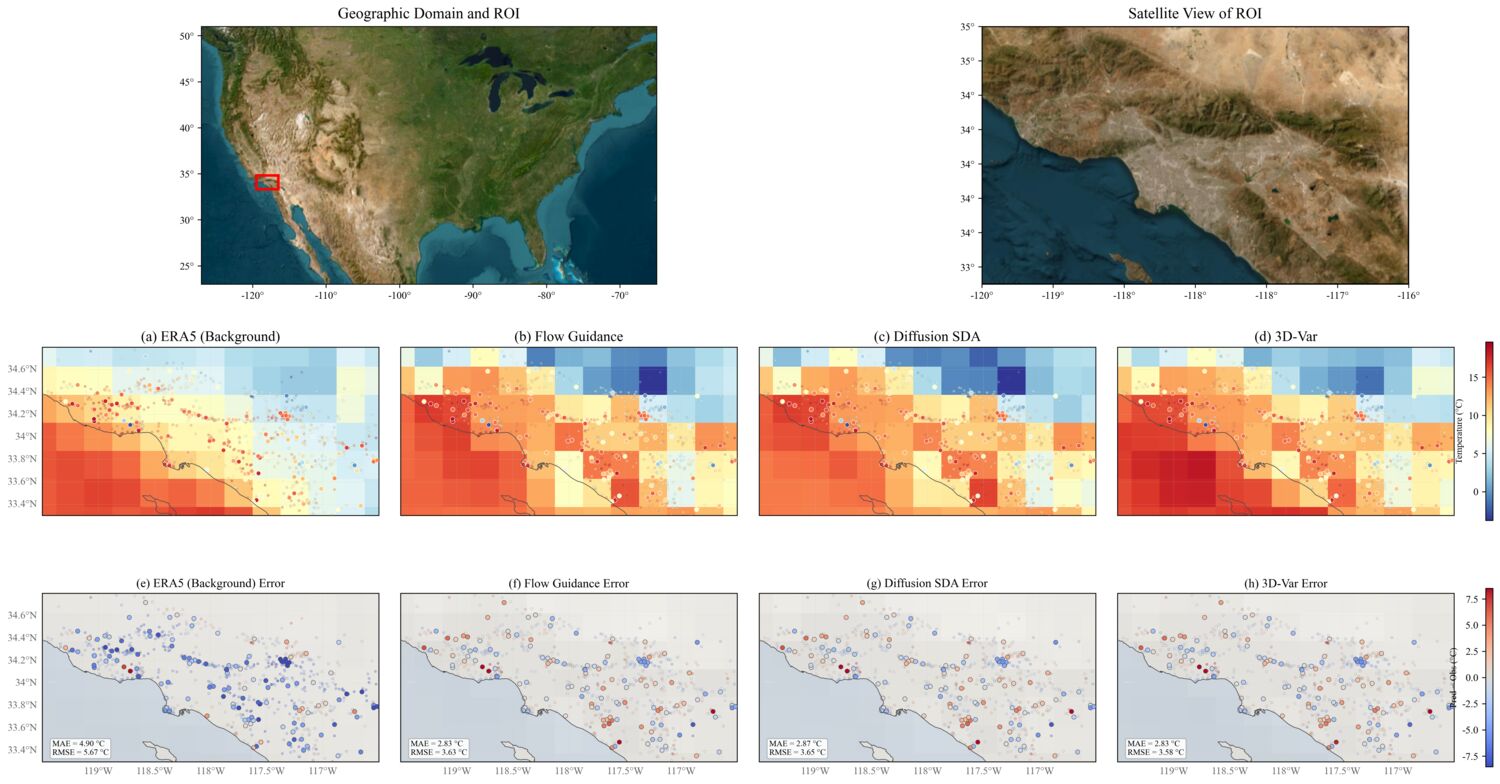}
\caption{Los Angeles basin case study for \ttwom. \textbf{Top row:} geographic context (left) and satellite view (right) of the region of interest. \textbf{Middle row:} (a) ERA5 background, (b) Flow Guidance, (c) Diffusion-SDA, (d) 3D-Var. \textbf{Bottom row:} per-station errors (analysis $-$ observation) for each method, with MAE and RMSE annotated. ERA5 overestimates mountain temperatures and underestimates coastal urban temperatures; all three methods achieve nearly identical corrections on this case (RMSE $3.6$--$3.7$\,K, $36$--$37\%$ reduction).}
\label{fig:case_la}
\end{figure}

\paragraph{Los Angeles basin: complex terrain and dense station network.}
The Los Angeles basin combines sharp terrain (coastal ranges with ${\sim}2000$\,m relief over a few tens of kilometers) with the Pacific shoreline. \autoref{fig:case_la} shows a representative snapshot in which ERA5 overestimates mountain temperatures and underestimates coastal urban temperatures, with test-station MAE of $4.90$\,K (\ttwom) and localized errors reaching ${\sim}13$\,K. The LA basin is densely instrumented; this ROI contains 131 test stations in addition to hundreds of assimilated training stations, so all three methods have ample local observations to anchor the correction. Consistent with the Rockies case, the three methods achieve essentially identical improvement on this snapshot: Flow Guidance to RMSE $3.63$\,K ($36.1\%$ reduction), Diffusion-SDA to $3.65$\,K ($35.7\%$), and 3D-Var to $3.58$\,K ($36.9\%$). At individual stations, Flow Guidance reduces absolute error by up to $9.5$\,K, with roughly one-third of stations improved by more than $4$\,K. The test stations show consistent improvements across all three methods, indicating that the correction generalizes beyond the training observation layout.

\begin{figure}[htbp]
\centering
\includegraphics[width=\textwidth]{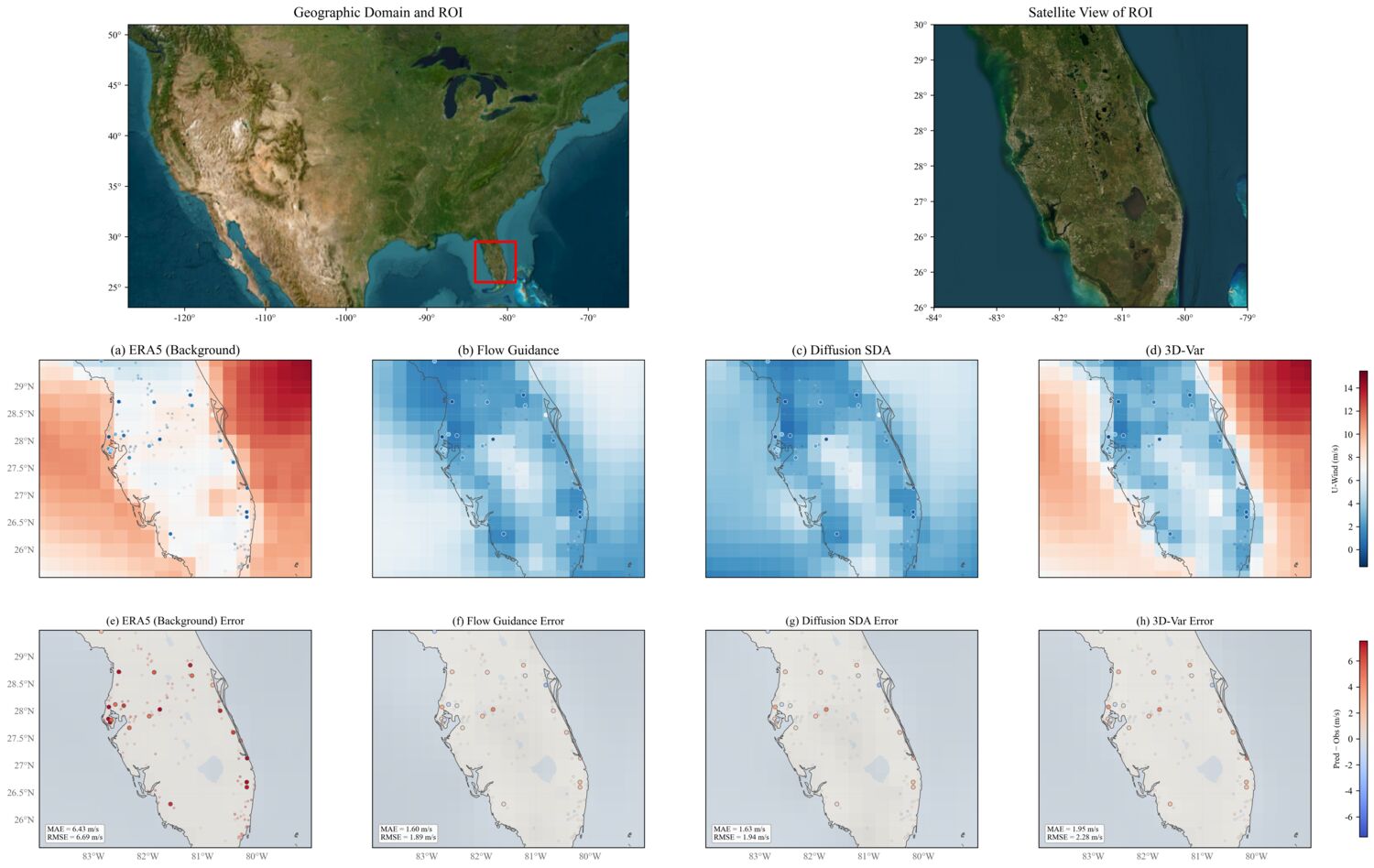}
\caption{Florida peninsula case study for \uten. Panels as in \autoref{fig:case_la}. Flow Guidance and Diffusion-SDA reduce wind RMSE by ${\sim}72\%$ on this snapshot versus $66\%$ for 3D-Var, a ${\sim}6$\,\text{pp} gap that mirrors the Great Lakes finding at smaller magnitude.}
\label{fig:case_florida}
\end{figure}

\paragraph{Florida: coastal winds and offshore propagation.}
The Florida peninsula is bordered on three sides by ocean, so a substantial fraction of the surrounding domain is offshore and unobserved by the station network. \autoref{fig:case_florida} shows a representative snapshot in which ERA5 substantially overestimates \uten\ wind speed at the gauged stations (test-station RMSE $= 6.69$\,m/s on $n=23$ stations). Flow Guidance reduces RMSE to $1.89$\,m/s ($71.8\%$ reduction) and Diffusion-SDA to $1.94$\,m/s ($71.0\%$), while 3D-Var reaches $2.28$\,m/s ($65.9\%$), a ${\sim}6$\,\text{pp} gap. This is smaller than the Great Lakes gap ($8$\,\text{pp} on \uten\ RMSE on the snapshot shown in \autoref{fig:case_lakes}) but qualitatively similar: in both cases, a sizeable portion of the domain lies over water that the station network does not directly cover, and the generative methods retain a margin over 3D-Var at the verifiable shoreline stations.

Taken together with the main-text cases, the pattern at the verifiable test stations is consistent: the gap between the generative methods and 3D-Var is small where observations densely cover the region (Rockies, LA basin) and larger where substantial portions of the surrounding domain are ungauged (Great Lakes, Florida).

\FloatBarrier
\subsection{Full benchmark in physical units}
\label{app:full_benchmark}

\autoref{tab:full_rmse} reports absolute RMSE in physical units (m\,s$^{-1}$ for wind, K for temperature and dewpoint) for the dense $11{,}849$-station benchmark. The headline percentage improvements in main-text \autoref{tab:benchmark_main} are computed from these absolute values relative to the ERA5 baseline.

\autoref{tab:benchmark_main} compares methods at their default ensemble sizes ($M{=}16$ for Flow Guidance and Diffusion-SDA, $M{=}1$ for D-Flow). To control for ensemble averaging, we also evaluate Flow Guidance and Diffusion-SDA at $M{=}1$ (\autoref{app:full_benchmark}). At $M{=}1$, both still outperform D-Flow ($34.0\%$ and $33.4\%$ vs.\ $30.5\%$), confirming that D-Flow's lower accuracy reflects a limitation of noise-space optimization rather than an ensemble-size disadvantage.

\begin{table}[htbp]
\centering
\caption{Absolute RMSE (train/validation/test) on the dense benchmark. Wind variables are in m\,s$^{-1}$; temperature and dewpoint are in K.}
\label{tab:full_rmse}
\tiny
\setlength{\tabcolsep}{2.5pt}\resizebox{\textwidth}{!}{%
\begin{tabular}{l ccc ccc ccc ccc ccc}
\toprule
& \multicolumn{3}{c}{\uten} & \multicolumn{3}{c}{\vten} & \multicolumn{3}{c}{\ttwom} & \multicolumn{3}{c}{\dtwom} & \multicolumn{3}{c}{Average} \\
\cmidrule(lr){2-4}\cmidrule(lr){5-7}\cmidrule(lr){8-10}\cmidrule(lr){11-13}\cmidrule(lr){14-16}
Method & Tr & Va & Te & Tr & Va & Te & Tr & Va & Te & Tr & Va & Te & Tr & Va & Te \\
\midrule
ERA5 (baseline)            & 1.87 & 1.89 & 1.88 & 1.93 & 1.94 & 1.91 & 2.40 & 2.44 & 2.38 & 2.62 & 2.63 & 2.53 & 2.20 & 2.22 & 2.18 \\
3D-Var                     & 0.90 & 1.21 & 1.21 & 0.93 & 1.26 & 1.24 & 1.29 & 1.76 & 1.68 & 1.31 & 1.76 & 1.66 & 1.11 & 1.50 & 1.45 \\
Diffusion-SDA              & 0.88 & 1.16 & 1.16 & 0.92 & 1.18 & 1.17 & 1.43 & 1.73 & 1.67 & 1.36 & 1.69 & 1.61 & 1.15 & 1.44 & 1.40 \\
Diffusion-SDA (PO)         & 0.96 & 1.16 & 1.16 & 1.00 & 1.19 & 1.18 & 1.68 & 1.93 & 1.90 & 1.63 & 1.95 & 1.87 & 1.32 & 1.56 & 1.52 \\
D-Flow                     & 0.92 & 1.21 & 1.22 & 0.97 & 1.23 & 1.22 & 1.59 & 1.87 & 1.82 & 1.55 & 1.84 & 1.79 & 1.26 & 1.54 & 1.51 \\
Flow Guidance              & 0.87 & 1.15 & 1.15 & 0.90 & 1.18 & 1.17 & 1.36 & 1.73 & 1.66 & 1.32 & 1.69 & 1.60 & 1.11 & 1.43 & 1.40 \\
FlowDPS, $K=20$            & 0.73 & 1.25 & 1.26 & 0.76 & 1.26 & 1.27 & 1.15 & 1.91 & 1.86 & 1.14 & 1.90 & 1.86 & 0.94 & 1.58 & 1.56 \\
\midrule
\multicolumn{16}{l}{\emph{Single-sample ($M{=}1$) comparison}} \\
Diffusion-SDA ($M{=}1$)    & 0.89 & 1.18 & 1.18 & 0.93 & 1.20 & 1.19 & 1.48 & 1.79 & 1.73 & 1.42 & 1.77 & 1.69 & 1.18 & 1.48 & 1.45 \\
Flow Guidance ($M{=}1$)    & 0.88 & 1.17 & 1.17 & 0.91 & 1.19 & 1.18 & 1.40 & 1.77 & 1.71 & 1.36 & 1.76 & 1.67 & 1.14 & 1.47 & 1.44 \\
\bottomrule
\end{tabular}%
}
\end{table}

\FloatBarrier
\subsection{Full-year benchmark}
\label{app:full_year}

\paragraph{Snapshot selection.}
For the dense benchmark, we drop time steps with too few available stations by applying a 70/15/15 split to a fixed budget of $8{,}000$ (a round-number threshold chosen below the full pool of $11{,}849$): a time step is dropped if fewer than $5{,}600$ training, $1{,}200$ validation, or $1{,}200$ test stations are available. For the sparse benchmark, the same filter is applied with thresholds scaled by $1{,}000 / 11{,}849$, giving per-split thresholds of $472 / 101 / 101$ (train/validation/test). Both benchmarks retain 360 snapshots: dense has $9{,}818$ stations available per snapshot on average (std $307$, range $8{,}934$--$10{,}481$), and sparse has $833$ stations available (std $28$, range $753$--$903$).

To verify that the 360-snapshot evaluation set is representative, we re-run the three headline methods on every valid hourly snapshot of 2023 ($n = 8{,}664$) under the dense benchmark protocol. Test-station improvements match the 360-snapshot averages to within $0.1$\,\text{pp} on the average test improvement and within $0.3$\,\text{pp} on every individual variable.

\begin{table}[htbp]
\centering
\caption{Full-year benchmark on the dense station network ($8{,}664$ snapshots, 2023). RMSE in physical units (top) and improvement over ERA5 (bottom).}
\label{tab:full_year}
\tiny
\setlength{\tabcolsep}{2.5pt}\resizebox{\textwidth}{!}{%
\begin{tabular}{l ccc ccc ccc ccc ccc}
\toprule
& \multicolumn{3}{c}{\uten} & \multicolumn{3}{c}{\vten} & \multicolumn{3}{c}{\ttwom} & \multicolumn{3}{c}{\dtwom} & \multicolumn{3}{c}{Average} \\
\cmidrule(lr){2-4}\cmidrule(lr){5-7}\cmidrule(lr){8-10}\cmidrule(lr){11-13}\cmidrule(lr){14-16}
Method & Tr & Va & Te & Tr & Va & Te & Tr & Va & Te & Tr & Va & Te & Tr & Va & Te \\
\midrule
\multicolumn{16}{l}{\emph{Absolute RMSE}} \\
ERA5            & 1.86 & 1.87 & 1.87 & 1.93 & 1.94 & 1.92 & 2.40 & 2.44 & 2.39 & 2.61 & 2.62 & 2.52 & 2.20 & 2.22 & 2.17 \\
3D-Var          & 0.90 & 1.21 & 1.21 & 0.93 & 1.26 & 1.23 & 1.29 & 1.76 & 1.69 & 1.31 & 1.76 & 1.66 & 1.11 & 1.50 & 1.45 \\
Diffusion-SDA   & 0.87 & 1.15 & 1.16 & 0.92 & 1.18 & 1.17 & 1.43 & 1.73 & 1.67 & 1.36 & 1.70 & 1.61 & 1.14 & 1.44 & 1.40 \\
Flow Guidance   & 0.87 & 1.14 & 1.15 & 0.90 & 1.17 & 1.16 & 1.36 & 1.73 & 1.67 & 1.32 & 1.69 & 1.60 & 1.11 & 1.43 & 1.40 \\
\midrule
\multicolumn{16}{l}{\emph{Improvement over ERA5 (\%)}} \\
3D-Var          & 51.5 & 35.3 & 34.9 & 52.0 & 35.0 & 35.4 & 46.0 & 27.7 & 29.1 & 49.3 & 32.2 & 33.6 & 49.7 & 32.5 & 33.2 \\
Diffusion-SDA   & 52.6 & 38.3 & 37.5 & 52.4 & 39.2 & 38.9 & 40.0 & 28.7 & 29.6 & 47.4 & 34.5 & 35.7 & 48.1 & 35.2 & 35.4 \\
Flow Guidance   & 53.1 & 38.6 & 37.9 & 53.2 & 39.2 & 39.1 & 43.1 & 29.0 & 29.9 & 48.9 & 34.8 & 35.8 & 49.6 & 35.4 & 35.7 \\
\bottomrule
\end{tabular}%
}
\end{table}

\end{document}